\documentclass[10pt,journal,compsoc]{IEEEtran}

\usepackage{amsmath,amssymb,amsfonts}

\usepackage[dvipsnames,table]{xcolor}
\definecolor{myblue}{rgb}{.894,.937,.965}
\definecolor{mydark_blue}{rgb}{.68,.776,.906}
\definecolor{NO3}{rgb}{.749,.902,.808}
\definecolor{NO2}{rgb}{.999,.973,.773}
\definecolor{NO1}{rgb}{0.98, 0.78, 0.57}

\usepackage{graphicx}

\usepackage{array}
\usepackage{booktabs}
\usepackage{multirow}
\usepackage{multicol}
\usepackage{colortbl}
\usepackage{makecell}
\usepackage{adjustbox}
\usepackage{longtable}

\usepackage[caption=false,font=normalsize,labelfont=sf,textfont=sf]{subfig}
\usepackage{stfloats}
\usepackage{rotating}

\usepackage{textcomp}
\usepackage{url}
\usepackage{cite}
\usepackage{soul}

\usepackage[breaklinks,colorlinks]{hyperref}

\usepackage{pifont}
\usepackage{wasysym}       % \CIRCLE \LEFTcircle \Circle for Harvey balls
\newcommand{\cmark}{\ding{51}}   % checkmark

\newcommand{\queryitem}[3]{%
  \par\vspace{1pt}\noindent
  \textsf{\textbf{#1}}\enspace\textit{#2}\par\nobreak
  \noindent\hspace*{1.9em}%
  \begin{minipage}[t]{\dimexpr\columnwidth-1.9em\relax}
    \raggedright\color{black!72}\footnotesize\ttfamily #3
  \end{minipage}\par}

\begin{document}

\title{From Generation to Simulation: How Far Are World Models\\ from Being True Simulators?}

\author{Tong~Wang, Huan~Deng, Mucheng~Yang, Yang~He, Xiaohui~Kuang, and Gang~Zhao%
\thanks{The authors are with the Institute of Systems Engineering, Academy of
Military Sciences, Beijing 100101, China. E-mail: tongwss@foxmail.com;
DHuan56@foxmail.com; 4349467ymc@gmail.com; heang@stu.xidian.edu.cn;
xhkuang@bupt.edu.cn; bisezhaog@163.com.}%
\thanks{Corresponding author: Gang Zhao.}%
\thanks{ORCID iDs: Tong Wang, 0000-0001-6981-916X; Huan Deng,
0000-0002-5116-1766; Mucheng Yang, 0009-0008-4567-4099; and Gang Zhao,
0009-0006-8668-777X.}%
\thanks{Manuscript prepared July 2026. This study analyzes a curated corpus of
200 papers published between 2018 and 2026.}}

\markboth{From Generation to Simulation: A Capability Audit,~2026}%
{From Generation to Simulation: Benchmarking Generative World Models Against Simulator Capabilities}

\IEEEtitleabstractindextext{
\begin{abstract}
With the rapid progress of diffusion models and large-scale video generation, generative world models are increasingly expected to replace traditional simulators---physics engines, game engines, and reinforcement-learning environments---and to become a new generation of data-driven, general-purpose simulation environments. Yet how far this road from \emph{generation} to \emph{simulation} still runs remains without a systematic answer: existing surveys mostly organize the literature along the architecture, function, or application domain of world models themselves, and thus struggle to characterize their true gap as simulators. In this paper, we present a systematic, capability-based study using an external yardstick, namely the eight capabilities of a traditional simulator: asset construction, physics engine, interaction, controllability, stability, state feedback, diversity, and evaluation metrics. Around this route-independent yardstick, we first trace the evolution of three main technical routes---latent dynamics, video generation, and joint-embedding prediction---and their recent trend toward convergence; we then map exactly 200 representative works from 2018 to June 2026 onto the eight capabilities, systematically comparing the coverage and gaps of each route. Our analysis shows that, aided by advances such as autoregressive-diffusion distillation, latent-action learning, and world-foundation-model platforms, world models have achieved functional substitution in interaction and controllability for specific scenarios, yet remain a critical step short of traditional simulators in formally guaranteeing physical law, in the richness of structured state feedback, and in the reproducibility of long-horizon evolution; among these, state feedback is a shortcoming that cuts across all technical routes yet is the most neglected: only 6 of 163 implementation papers expose a runtime interface for queryable entity states or physical parameters. Accordingly, we look ahead to six research directions---formalized physics, a unified action interface, first-class state feedback, long-horizon stability, downstream-utility evaluation, and cross-route hybridization---charting an evidence-grounded path toward the next generation of true simulators. Project page: \url{https://github.com/AtongWang/world-model-simulators}.
\end{abstract}

\begin{IEEEkeywords}
World models, generative simulation, simulator capabilities, physical plausibility, state feedback, long-horizon stability, embodied AI, autonomous driving, evaluation.
\end{IEEEkeywords}}

\maketitle
\IEEEdisplaynontitleabstractindextext
\IEEEpeerreviewmaketitle

\IEEEraisesectionheading{\section{Introduction}}

\IEEEPARstart{T}he core idea of a \emph{world model} is to predict the future state of an environment by learning the physical regularities that govern space and time. Since the pioneering work of Ha and Schmidhuber in 2018, which built an interactive latent environment from a VAE and an RNN \cite{1803.10122}, world models have undergone several paradigm shifts---from latent dynamics to video-generative, and further to joint-embedding predictive (JEPA) formulations---with a steadily expanding capability boundary and application scope that now spans game engines \cite{2408.14837,2501.08325,2604.08995,2411.00769}, autonomous-driving simulation \cite{2403.09630,2311.13549,2606.03159,2412.18607}, robotic manipulation \cite{2605.16395,2506.09985,2606.01027,2508.17600,2603.16669}, and open-domain exploration \cite{2402.15391,2406.09394,2511.08536}. This trend naturally raises a fundamental question: given that the core capability of a world model is to predict future states, can it replace a traditional simulator and become a new generation of data-driven, general-purpose simulation environment?

\begin{figure*}[t]
\centering
\includegraphics[width=\textwidth]{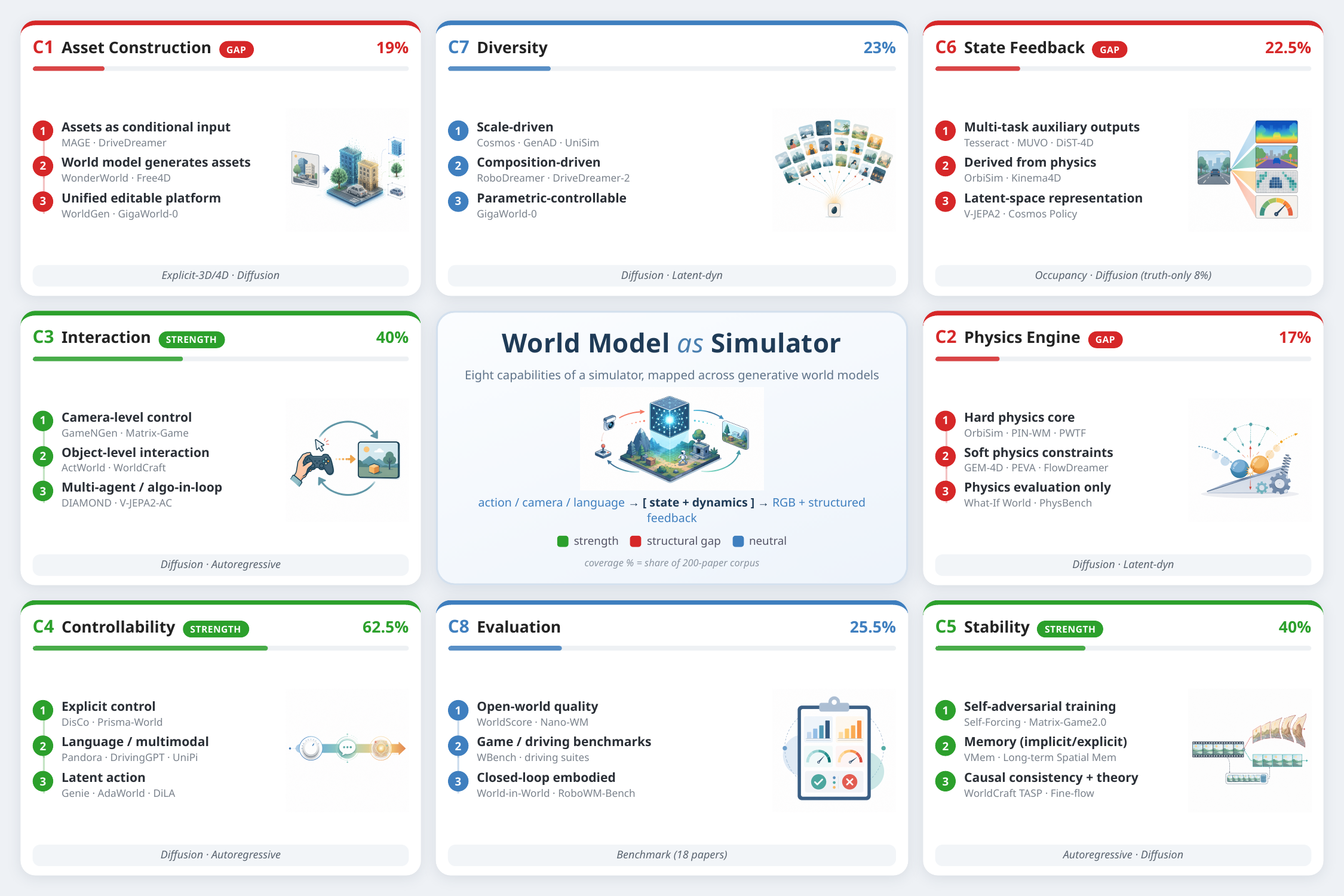}
\caption{\textbf{Overview of the survey.} Centred on ``world model as simulator'', the eight capabilities of a traditional simulator are arranged around it: asset construction, physics engine, interaction, controllability, stability, state feedback, diversity, and evaluation metrics. Each capability panel reports its coverage as a principal contribution across the 200-paper corpus, a status tag (\emph{strength} / \emph{structural gap} / \emph{neutral}), and representative works. Controllability (62.5\%), interaction (40\%), and stability (40\%) are tagged as strengths; asset construction (19\%), the physics engine (17\%), and state feedback (22.5\%) as cross-route structural gaps; and diversity (23\%) and evaluation metrics (25.5\%) remain neutral.}
\label{fig:overview}
\end{figure*}

Following the perspective of Cosmos Policy \cite{2601.16163}, world models can be divided into two paradigms by function. The \emph{policy} route uses a world model to directly produce executable actions or value functions that support control decisions; for example, Cosmos Policy, V-JEPA~2-AC \cite{2506.09985}, and $\tau_0$-WM \cite{2606.01027} unify policy learning and video prediction within a single framework. The \emph{generative} route instead aims to generate future observations---images, video, 3D structure, or structured state---to serve simulation, planning, or data augmentation \cite{2501.03575,2511.19861}. The policy route also models world dynamics, but its output is an action vector or a scalar value, which differs fundamentally from the state-feedback interface of a traditional simulator. This survey therefore focuses primarily on the generative route---works that generate future observations as pixels, tokens, or explicit 3D representations---and examines to what extent they can play the role of a traditional simulator. Formally, the world model we consider is a mapping $T\colon\, \mathcal{S}\times\mathcal{A}\to\Pi(\mathcal{S})$ that learns the distribution of future states given the current state and action, thereby approximating the true environment dynamics \cite{2601.16163}.

Although world models exhibit potential to surpass traditional simulators in visual realism and learnability, their current state remains markedly short of a rigorous simulator. Hallucination is a systematic problem shared by almost all generative models, manifesting as objects appearing or disappearing without cause \cite{2408.14837}, geometry drifting over long rollouts \cite{2604.13036,2605.31336}, collision and contact behaviour violating physical intuition \cite{2605.27589,2603.23376}, and identical actions producing inconsistent visual consequences at different times \cite{2606.07967}. The root of these problems is that current world models are conditional-distribution samplers rather than physical-evolution solvers: they learn the conditional distribution of visual patterns in the training data, not the invariant physical laws of the real world \cite{2405.03520}. In addition, world models still exhibit critical gaps against traditional simulators in the precision of controllability, the richness of state feedback, and the verifiability of their output. Section~\ref{sec:c6} reports a paper-level audit of state-feedback interfaces across all 163 implementation papers.
Motivated by these observations, this survey systematically answers three core questions: (a) Can the generative process of a generative world model become a simulator in the strict sense? (b) On each simulator capability, how far has each technical route progressed, and how far does it remain from the target? (c) In which directions should the next steps be taken? To answer them, Section~\ref{sec:collection} describes the literature-collection process and inclusion criteria; Section~\ref{sec:background} provides the technical background of world models and the capability yardstick of traditional simulators; Section~\ref{sec:comparison}, the core of the survey, uses eight key simulator capabilities as a unified framework (asset construction C1, physics engine C2, interaction C3, controllability C4, stability C5, state feedback C6, diversity C7, evaluation metrics C8), summarized in Figure~\ref{fig:overview}, and compares, capability by capability, the concrete works, progress, and shortcomings of each route---previewing our central finding that controllability, interaction, and stability stand out as relative strengths, whereas asset construction, the physics engine, and state feedback remain structural gaps; Section~\ref{sec:future} distills six future research directions from the comparative analysis; and Section~\ref{sec:conclusion} gives our concluding judgement.

The main contributions of this capability-based study are fourfold:
\begin{itemize}
\item \textbf{A comparative analysis framework.} We examine generative world models through the unified yardstick of a traditional simulator's capabilities, establishing an eight-dimensional comparison system.
\item \textbf{A reproducible full-corpus evidence map.} Based on paper-by-paper coding of a curated 200-paper corpus, we quantify the progress and key shortcomings of each technical route capability by capability.
\item \textbf{Quantification of the state-feedback shortcoming.} We identify state feedback---a capability so far severely neglected yet indispensable to a simulator---as a structural shortcoming, and quantify this structural absence through a five-level labelling scheme (B1--B5).
\item \textbf{An actionable research roadmap.} We provide six clear and actionable research directions, each backed by nascent work as evidence of its feasibility.
\end{itemize}

The corpus annotations, evidence map, and analysis scripts supporting this study
are maintained on the project page:
\url{https://github.com/AtongWang/world-model-simulators}.

\section{Literature Collection}\label{sec:collection}

\subsection{Search Strategy and Inclusion Criteria}

Our literature collection covers four complementary sources---Google Scholar, arXiv, DBLP, and Crossref---with a search cut-off of 30 June 2026. Each source has its own emphasis: arXiv aggregates the vast majority of the field's latest preprints and is the primary source for frontier work; Google Scholar is used for cross-database broad discovery and citation tracing; and DBLP and Crossref respectively provide canonical bibliographic records for computer-science conferences/journals and DOI-based publication metadata, used to verify the formal publication status of each paper. The purpose of using multiple sources is to cover the blind spots of any single database---arXiv is timely but holds only preprints, whereas DBLP and Crossref are authoritative but lag in indexing the newest work.

In constructing the queries, rather than using a single broad term, we decomposed each topic into several concept terms combined by Boolean \texttt{AND}, striking a balance between recall and precision. For example, for the topic of interactive video world models we used \texttt{interactive AND world AND model AND simulation AND video AND generation}; for the differentiable-physics direction we used \texttt{differentiable AND physics AND simulation AND world AND model AND learned}. Around the mainstream technical routes (latent dynamics, autoregressive, diffusion, game engines, driving simulation, embodied robotics, evaluation benchmarks) and several angles easily missed by the main lines (JEPA self-supervision, language models as world models, 3D Gaussians and NeRF, long-horizon memory, occupancy modelling, multimodality, physical-plausibility evaluation), we designed more than twenty queries in total; the complete list appears in Appendix~\ref{app:queries}. Search results were sorted by relevance, truncated to the top-ranked entries per query, automatically de-duplicated against the accumulated set, and then screened manually.

Because keyword search cannot exhaust a rapidly evolving field, we further expanded the corpus along the citation network. On one hand, we combed the reference lists of six representative surveys to fill gaps, covering world models for robot learning \cite{2605.00080}, interactive video world models \cite{2606.01164}, world models for robotic manipulation \cite{2606.00113}, embodied intelligence fusing physical simulators and world models \cite{2507.00917}, unified 2D/video/3D/4D multimodal generation \cite{2503.04641}, and world models for embodied AI \cite{2510.16732}. On the other hand, we expanded along the forward and backward citations of four foundational or milestone works---World Models \cite{1803.10122}, DreamerV3 \cite{2301.04104}, Genie \cite{2402.15391}, and Cosmos \cite{2601.16163}---to include peripheral but relevant work. In addition, some industrial-grade world models (such as DeepMind's Genie~2/3) are released only as technical reports or blog posts without a formal paper; such works are cited individually in the text and are not counted in the searchable-corpus statistics.

Because arXiv is a preprint platform and cannot reflect the formal publication status of a paper, we cross-verified every included paper in DBLP and Crossref and generated canonical references accordingly: anchored on each paper's authoritative record (title, authors, year), we searched both databases by title similarity and first-author matching, judging a paper formally published only when the title matched closely, the authors agreed, and the venue was not a preprint. Verification showed that, of the 200 papers, 72 have been published at peer-reviewed conferences or journals---22 at ICML, NeurIPS, or ICLR; 20 at CVPR, ICCV, or ECCV; 6 at CoRL, ICRA, or RSS; 6 at AAAI; 3 in \emph{Nature}; and 15 others at ICASSP, IROS, WACV, TPAMI, TIP, \emph{ACM Computing Surveys}, and domain-specific journals---while the remaining 128, mostly recent results from late 2025 to 2026, were still arXiv preprints as of the search date. For published work, the references always adopt the metadata of the formal version; verification also corrected two mislabelled identifiers and three publisher-name errors in the text.

Inclusion follows one core criterion: whether a paper substantively accepts some action or conditional input (camera pose, keyboard/mouse operation, language instruction, latent action, trajectory signal, or 3D geometric condition) and, on that basis, performs a forward rollout to generate future states (video frames, 3D scenes, occupancy, point clouds, or structured state), with results that can serve interaction, control, or planning. On this basis, we exclude four kinds of work:
\begin{enumerate}
\item pure text-to-video/image-to-video backbones unrelated to interaction or conditional control;
\item work targeting pure representation learning without action conditioning;
\item pure prediction models that forecast only object trajectories without environment-state evolution;
\item papers whose title or abstract contains the phrase ``world model'' but which actually belong to non-generative directions such as language-model encoding or cognitive modelling.
\end{enumerate}

\subsection{Corpus Overview}

\begin{figure*}[t]
\centering
\includegraphics[width=\textwidth]{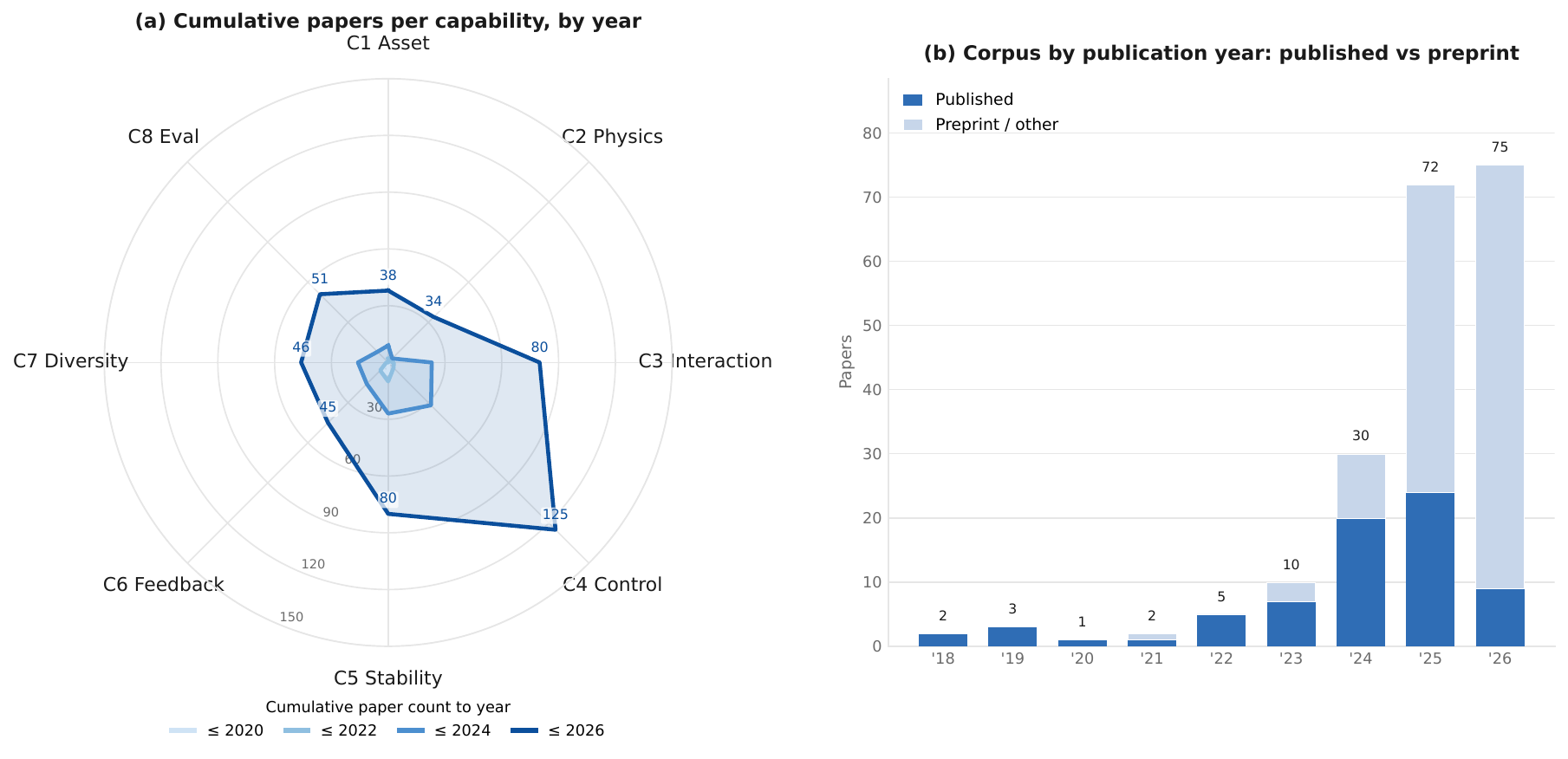}
\caption{\textbf{Research attention across the eight simulator capabilities and the corpus composition, 2018--2026.} \textbf{(a)} Cumulative papers contributing each capability, drawn as nested year slices (to 2020/2022/2024/2026); the radial value is the number of papers for which that capability is a principal contribution, and the number at each vertex is its 2026 total. Controllability, interaction, and stability dominate, while asset construction, the physics engine, and state feedback stay low. \textbf{(b)} The 200-paper corpus by publication year, split into published (72) and preprint / other (128); the corpus is increasingly preprint-dominated in 2025--2026.}
\label{fig:radar}
\end{figure*}

After several rounds of screening, 200 records were finally included, spanning 2018 to 2026. We performed structured classification at the abstract level or above for all records---covering paradigm membership, technical route, action-interface type, state-feedback capability where annotated, and principal contribution dimensions, covering the milestone works of all major routes as well as representative benchmarks and surveys. The corpus contains 163 implementation papers assigned to six technical families: latent-dynamics and latent-action models (29), autoregressive generation (37), diffusion models (63), JEPA-predictive models (5), explicit 3D/4D reconstruction (25), and occupancy-centric methods (4). The remaining 37 records provide context: evaluation benchmarks (18), surveys (16), and others (3). Figure~\ref{fig:radar} profiles the complete 200-record corpus along two views: (a) the cumulative number of papers contributing each of the eight capabilities, sliced by year (2020/2022/2024/2026); and (b) the corpus by publication year, split into published versus preprint. Two trends stand out. First, research attention is highly uneven: controllability, interaction, and stability accumulate far more papers than asset construction, the physics engine, and state feedback---a gap the capability-by-capability analysis of Section~\ref{sec:comparison} examines in detail. Second, the corpus is increasingly preprint-dominated: 72 of the 200 records are published while 128 remain arXiv preprints or other records, the latter concentrated in 2025--2026, reflecting how fast the field is moving relative to the review cycle.

\subsection{Distinction from Existing Surveys}

Our corpus contains 16 surveys, grouped by coverage into five categories: general world models \cite{2411.14499,2511.08585,2405.03520}, autonomous driving \cite{2501.11260,2502.10498,2411.02914,long2026}, robotics and embodiment \cite{2605.00080,2606.00113,2507.00917,2510.16732,evolving2026}, interactive video \cite{2606.01164}, and efficiency/multimodality \cite{2603.28489,2503.04641,2506.20134}. Their common feature is that they centre on the world model itself, classifying by architecture, function, or application domain. The core difference of this survey is that it uses simulator capability as an \emph{external} yardstick: the eight capabilities (C1--C8) derive from the capability set of traditional simulators (physics engine G1, game engine G2, RL environment G3) rather than from the internal attributes of world models. This externality makes the yardstick independent of any specific route, so heterogeneous routes can be projected into a unified capability space for quantitative comparison.

Methodologically, this study is distinguished by two empirical mechanisms. The first is full-corpus evidence mapping: we annotate all 200 records paper by paper with paradigm, route, one to three principal-contribution capabilities, and a summary of novelty and evidence. The resulting distributions measure research attention rather than capability attainment. The second is a five-variable state-interface audit across all 163 implementation papers: ego information (B1), sensor output (B2), reward and termination (B3), runtime entity/physics annotations (B4), and closed-loop interaction (B5). Present, absent, and unresolved judgements are reported separately so that insufficient evidence is not treated as absence.

Relative to the most relevant surveys, this survey forms a complementary division of labour rather than a replacement. Reference \cite{2605.00080} concerns how world models assist policy learning, whereas this survey concerns whether the simulator attributes of world models themselves are complete; Reference \cite{2606.01164} concerns the fluency of interaction (controllability, memory, real time), whereas this survey further examines the breadth and depth of interaction objects (algorithm-in-the-loop and human-in-the-loop); Reference \cite{2606.00113} provides a fine-grained functional taxonomy, which this survey projects into a capability-shortcoming space; and the judgement in \cite{2405.03520} that Sora possesses preliminary physical understanding rather than strict simulation aligns in direction with our core conclusion, for which our full analysis of 200 papers provides broader empirical support. In short, existing surveys answer what world models exist and what they can do; this survey asks what a world model still \emph{cannot} do as a simulator, how far it remains, and why---thereby turning the proposition of ``world model as simulator'' from a vision into an actionable research agenda with concrete dimensions, checkable data, and clearly identified shortcomings.

\section{Background}\label{sec:background}

\subsection{Concept and Evolution of World Models}

\begin{figure*}[!t]
\centering
\includegraphics[width=\textwidth]{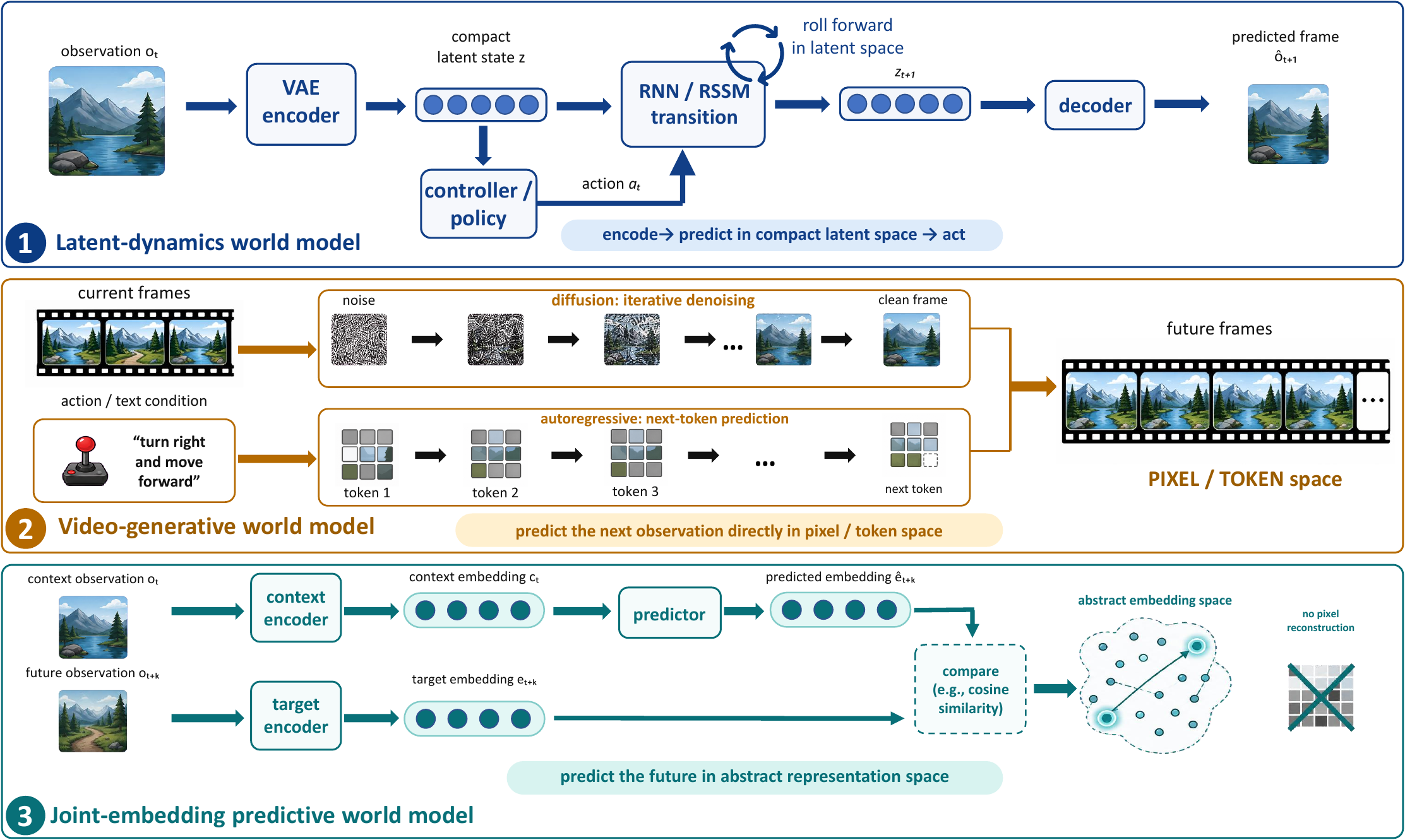}
\caption{\textbf{Three principal routes to world modelling, distinguished by prediction space.} A latent-dynamics model encodes observations into a compact state and rolls that state forward for decoding and control; a video-generative model predicts future observations directly in pixel or token space through diffusion or autoregressive generation; and a joint-embedding predictive model predicts a future representation against a target embedding without reconstructing pixels. The routes therefore expose different simulator interfaces: compact state for imagination and control, rendered observations for direct inspection, and abstract embeddings for efficient downstream prediction and planning.}
\label{fig:three-routes}
\end{figure*}

World models have developed along three main technical routes and have recently shown a clear trend toward convergence. This section reviews that evolution along the twin threads of chronology and modelling mechanism: first the latent-dynamics route, which was the earliest to model environment dynamics in a compact latent space; then the video-generation route, which predicts the future directly in pixel or token space; next the Joint-Embedding Predictive Architecture, which instead predicts in an abstract representation space; and finally the overall trend, since 2024, of these three routes borrowing from one another and moving toward hybrid fusion. This trajectory reflects both a migration of the representation space---from latent state, to pixels, to abstract embeddings---and a shift of research emphasis from policy learning, to high-fidelity generation, to downstream utility. Figure~\ref{fig:three-routes} makes the principal architectural distinction explicit: the three routes differ primarily in the space in which future evolution is represented and predicted.

\subsubsection{Latent-Dynamics World Models}

The concept of world models traces back to the pioneering work of Ha and Schmidhuber in 2018 \cite{1803.10122}, whose core idea models the environment as three components: a VAE that compresses high-dimensional observations into a compact latent state, an MDN-RNN-based dynamics model that predicts future states in the latent space, and a linear controller that performs policy learning within this latent environment. This framework established the basic paradigm of the world model as an internal simulator of the environment, its key being the reduction of complex high-dimensional environment interaction to two steps---prediction and control---in latent space. Subsequently, PlaNet \cite{1811.04551} and the Dreamer series \cite{1912.01603,2010.02193,2301.04104} systematically developed this framework into a latent-dynamics system based on the Recurrent State-Space Model (RSSM). RSSM contains both deterministic and stochastic latent-state variables, enabling long-horizon imagination in a compact latent space and policy optimization thereupon. DreamerV2 \cite{2010.02193} raised the representational power of world models to Atari human level via discrete latent representations, while DreamerV3 \cite{2301.04104}, through symlog prediction and world-model regularization, achieved stable training without manual tuning across more than 150 environments. The core value of this route is \emph{learning in imagination}: DayDreamer \cite{2206.14176} deployed Dreamer on a real robot for online learning, using the world model to accelerate simulation of real interaction in latent space; MoDem-V2 \cite{modem2024} follows the same idea, extending a visuo-motor latent-dynamics world model to online learning for real-robot manipulation and confirming the paradigm's viability on contact-rich tasks.

\subsubsection{Video-Generative World Models}

In parallel, video-generative world models developed rapidly between 2023 and 2026. Unlike the latent-dynamics route that operates in a compact latent space, the video-generation route models the future directly in pixel or token space; its advantages are high visual realism, direct suitability for human evaluation, and native support for generating complex visual scenes. Early works such as GAIA-1 \cite{2309.17080} focused on neural world models for driving scenes, while Genie \cite{2402.15391} inferred latent actions without labels from internet videos and generated interactive 2D game worlds. Sora then prompted broad discussion of whether video-generation models can serve as world simulators \cite{2405.03520}. Cosmos \cite{2601.16163,2501.03575} packages this route as a World Foundation Model platform for Physical AI, including a video-curation pipeline, pretrained models, tokenizers, and fine-tuning tools. At the mechanism level, video-generative world models fall mainly into diffusion and autoregressive classes. SimPLe \cite{1903.00374} showed that a video-prediction world model improves the sample efficiency of RL on Atari, while Phenaki \cite{2210.02399} achieved variable-length video generation driven by time-varying text prompts. Representative diffusion works include DIAMOND \cite{2405.12399}, which trains RL agents in a diffusion world model; GameNGen \cite{2408.14837}, which demonstrates about 20-FPS interaction in a constrained game domain; and the Matrix-Game series \cite{2508.08601,2512.14614,2604.08995}, which reports up to 720p and about 40 FPS. Representative autoregressive works include IRIS \cite{2209.00588}, which combines discrete autoencoding with an autoregressive Transformer; Oasis and RELIC \cite{2512.04040}, with KV-cache camera-aware memory for real-time generation; GameFactory \cite{2501.08325}, with scene-generalizing keyboard/mouse control; and GameGen-X \cite{2411.00769}, an interactive, controllable open-world game DiT. The mechanisms are also converging: causal-forcing distillation \cite{2605.30263} and few-step autoregressive diffusion \cite{2508.13009} combine diffusion-based generation with autoregressive inference.

\subsubsection{Joint-Embedding Predictive Architecture}

The third route is the Joint-Embedding Predictive Architecture (JEPA), represented by V-JEPA \cite{2404.08471} and its successors V-JEPA~2 \cite{2506.09985} and UWM-JEPA \cite{2605.25313}. JEPA's design philosophy differs sharply from the two generative routes above: it predicts the future not in pixel space but in a learned abstract representation space. Through self-supervised pretraining of a 1.2B-parameter ViT on over one million hours of internet video, combined with fine-tuning on merely 62 hours of unlabelled robot data, V-JEPA~2 achieves zero-shot transfer Franka manipulation, requiring no data from that robot and no task-specific training or reward signal. In planning efficiency, V-JEPA~2-AC requires only about 16 seconds per action for latent-space planning, whereas pixel-generation planning under comparable conditions requires about 4 minutes---roughly a 15$\times$ speed advantage \cite{2506.09985}. UWM-JEPA \cite{2605.25313} further extends the JEPA framework to partially observable settings: by introducing a density-matrix latent representation and a unitary-transformation predictor, it maintains the structural integrity of belief space under blind rollouts with occluded observations, reaching 0.77 accuracy in a five-step forward-simulation task, versus 0.53 for a parameter-matched LSTM-JEPA. A cluster of deepening work has recently appeared on the mechanism of JEPA as a world model: a systematic dissection of the success factors of JEPA in physical planning \cite{2512.24497} points out that the combination of action conditioning and prediction in representation space (rather than pixel space) is key to JEPA's support for planning, providing an interpretable design rationale for this route; variational JEPA \cite{2601.14354} turns the originally deterministic predictor into a probabilistic latent-space world model that can characterize multimodal uncertainty about the future (so far verified only at the theoretical and toy-experiment level); and, addressing test-time representation collapse in JEPA, one work \cite{2602.18639} repairs it by learning invariant visual representations and, on top of them, realizes reward-free latent-space MPC planning. A unified probing study \cite{2606.07687} further shows that it is the prediction (rather than reconstruction) objective that drives the action-relevance of video-world-model latents, giving probe-level evidence for the claim that representation-space prediction beats pixel reconstruction. The cost of the JEPA route, however, is that it produces no readable pixel output, and its latent-vector feedback is opaque to non-learning downstream consumers.

\subsubsection{The Trend of Route Convergence}

Between 2024 and 2026, a clear trend of route convergence emerged; noteworthy instances include the hybridization of diffusion and autoregression (causal-forcing distillation \cite{2605.30263}, few-step autoregressive diffusion \cite{2508.13009}), the complementarity of generative and JEPA approaches (V-JEPA~2-AC's latent-conditioned planning \cite{2506.09985}), the synergy of explicit 3D and implicit generation (Lyra~2.0's \cite{2604.13036} geometry routing plus generated appearance, and GWM's \cite{2508.17600} use of a 3D-Gaussian world model as a neural simulator, built on the 3D Gaussian Splatting representation \cite{kerbl2023gaussians}), and the unification of VLA and world models ($\tau_0$-WM \cite{2606.01027} and WorldVLA \cite{2506.21539} integrating policy learning, video prediction, and action evaluation in a single framework). These trends may foreshadow the architectural direction of the next generation of world models: no longer a pure diffusion or latent-dynamics model, but a hybrid system composed of functional modules.

\subsection{Capability Yardstick of Traditional Simulators}

Before comparing world models, we must first make explicit the capability yardstick of a rigorous simulator. Three classes of traditional simulator baselines jointly define the capability set that a functionally complete simulation environment should possess: physics engines (e.g., MuJoCo \cite{todorov2012mujoco}, Isaac Sim \cite{nvidia2026isaacsim}, and PyBullet \cite{coumans2021pybullet}; denoted G1), game engines (e.g., Unreal Engine \cite{epicgames2026unreal} and Unity \cite{juliani2020unity}; G2), and reinforcement-learning training environments (e.g., the Atari-based Arcade Learning Environment \cite{bellemare2013ale}, CARLA \cite{dosovitskiy2017carla}, and DMControl \cite{tunyasuvunakool2020dmcontrol}; G3). We formalize the eight capabilities as follows:
\begin{itemize}
\item \textbf{Asset construction (C1)}: traditional simulators rely on artists and modellers to hand-create or procedurally generate static assets such as maps, characters, objects, and materials; the core strength is fully controllable, instance-level decomposition and editability---every object can be independently selected, modified, or replaced---at the cost of high labour and limited content generalization.
\item \textbf{Physics engine (C2)}: based on numerical solvers (rigid-body dynamics, collision detection, contact-force computation, elastic deformation, etc.), it provides physically deterministic motion simulation, whose formal guarantees---energy and momentum conservation, collision impenetrability---are the cornerstone of scientific computing and safety-critical simulation.
\item \textbf{Interaction (C3)}: it supports any agent interacting with any element of the environment at any moment, with response latency down to the millisecond level, through interfaces such as API calls, physical controllers, or scripting languages.
\item \textbf{Controllability (C4)}: the configuration and evolution of the simulated environment are fully controlled by the user through a precise parametric interface, with no approximation or semantic ambiguity.
\item \textbf{Stability (C5)}: based on deterministic solvers, the simulator's evolution is fully reproducible for the same input sequence.
\item \textbf{State feedback (C6)}: a traditional simulator can extract rich structured information from any intermediate state, including ego pose (B1), sensor data (B2), task-evaluation signals (B3), and runtime entity/physics annotations (B4), and supports closed-loop interaction (B5).
\item \textbf{Diversity (C7)}: controlled scene diversity through parameter randomization and procedural generation.
\item \textbf{Evaluation metrics (C8)}: a mature evaluation system, including reward-based RL performance metrics and directly accessible ground-truth labels.
\end{itemize}

On this yardstick, we express the working definition of a rigorous simulator as a function $T\colon\, \mathcal{S}\times\mathcal{A}\to\Pi(\mathcal{S}\times\mathcal{F})$, where $\mathcal{S}$ is the state space, $\mathcal{A}$ the action space, and $\mathcal{F}$ a structured state-feedback space containing at least B1 through B4, subject to four conditions: state evolution is physically consistent and verifiable, temporally consistent without drift, statistically reproducible (deterministic or controllably stochastic), and supports closed-loop interactive evaluation.

\section{World-Model--Simulator Comparison}\label{sec:comparison}

Before the capability-by-capability comparison, we first compare world models and traditional simulators at a high level. World models hold two fundamental advantages over traditional simulators. First, a world model learns environment dynamics from data, without manual geometric modelling, material editing, or physical-parameter calibration. Cosmos's \cite{2501.03575} WFM is positioned as a digital twin for Physical AI and can generate customized world models for driving, robotic manipulation, indoor navigation, and other scenarios from one pretrained foundation model, whereas a traditional pipeline must re-model each new scene. Second, generative world models---especially diffusion-based ones---can achieve visual realism that approaches or even surpasses real photographs; in GameNGen's \cite{2408.14837} human study, only 58--60\% of subjects could distinguish real from generated \textsc{Doom} screenshots, whereas a traditional rendering pipeline, though stronger in physical accuracy, is limited in expressiveness by model complexity and compute.

World models, however, exhibit three recurring limitations. The \emph{first} is the absence of formal guarantees. Traditional simulators can verify physical invariants at each step, whereas world models generally provide plausibility evidence rather than proof of physical correctness. The \emph{second} is a sparse information interface: only 45 of 163 implementation papers explicitly report sensor-level outputs beyond RGB, whereas traditional simulators expose pose, depth, LiDAR, semantics, collision events, rewards, and labels through APIs. The \emph{third} is limited closed-loop reproducibility. Traditional simulators commonly expose random seeds and deterministic modes, but the included literature provides little comparable evidence on seed control, deterministic execution, tolerance bands, or repeat-run variance.
These three recurring limitations constitute the core focus of the capability-by-capability comparison that follows. For each simulator capability, we examine three angles: what the yardstick of a traditional simulator is on that capability, what concrete efforts each world-model route has made (identifying representative works and their methods), and where the gap to a traditional simulator lies and what its essential cause is.

As profiled in Figure~\ref{fig:radar} of Section~\ref{sec:collection}, research attention is severely imbalanced across the eight capabilities. Measured by the cumulative number of papers for which a capability is a principal contribution, controllability C4 alone has reached 125 papers, interaction C3 has reached 80, and stability C5 has reached 80, whereas asset construction C1, physics engine C2, and state feedback C6 have long stalled in the 35--45 range. This gap is no accidental research vacuum but exactly the three foundational capabilities a world model most lacks as a simulator: the construction of the static world, adherence to physical law, and the feedback of structured state.

The analysis uses two complementary evidence views. Table~\ref{tab:checklist} lists representative systems that accept an action or conditional input and roll forward to predict future observations or state, together with their principal-contribution topics. Table~\ref{tab:matrix} aggregates all 163 implementation papers by six model families. Benchmarks, surveys, and the three other contextual records are excluded from these family denominators.
\begin{table*}[p]
\centering
\scriptsize
\caption{\textbf{Representative simulator-capable world models, grouped by their primary simulator capability.} Each work is marked on every capability it substantively contributes (C1--C8), with a \emph{Category} following the taxonomy of Section~\ref{sec:comparison} and a one-line \emph{Highlight}. For C8 we list representative evaluation benchmarks. Routes: Latent (latent dynamics), AR (autoregressive), Diff.\ (diffusion), 3D/4D, Occ.\ (occupancy), Bench.\ (benchmark); \emph{arXiv} venues are preprints as of the search cutoff.}
\label{tab:checklist}
\resizebox{\linewidth}{!}{%
\begin{tabular}{@{}ll l *{8}{c} l l@{}}
\toprule
\textbf{Method} & \textbf{Venue} & \textbf{Route} & \textbf{C1} & \textbf{C2} & \textbf{C3} & \textbf{C4} & \textbf{C5} & \textbf{C6} & \textbf{C7} & \textbf{C8} & \textbf{Category} & \textbf{Highlight} \\
\midrule
\rowcolor{myblue}\multicolumn{13}{@{}l}{\textit{C1 -- Asset Construction}} \\
Lyra2~\cite{2604.13036} & arXiv'26 & 3D/4D & \cmark &  & \cmark &  & \cmark &  &  &  & Asset generation & Memory routing removes drift \\
DecMem~\cite{2605.31336} & arXiv'26 & Diff. & \cmark &  &  &  & \cmark &  &  &  & Asset generation & Decoupled memory, minute-scale \\
CP4D~\cite{2606.09187} & arXiv'26 & 3D/4D & \cmark & \cmark &  &  & \cmark &  &  &  & Asset generation & Physics-aware compositional 4D \\
Prisma-World~\cite{2606.09507} & arXiv'26 & Diff. & \cmark &  &  & \cmark & \cmark &  &  &  & Asset as condition & Multi-agent cross-view consistency \\
MoVerse~\cite{2606.13376} & arXiv'26 & 3D/4D & \cmark &  & \cmark &  & \cmark &  &  &  & Asset generation & Panorama-Gaussian navigable world \\
Cosmos~\cite{2501.03575} & arXiv'25 & Diff. & \cmark &  &  &  &  &  & \cmark &  & Unified platform & Fine-tunable foundation platform \\
GWM~\cite{2508.17600} & arXiv'25 & 3D/4D & \cmark & \cmark &  & \cmark &  &  &  &  & Asset generation & Gaussian primitives as state \\
Aether~\cite{2503.18945} & arXiv'25 & 3D/4D & \cmark &  &  & \cmark &  &  & \cmark &  & Asset generation & Unified 4D reconstruct+predict+plan \\
WonderWorld~\cite{2406.09394} & arXiv'24 & 3D/4D & \cmark &  & \cmark &  & \cmark &  &  &  & Asset generation & Single image, $<10\,\mathrm{s}$ 3D \\
UniScene~\cite{2412.05435} & arXiv'24 & Occ. & \cmark &  &  &  &  &  & \cmark & \cmark & Asset generation & Occupancy-centric multi-sensor generation \\
\midrule
\rowcolor{myblue}\multicolumn{13}{@{}l}{\textit{C2 -- Physics Engine}} \\
RealWonder (PhysBridge)~\cite{2603.05449} & arXiv'26 & Diff. &  & \cmark & \cmark &  &  &  &  &  & Simulator bridge & Simulator flow, 4-step refine \\
OrbiSim~\cite{2605.16395} & arXiv'26 & Latent &  & \cmark &  & \cmark &  & \cmark &  &  & Hard physics & Differentiable physics core \\
RoboWM-Bench~\cite{2604.19092} & arXiv'26 & Bench. &  & \cmark &  & \cmark &  &  &  & \cmark & Physical eval & Generation-to-execution verification \\
What-If World~\cite{2605.27589} & arXiv'26 & Bench. &  & \cmark &  &  & \cmark &  &  & \cmark & Physical eval & Causal-intervention physics benchmark \\
ActWorld~\cite{2606.17730} & arXiv'26 & Diff. &  & \cmark &  & \cmark & \cmark &  &  &  & Soft constraint & Hierarchical action-aware memory \\
PIN-WM~\cite{2504.16693} & RSS'25 & 3D/4D & \cmark & \cmark &  &  &  &  & \cmark &  & Hard physics & Physics-parameter identification \\
PEVA~\cite{2506.21552} & NeurIPS'25 & Latent &  & \cmark &  & \cmark &  &  &  & \cmark & Kinematic prior & Whole-body pose conditioning \\
WorldVLA~\cite{2506.21539} & arXiv'25 & AR &  & \cmark &  & \cmark & \cmark &  &  &  & Soft constraint & Unifies VLA and WM \\
RoboScape~\cite{2506.23135} & arXiv'25 & Diff. & \cmark & \cmark &  &  &  & \cmark &  &  & Soft constraint & Depth + keypoint dynamics \\
ReconDreamer-RL~\cite{2508.08170} & arXiv'25 & Diff. &  & \cmark & \cmark &  &  &  &  & \cmark & Simulator bridge & Kinematics + diffusion for RL \\
\midrule
\rowcolor{myblue}\multicolumn{13}{@{}l}{\textit{C3 -- Interaction}} \\
Matrix-Game3~\cite{2604.08995} & arXiv'26 & AR &  &  & \cmark & \cmark & \cmark &  &  &  & Camera-level & 720p 40FPS streaming \\
PointWorld~\cite{2601.03782} & arXiv'26 & 3D/4D &  & \cmark & \cmark &  &  & \cmark &  &  & Algorithm-in-loop & 3D point flow, MPC-ready \\
DreamDojo~\cite{2602.06949} & arXiv'26 & Diff. &  &  & \cmark & \cmark &  &  & \cmark &  & Object-level & 44k-hour human-video latent action \\
WorldCraft~\cite{2605.25077} & arXiv'26 & AR &  &  & \cmark & \cmark & \cmark &  &  &  & Object-level & Object-level trajectory control \\
GameFactory~\cite{2501.08325} & ICCV'25 & AR &  &  & \cmark & \cmark &  &  & \cmark &  & Object-level & Scene-generalizing key/mouse control \\
DIAMOND~\cite{2405.12399} & NeurIPS'24 & Diff. &  &  & \cmark &  & \cmark & \cmark &  &  & Algorithm-in-loop & Diffusion WM trains RL agent \\
GameNGen~\cite{2408.14837} & ICLR'24 & Diff. &  &  & \cmark & \cmark & \cmark &  &  &  & Camera-level & Real-time neural game engine \\
GenAD~\cite{2403.09630} & arXiv'24 & AR &  &  & \cmark & \cmark &  &  & \cmark &  & Camera-level & First large driving predictor \\
UniSim~\cite{2310.06114} & ICLR'23 & Diff. &  &  & \cmark & \cmark &  &  & \cmark &  & Object-level & Universal interaction simulator \\
DayDreamer~\cite{2206.14176} & CoRL'22 & Latent &  & \cmark & \cmark &  &  &  &  &  & Algorithm-in-loop & Online learning on robot \\
\midrule
\rowcolor{myblue}\multicolumn{13}{@{}l}{\textit{C4 -- Controllability}} \\
DiLA~\cite{2605.15725} & arXiv'26 & Latent &  &  &  & \cmark & \cmark &  & \cmark &  & Latent action & Content-structure decoupling \\
UWM-JEPA~\cite{2605.25313} & arXiv'26 & JEPA &  &  &  & \cmark & \cmark &  &  &  & Latent action & Density-matrix belief space \\
DisCo~\cite{2606.07967} & arXiv'26 & AR &  &  &  & \cmark & \cmark &  &  &  & Explicit & Discrete motion primitives \\
AdaWorld~\cite{2503.18938} & ICML'25 & AR &  &  & \cmark & \cmark &  &  &  &  & Latent action & Transferable latent actions \\
GAIA-2~\cite{2503.20523} & arXiv'25 & Diff. &  &  &  & \cmark & \cmark &  & \cmark &  & Explicit & Structured multi-view driving control \\
V-JEPA2~\cite{2506.09985} & arXiv'25 & JEPA &  &  &  & \cmark & \cmark &  & \cmark &  & Latent action & Self-supervised, 15x planning \\
Genie~\cite{2402.15391} & ICML'24 & Latent & \cmark &  & \cmark & \cmark &  &  &  &  & Latent action & Unsupervised latent action \\
Pandora~\cite{2406.09455} & arXiv'24 & AR &  &  & \cmark & \cmark &  &  & \cmark &  & Language/multimodal & Anytime free-text control \\
GameGen-X~\cite{2411.00769} & ICLR'24 & Diff. & \cmark &  &  & \cmark &  &  & \cmark &  & Language/multimodal & First open-world game DiT \\
RoboDreamer~\cite{2404.12377} & ICML'24 & Diff. &  &  &  & \cmark & \cmark &  & \cmark &  & Language/multimodal & Compositional text-to-video \\
\midrule
\rowcolor{myblue}\multicolumn{13}{@{}l}{\textit{C5 -- Stability}} \\
OpenWM~\cite{2601.20540} & arXiv'26 & AR &  &  & \cmark &  & \cmark &  & \cmark &  & Implicit memory & Minute-scale open-source WM \\
FastVDM~\cite{2602.01801} & arXiv'26 & AR &  &  & \cmark &  & \cmark &  &  &  & Drift suppression & Training-free attention acceleration \\
LiveWorld~\cite{2603.07145} & arXiv'26 & 3D/4D &  &  &  &  & \cmark &  &  & \cmark & Causal consistency & Off-sight dynamics synchronized \\
SWM~\cite{2603.15583} & arXiv'26 & AR &  &  &  & \cmark & \cmark &  &  &  & Drift suppression & Retrieval-anchored to real city \\
VectorWorld~\cite{2603.17652} & arXiv'26 & Diff. &  &  & \cmark &  & \cmark & \cmark &  &  & Drift suppression & Vector-map streaming diffusion \\
LT-SpatialMem~\cite{2506.05284} & arXiv'25 & AR & \cmark &  &  &  & \cmark &  &  &  & Explicit 3D memory & Geometry-grounded spatial memory \\
RELIC~\cite{2512.04040} & arXiv'25 & AR &  &  & \cmark & \cmark & \cmark &  &  &  & Implicit memory & KV-cache camera memory \\
GenWorldExplorer~\cite{2411.11844} & arXiv'24 & Diff. &  &  & \cmark &  & \cmark & \cmark &  &  & Explicit 3D memory & Belief update via exploration \\
IRIS~\cite{2209.00588} & ICLR'22 & AR &  &  &  &  & \cmark & \cmark &  &  & Implicit memory & Discrete AR Transformer WM \\
World Models~\cite{1803.10122} & arXiv'18 & Latent &  &  & \cmark &  & \cmark &  &  &  & Implicit memory & VAE+RNN, train in dream \\
\midrule
\rowcolor{myblue}\multicolumn{13}{@{}l}{\textit{C6 -- State Feedback}} \\
Cosmos Policy~\cite{2601.16163} & arXiv'26 & Diff. &  &  & \cmark & \cmark &  & \cmark &  &  & Latent feedback & Latent feedback into policy \\
TesserAct~\cite{2504.20995} & ICCV'25 & 3D/4D & \cmark &  &  &  &  & \cmark &  &  & Multi-task output & RGB-depth-normal 4D output \\
ChronoDreamer~\cite{2512.18619} & arXiv'25 & AR &  & \cmark & \cmark &  &  & \cmark &  &  & Physics-derived & Explicit contact-force prediction \\
OccSora~\cite{2405.20337} & TIP'24 & Occ. &  &  &  & \cmark & \cmark & \cmark &  &  & Multi-task output & Diffusion 4D-occupancy simulator \\
OccLLaMA~\cite{2409.03272} & arXiv'24 & Occ. &  &  & \cmark & \cmark &  & \cmark &  &  & Multi-task output & Occupancy-language-action vocab \\
DreamerV2~\cite{2010.02193} & ICLR'20 & Latent &  &  &  &  & \cmark & \cmark &  &  & Reward/termination & Discrete latent, human-level \\
SimPLe~\cite{1903.00374} & ICLR'19 & AR &  &  &  &  & \cmark & \cmark &  &  & Reward/termination & Sample-efficient Atari RL \\
MuZero~\cite{1911.08265} & Nature'19 & Latent &  &  &  &  & \cmark & \cmark &  &  & Reward/termination & Latent model + tree search \\
Dreamer~\cite{1912.01603} & ICLR'19 & Latent &  &  &  &  & \cmark & \cmark &  &  & Reward/termination & Value gradients in imagination \\
PlaNet~\cite{1811.04551} & ICML'18 & Latent &  &  &  &  & \cmark & \cmark &  &  & Reward/termination & Latent overshooting planning \\
\midrule
\rowcolor{myblue}\multicolumn{13}{@{}l}{\textit{C7 -- Diversity}} \\
Cosmos-Drive~\cite{2506.09042} & arXiv'25 & Diff. &  &  &  & \cmark &  &  & \cmark &  & Composition & Synthetic long-tail driving \\
GigaWorld-0~\cite{2511.19861} & arXiv'25 & Diff. & \cmark & \cmark &  &  &  &  & \cmark &  & Parametric & Controllable data-engine diversity \\
AccidentGen~\cite{end2025} & arXiv'25 & Diff. &  &  &  & \cmark &  &  & \cmark &  & Composition & Rare accident scene generation \\
DreamerV3~\cite{2301.04104} & Nature'23 & Latent &  &  &  &  & \cmark &  & \cmark &  & Scale-driven & One config, 150+ domains \\
\midrule
\rowcolor{myblue}\multicolumn{13}{@{}l}{\textit{C8 -- Evaluation Metrics}} \\
DrivingGen~\cite{2601.01528} & arXiv'26 & Bench. &  &  &  & \cmark &  &  & \cmark & \cmark & Unified quality & Driving WM benchmark \\
WorldBench~\cite{2601.21282} & arXiv'26 & Bench. &  & \cmark &  &  &  &  &  & \cmark & Physical plausibility & Single-concept physics diagnosis \\
Omni-WorldBench~\cite{2603.22212} & arXiv'26 & Bench. &  &  & \cmark &  &  &  &  & \cmark & Closed-loop embodied & Interaction-centric 4D benchmark \\
GameWorld~\cite{2604.07429} & arXiv'26 & Bench. &  &  & \cmark &  &  &  &  & \cmark & Closed-loop embodied & Verifiable game-agent benchmark \\
OpenGame~\cite{2604.18394} & arXiv'26 & Other &  &  &  &  &  &  &  & \cmark & Generation quality & Agentic playable-game generation \\
WBench~\cite{2605.25874} & arXiv'26 & Bench. &  &  & \cmark &  &  &  &  & \cmark & Closed-loop embodied & Multi-turn interaction benchmark \\
WorldScore~\cite{2504.00983} & arXiv'25 & Bench. &  &  &  & \cmark &  &  &  & \cmark & Unified quality & First unified WM benchmark \\
World-in-World~\cite{2510.18135} & arXiv'25 & Bench. &  &  & \cmark & \cmark &  &  &  & \cmark & Closed-loop embodied & Visual quality != task success \\
Genie-Envisioner~\cite{2508.05635} & arXiv'25 & Diff. &  & \cmark &  & \cmark &  &  &  & \cmark & Unified quality & Unified policy/eval/sim platform \\
1X-WorldModel~\cite{2510.07092} & arXiv'25 & Bench. &  &  &  & \cmark &  & \cmark &  & \cmark & Closed-loop embodied & Humanoid interaction benchmark \\
\bottomrule
\end{tabular}}
\end{table*}

\begin{table*}[!t]
\centering
\caption{\textbf{Model-family $\times$ principal-topic counts for the 163 implementation papers.} Columns denote C1 asset construction, C2 physics engine, C3 interaction, C4 controllability, C5 stability, C6 state feedback, C7 diversity, and C8 evaluation metrics. Each cell reports $n/N$, where $N$ is the complete family denominator and $n$ is the number carrying that principal-topic label. Counts measure research attention, not capability attainment.}
\label{tab:matrix}
\small
\setlength{\tabcolsep}{4pt}
\begin{tabular}{@{}l cccc cccc@{}}
\toprule
Route & C1 & C2 & C3 & C4 & C5 & C6 & C7 & C8 \\
\midrule
Latent dynamics/action & 2/29 & 5/29 & 11/29 & 18/29 & 11/29 & 10/29 & 9/29 & 3/29 \\
Autoregressive & 2/37 & 3/37 & 21/37 & 25/37 & 23/37 & 7/37 & 8/37 & 7/37 \\
Diffusion & 12/63 & 11/63 & 26/63 & 52/63 & 27/63 & 13/63 & 14/63 & 9/63 \\
JEPA predictive & 0/5 & 0/5 & 1/5 & 5/5 & 4/5 & 0/5 & 1/5 & 0/5 \\
Explicit 3D/4D & 20/25 & 5/25 & 6/25 & 11/25 & 8/25 & 6/25 & 7/25 & 1/25 \\
Occupancy & 1/4 & 0/4 & 1/4 & 3/4 & 1/4 & 3/4 & 1/4 & 1/4 \\
\bottomrule
\end{tabular}
\vspace{5pt}

\begin{minipage}{.86\textwidth}
\footnotesize
\textit{Reading guide.} Controllability is the most frequently coded topic in five families. Explicit 3D/4D papers concentrate on asset construction, and three of four occupancy papers discuss state feedback. These distributions motivate route-specific questions, but the small JEPA and occupancy denominators and the absence of a shared performance threshold preclude family maturity rankings.
\end{minipage}
\end{table*}

\subsection{Asset Construction}\label{sec:c1}

Asset construction is a foundational capability of a traditional simulator, encompassing the creation and management of static elements such as maps, characters, objects, materials, and environmental lighting. In a traditional production pipeline, this process depends heavily on human artists and modelling experts; quality is controllable and instances are decomposable and editable, but the cost is high and generalization is hard. Of the 200 papers, only 38 (19.0\%) list asset construction as a principal contribution dimension---a low proportion worth noting: without high-quality, editable static assets as a foundation, the subsequent dynamic simulation lacks a reliable carrier.

We organize the analysis of asset construction around one core distinction: a world model can accept assets as a conditional input for generation---given a scene layout, make it move; it can also generate assets as its own output---creating an explorable world by itself; and it can serve as a unified platform integrating both. These three perspectives correspond to different maturity stages of the capability, from using known assets, to creating new assets, to integrating asset construction with dynamic simulation.

\subsubsection{Assets as Conditional Input}

Many world models use explicit 2D or 3D geometric information as a generation condition; such geometry is essentially the assets of a traditional simulator, including road layouts, building geometry, object positions, and robot structure. In driving scenes this conditioning is especially natural. MAGE \cite{2310.02601} injects BEV layout, camera parameters, and 3D bounding boxes as multiple geometric conditions into a street-view generation diffusion model, where the bird's-eye view defines road topology and lane structure, the 3D boxes fix the spatial positions of traffic participants, and cross-view attention guarantees geometric consistency across front, rear, and surround views, so that the generated street scene strictly obeys the specified geometric constraints rather than freely generating scenes inconsistent with the road structure. DriveDreamer \cite{2309.09777} precisely generates driving videos that conform to traffic-structure constraints via a controllable diffusion model, incorporating road topology and lane lines as hard geometric constraints; its key value is to guarantee structurally reasonable content, i.e., vehicles driving in lanes rather than floating in the air. DriveDreamer-2 \cite{2403.06845} further lengthens the asset-conditioning chain: an LLM first converts a user's language query (e.g., generate a congestion scene on a multi-lane highway) into agent trajectories, then generates an HDMap conforming to traffic rules from the trajectories, and finally produces a spatio-temporally consistent driving video via a unified multi-view model. This pipeline---from language to trajectory to HDMap to multi-view video---essentially introduces asset conditions at different abstraction levels at different stages of generation: the semantic layer where the LLM understands congestion, the geometric layer where the HDMap defines road structure, and the instance layer where trajectories define specific vehicle positions and time curves. Such layered conditioning lets users control content at the most appropriate abstraction level rather than being forced to intervene at the pixel level.

\begin{figure*}[!t]
\centering
\includegraphics[width=.82\textwidth]{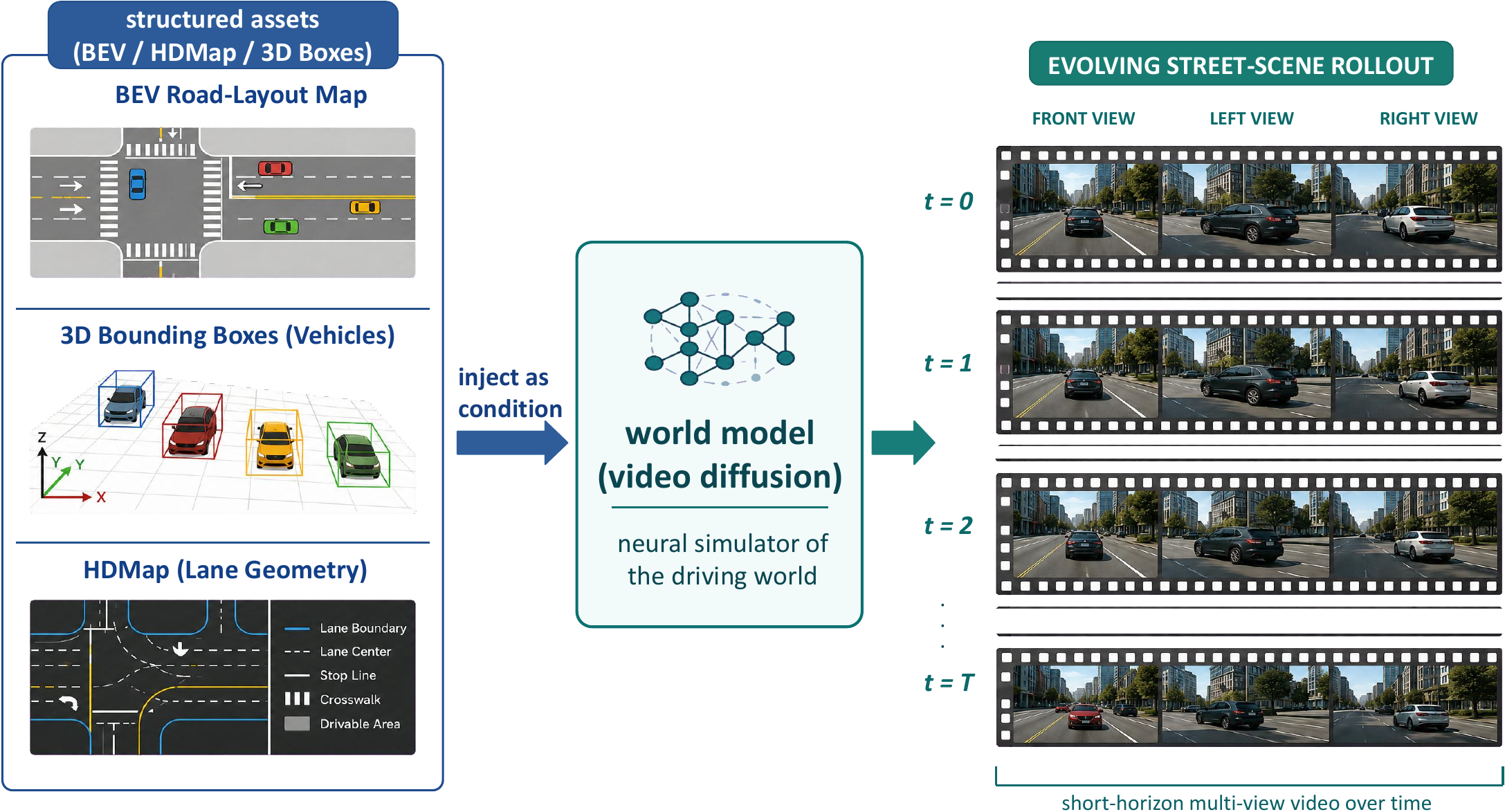}
\caption{\textbf{Structured assets as conditional input to a world model.} BEV road layouts, HD maps, and 3D bounding boxes specify the geometry and traffic configuration supplied to a video-diffusion world model, which produces a temporally evolving, multi-view street-scene rollout consistent with those conditions.}
\label{fig:asset-conditional}
\end{figure*}

Figure~\ref{fig:asset-conditional} schematizes this conditional-input route: structured geometry fixes the scene configuration, while the world model is responsible for rolling the conditioned scene forward over time.

In robotic scenes, asset conditions are even more structured, because the robot's own kinematic structure is known and precise. VerseCrafter \cite{2601.05138} proposes a 4D geometric-control representation: it encodes the scene background as a 3D point cloud (static asset), represents dynamic objects as 3D-Gaussian trajectories (spatio-temporal encoding of dynamic assets), and encodes occupancy as spatial constraints, jointly driving a video-diffusion model. The essence of this representation is to separate the geometric state of the world from its pixel projection---controlling object motion not on the 2D image plane but by precisely specifying, in the 3D world coordinate system, that an object is at $(x_1,y_1,z_1)$ at $t{=}1$ and $(x_2,y_2,z_2)$ at $t{=}2$, then rendering video from this 4D geometric state. This separation of geometric state from visual rendering is structurally isomorphic to the world-state-to-rendering-pipeline design of a traditional game engine. Kinema4D \cite{2603.16669} drives 4D embodied simulation conditioned on a URDF (Unified Robot Description Format) kinematic model, outputting RGB and pointmap as dual feedback; URDF is a standard asset format in robotics that describes joint structure, link lengths, mass distribution, and kinematic chains. Driving generation with a standardized robot-description format lets a world model plug directly into existing robotics software ecosystems, such as ROS \cite{quigley2009ros} and Gazebo \cite{koenig2004gazebo}, without users learning a new model-specific interface. The URDF embodied mask \cite{2602.03793} further solves the alignment between coordinate actions and pixel video: by introducing an embodied mask into the diffusion model's attention, it precisely maps actions in 3D coordinate space (e.g., move the end-effector to $(0.3, 0.1, 0.5)$) to visual changes in 2D pixel space, while unifying the control architecture across different robot embodiments.

Together, these works reveal a regularity: when a world model accepts explicit, structured assets as input conditions, the precision and controllability of generation improve markedly. This is because the model no longer needs to simultaneously guess what the world looks like and predict how it changes; given the world's appearance, it focuses on predicting the world's evolution, and the precision of the asset condition translates directly into the precision of generation control. In current practice, however, a critical gap remains between asset conditions and a queryable, editable world state: BEV layouts and 3D point clouds are injected as one-off conditions and cannot be queried or modified after generation, so users cannot edit the scene mid-generation (e.g., change the third building to blue), which is a basic operation of a traditional simulator.

\subsubsection{Asset Generation by World Models}

Complementary to accepting assets as conditions is the world model generating assets itself. On the explicit 3D/4D reconstruction route, several works attempt to generate navigable, explorable, geometrically consistent 3D worlds directly from extremely sparse input (a single image or monocular video). WonderWorld \cite{2406.09394} uses FLAGS (Fast Layered Gaussian Surfels) as its scene representation, generating an interactively explorable 3D scene from a single image in under 10 seconds and achieving geometrically consistent stitching via guided depth diffusion. Its key innovation is to avoid the expensive per-view optimization of traditional 3D reconstruction, using geometric initialization to drastically reduce optimization time and make real-time interactive 3D-world construction possible. Lyra~2.0 \cite{2604.13036} specifically targets 3D-consistency degradation under long-range camera trajectories---the key bottleneck of all autoregressive scene generation---and proposes a dual solution: spatial-memory routing (using the generated per-frame 3D geometry to retrieve relevant historical frames and establish dense correspondences) to counter spatial forgetting, and self-augmented history training (exposing the model to its own imperfect outputs) to counter temporal drift; together they greatly extend the scale of explorable 3D worlds and support geometric consistency upon revisiting a location. MoVerse \cite{2606.13376} adopts a build-the-world-then-render-observations three-stage strategy: the first stage uses topology-aware diffusion to expand a single narrow-FoV image into a 360$^\circ$ panorama to bridge the missing field of view, the second lifts the panorama into a persistent 3D-Gaussian scaffold to establish explicit spatial memory, and the third generates a real-time roaming video stream along a user-specified trajectory via a Gaussian-conditioned video renderer (about 8 FPS on an RTX 4090). Separating world construction and observation rendering in time---completing geometric reconstruction before visual rendering---is architecturally closer to the design paradigm of a traditional game engine than the generate-while-rendering pure-autoregressive scheme.

\begin{figure*}[!t]
\centering
\includegraphics[width=.82\textwidth]{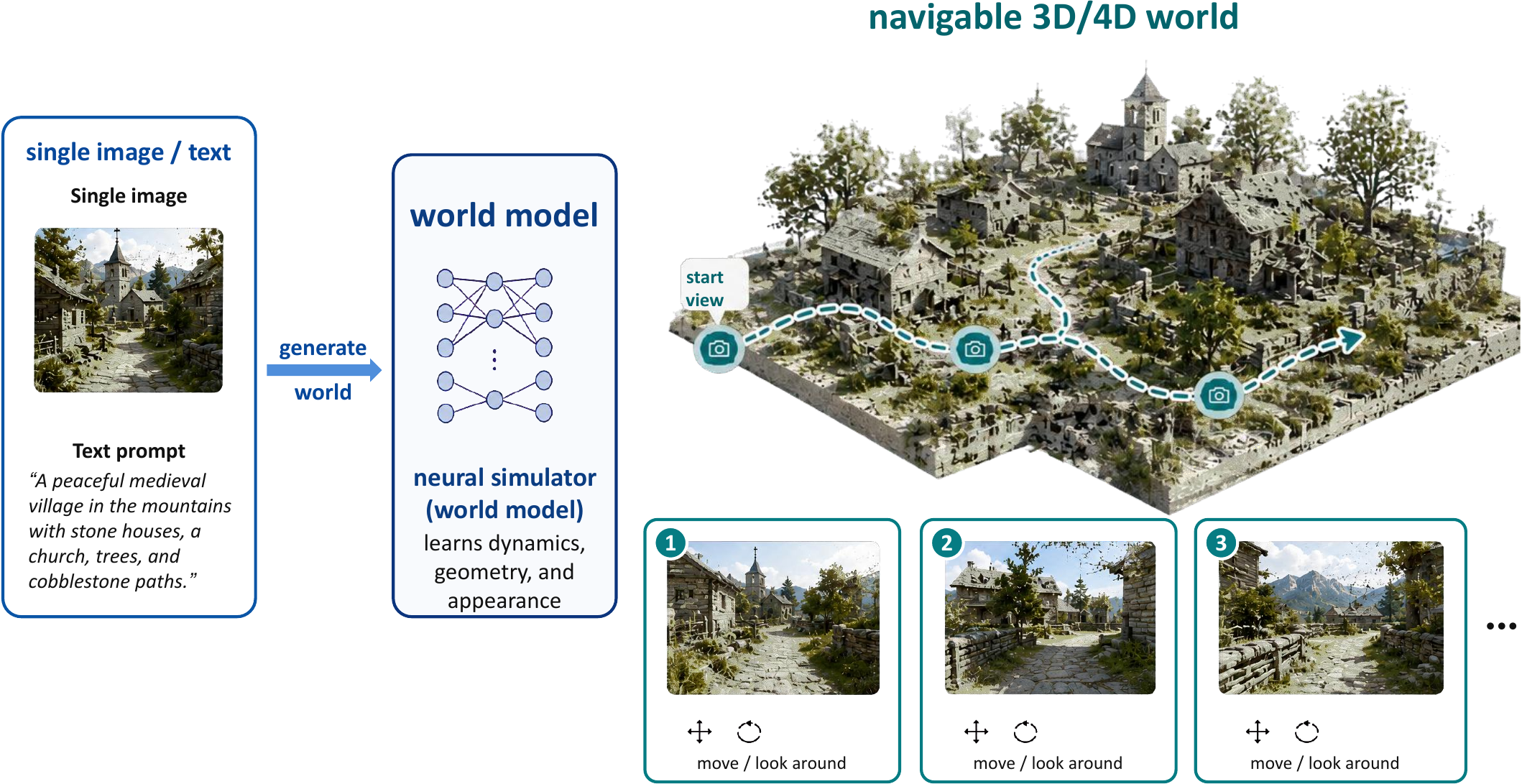}
\caption{\textbf{Generation of a navigable 3D/4D world from sparse input.} A single image or text description conditions a world model that constructs an environment with spatial extent and renders consistent observations along a user-controlled camera path.}
\label{fig:asset-generation}
\end{figure*}

The sparse-input-to-world process is illustrated in Figure~\ref{fig:asset-generation}: the output must support viewpoint changes along a navigation trajectory rather than merely depict a single generated object or frame.

In 4D-scene (dynamic 3D) generation, a body of work explores synthesizing explicit, spatio-temporally consistent 4D assets directly from sparse input. Free4D \cite{free4d2025} generates spatio-temporally consistent 4D-Gaussian scenes without per-scene tuning, lifting a single image or text prompt into a freely-navigable dynamic world; HoloTime \cite{holotime2025} tames a video-diffusion model to generate panoramic 4D Gaussians, supporting 360$^\circ$ roaming within the generated scene and extending asset generation from narrow FoV to panorama. Comp4D \cite{comp4d2026} targets multi-object interaction, a difficulty of 4D generation: an LLM first decomposes the scene into objects and their motion trajectories, then compositional score distillation jointly optimizes inter-object interaction so that the generated dynamic scene is more than a simple superposition of independent objects. A unified text-and-image-guided 4D scene generator \cite{approach2024} supports both text and image conditions driving D-NeRF-style dynamic-scene synthesis under one input interface, widening the entry points for 4D assets; a monocular-video route toward real data \cite{dynamic2026} reconstructs an editable, interactive dynamic world from a single monocular video, directly answering the pain point that generated assets are hard to edit afterwards. These works are of a piece with CP4D \cite{2606.09187}, which decomposes a 4D scene into two subproblems---static 3D environment background and dynamic foreground objects. The significance of this decomposition exceeds engineering convenience; it reflects a structural understanding of the world: changes usually involve only a few objects, while most environment elements are static. CP4D combines the motion constraints of a physics simulator with the commonsense-motion prior of a video-diffusion model for hybrid motion synthesis, generating physically plausible trajectories while retaining visual richness. FR3D \cite{2606.18250} approaches from another angle, explicitly decoupling ego-motion from scene dynamics. In traditional video generation, camera motion and scene motion are mixed on the image plane, breaking geometric consistency because it is hard to judge whether an object's displacement arises from camera or object motion. FR3D treats ego-motion as a latent proxy action and, via teacher--student distillation, extracts spatial commonsense from a foundation model to achieve future-2-second dynamic 3D reconstruction without action annotation, with zero-shot generalization. Aether \cite{2503.18945}, through task-interleaved feature learning, unifies 4D dynamic reconstruction, action-conditioned video prediction, and goal-conditioned visual planning in one framework, where the three tasks share the same underlying geometric representation and mutually reinforce; its zero-shot sim-to-real generalization---reconstruction performance rivalling or surpassing domain-specific models even without seen real data---demonstrates the feasibility of geometric understanding as a transferable foundational capability.

The most noteworthy new paradigm in this direction is to simulate the world directly in 3D asset space rather than simulating in 2D pixel space and then reconstructing 3D. GWM (Gaussian World Model) \cite{2508.17600} takes 3D-Gaussian primitives---rather than pixels or latent features---as the world model's primitive state representation; a DiT-based neural simulator propagates these primitives under actions to reconstruct future 3D scenes, thereby supporting model-based RL policy training in native 3D space. Its advantages are built-in geometric consistency (3D Gaussians natively carry spatial coordinates), physical interaction modelled in 3D space to avoid 2D-projection ambiguity, and output directly usable for the 3D positions, orientations, and occupancy needed by downstream robotic manipulation. The Gaussian Action Field \cite{2506.14135} further couples action and asset on the same 3D-Gaussian representation: it drives the deformation of 3D Gaussians directly by gripper pose, defining robot action as an explicit spatial transformation of the Gaussian field, so that the evolution of the 4D dynamic world representation and the manipulation action align in the same geometric space rather than being conveyed indirectly through pixels or latent vectors. PlayerOne \cite{2506.09995} represents another path---reconstructing the world from a first-person view: given a user's egocentric scene image, it builds a corresponding explorable world through a coarse-to-fine training pipeline (large-scale pretraining plus scene-specific fine-tuning) and generates egocentric video strictly aligned with the user's real motion, its 4D reconstruction guaranteeing scene consistency across long video sequences. GigaWorld-0 \cite{2511.19861} positions the world model as a complete data engine for embodied AI, where GigaWorld-0-Video generates diverse, texture-rich embodied video sequences under fine-grained control of appearance, camera view, and scene layout, and GigaWorld-0-3DGS generates 3D scenes in a physically plausible way; together they provide controllable, diverse training data for VLA learning, elevating asset generation from visual display to the level of downstream function. In driving, the Xiaomi Auto World Model \cite{2605.18137} unifies reconstruction and generation in one industrial-grade framework: the reconstruction route yields high-fidelity, editable static-scene assets, while the generation route fills in rare operating conditions, demonstrating the fusion of asset construction and dynamic simulation in a real production line. WorldGen \cite{2511.16825} organizes scene elements (terrain, buildings, vegetation, roads) in an editable layered structure, supporting instance-level control of generated assets---an important advance on the key dimension of editability of generated assets. MagiCity4D \cite{magicity4d2026} advances the same goal at city scale: an MLLM agent parses language and layout sketches to drive procedural content generation (PCG), producing real-time-editable 4D city scenes and letting users add, remove, or adjust buildings after generation, directly answering the earlier-noted gap that asset conditions cannot be queried or edited once injected. Playable Environments \cite{2203.01914} pioneeringly learns interactive 3D scenes from monocular video, learning a per-frame environment-state representation manipulated by an action module and rendered by a style-modulated NeRF, demonstrating the feasibility of inversely inferring an interactive 3D environment from video observation. GameGen-X \cite{2411.00769}, the first DiT designed for open-world game-video generation and interactive control, unifies text, keyboard/mouse, and visual prompts via InstructNet to interactively control new characters, dynamic environments, complex actions, and diverse events. OpenGame \cite{2604.18394} approaches from the entirely different angle of code generation, letting an LLM agent generate a playable web game via a template skill library and a debugging protocol library---demonstrating another possibility for asset construction: not generating visual assets, but generating the code that creates assets.

\begin{figure*}[!t]
\centering
\includegraphics[width=.82\textwidth]{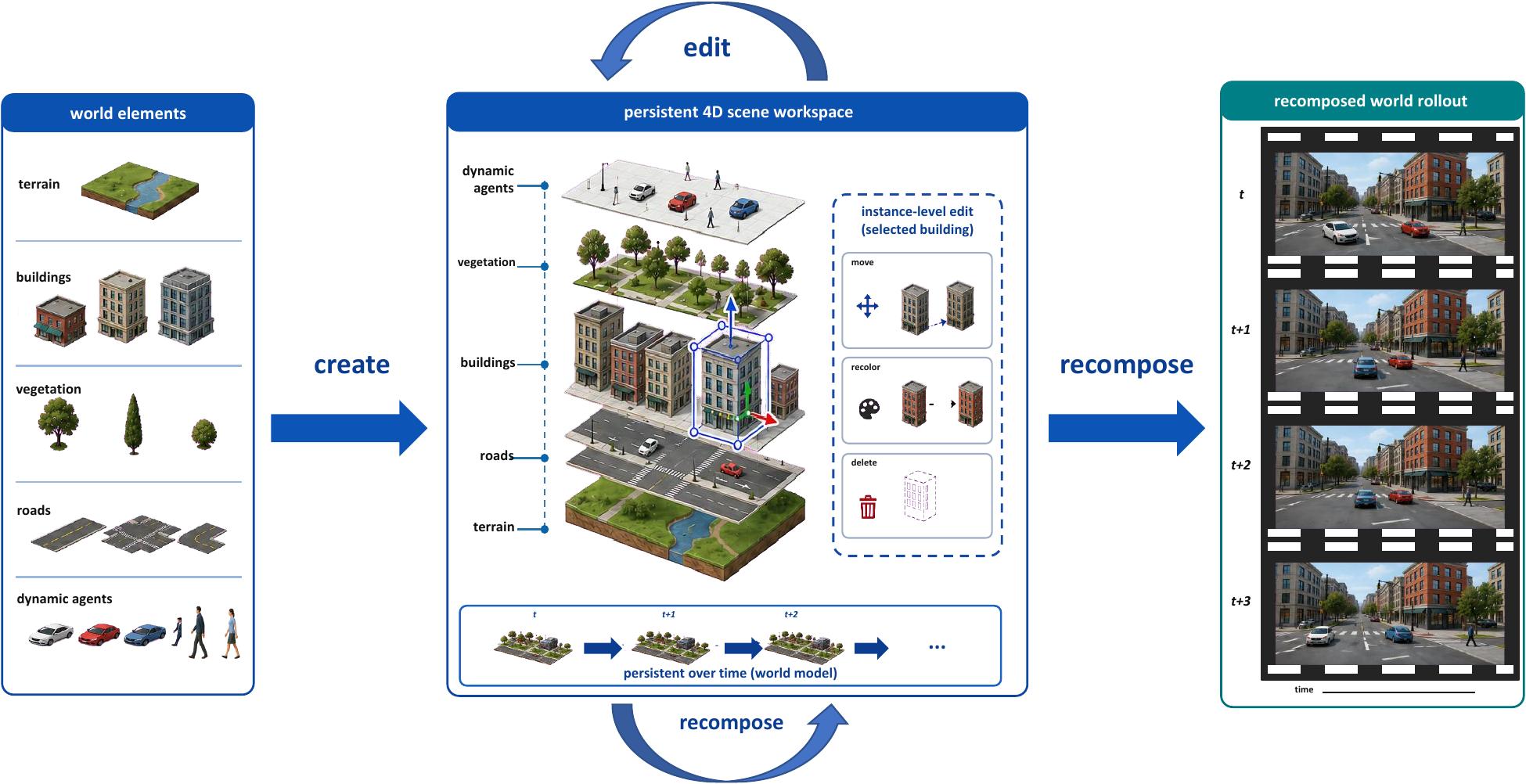}
\caption{\textbf{A unified platform for persistent and editable world assets.} Terrain, buildings, vegetation, roads, and dynamic agents are maintained as separable layers; instance-level edits update the persistent scene before the world model generates a recomposed rollout.}
\label{fig:asset-platform}
\end{figure*}

Figure~\ref{fig:asset-platform} shows the stronger platform-level requirement implied by these systems: assets should remain decomposed, persistent, and editable while subsequent world evolution is recomposed from the modified scene.

The gap in asset construction is essentially the chasm between implicit and explicit, operable representations. Almost all 3D scenes generated by world models are stored implicitly or semi-implicitly---as NeRF weights \cite{mildenhall2020nerf}, 3D-Gaussian sets \cite{kerbl2023gaussians}, MLP latent features, or even the diffusion-model parameters themselves---lacking the instance-level decomposability that downstream editing tools require. Users cannot individually select, edit, or replace an object as in Unreal Engine \cite{epicgames2026unreal}, cannot export to standard mesh, material, and collision formats, and cannot modify scene parameters mid-generation. This chasm is not merely an engineering format-conversion problem; it reflects a core design trade-off: current generative models take end-to-end differentiability as the highest principle, whereas editability inherently requires modularity and interruptibility. How to introduce a structured, editable intermediate representation while preserving generation quality---even at the cost of partially sacrificing end-to-end differentiability---is key to a breakthrough in asset construction.

\subsection{Physics Engine}\label{sec:c2}

The physics engine is the core of a traditional simulator, precisely and deterministically simulating the motion and interaction of objects under forces through numerical solvers. Formal guarantees such as energy and momentum conservation, collision impenetrability, and friction-cone constraints form the unshakeable cornerstone of scientific computing and safety-critical simulation. The physical-law simulation capability of world models is one of the most core---and most worrying---dimensions for assessing whether they approach a rigorous simulator. Of the 200 papers, only 34 (17.0\%) list the physics engine as a principal contribution dimension. Physical correctness is, in most current research, treated as post-processing, downstream verification, or future work, rather than a first-class design goal of the architecture.

The most natural framework for analysing the physics-engine capability is to proceed by the strictness of physical constraints, from strong to weak. The strictest is to build the world model itself as a differentiable physics engine (\emph{hard physics}); the intermediate level injects physical constraints as soft guidance into the generation process (\emph{soft constraints}); and the outermost only verifies physical plausibility post hoc through benchmark evaluation (\emph{physical evaluation}). These three levels correspond not only to decreasing physical strictness but also to a shift in view from physical guarantee to physical inspiration.

\subsubsection{Hard Physics}

\begin{figure*}[!t]
\centering
\includegraphics[width=.82\textwidth]{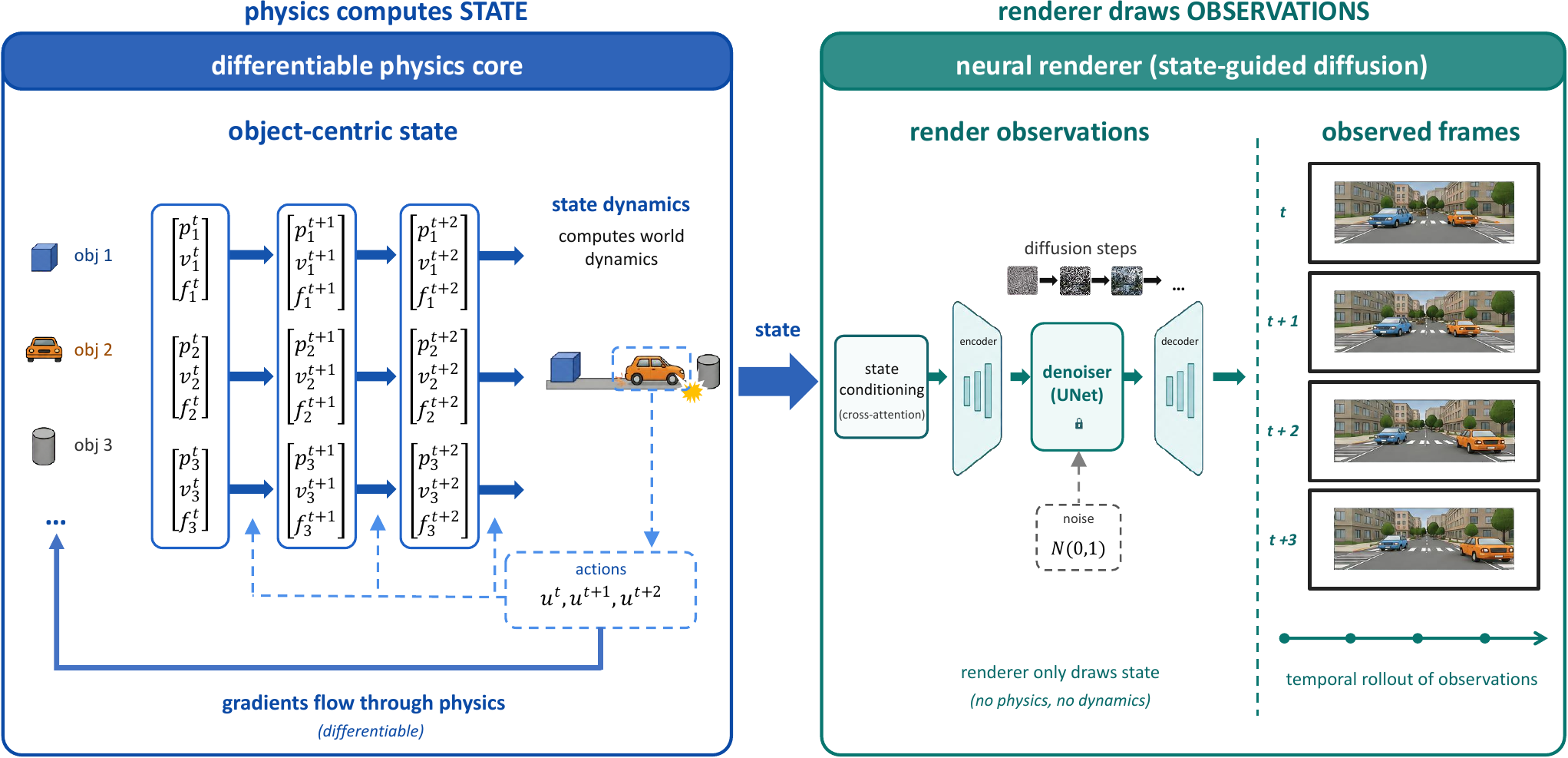}
\caption{\textbf{A world model with an explicit differentiable physics core.} Object-centric position, velocity, and force states are advanced under actions with gradients propagated through the dynamics; a state-guided neural renderer then converts the physical trajectory into observed frames.}
\label{fig:physics-hard}
\end{figure*}

Figure~\ref{fig:physics-hard} separates physical state evolution from observation rendering: the differentiable core computes the dynamics, and the neural renderer only visualizes the resulting state.

This is the most radical direction: not content to make the model \emph{look} as if it obeys physics, it makes the model itself take on the role of a physics engine. OrbiSim \cite{2605.16395} is representative, with an architecture of two cooperating modules. OrbiSim-Dynamics is an object-centric differentiable physics core that computes physical evolution (position, velocity, object configuration) in an explicit state space, where all intermediate physical quantities are accessible and differentiable, so gradients can propagate from downstream policy loss all the way through the physics simulation back to action and physical parameters. OrbiSim-Vision renders visual observations from the physical state via state-guided diffusion, where the diffusion model acts as a neural renderer rather than a world simulator, and physical correctness is guaranteed by the Dynamics module, not the Diffusion module. This differentiable-physics-core-plus-neural-renderer architecture strikes a delicate balance between physical fidelity (explicit physical state and verifiable physical constraints) and visual quality (the generative power of diffusion), while creating two abilities that traditional physics engines lack: differentiability (supporting gradient-based policy optimization and system identification) and physical-parameter inference (inferring mass, friction, elasticity, and other unobservable physical quantities from observations). In experiments, OrbiSim markedly surpasses existing world models in both prediction fidelity and control performance, and its sustained responsiveness to asset configuration and physical parameters indicates its potential as a differentiable tool supporting robotic simulation and policy training.

PIN-WM \cite{2504.16693} approaches from the perspective of physics-informed networks (PINNs), a variant of the hard-physics route: rather than building a full differentiable physics engine, it embeds physical equations as constraint terms in neural-network training. Its core design treats the physical parameters of 3D rigid-body dynamics (mass, friction coefficient, elastic modulus) as learnable variables and jointly optimizes two losses during training---the world model's prediction loss (how close the predicted next state is to the true next state) and the physical-equation residual (how much the predicted state trajectory violates the Newton--Euler equations). The significance is that, even without a full physics solver, using known physical equations as a soft regularizer rather than a hard constraint can pull model predictions toward physically reasonable regions. Its differentiable physical-identification capability, recovering environmental physical parameters from visual observations, has direct value in robotic scenes requiring sim-to-real transfer. Prompting-with-the-Future \cite{prompting2025} goes further still: rather than trying to teach a neural network physics, it directly builds an interactive digital twin, computing action consequences precisely with a URDF physics engine and having a VLM act as a judge to select actions in a closed-loop MPC. This essentially pushes the hard-physics route to the extreme---physical correctness is fully guaranteed by an external engine, and the world model recedes to providing semantic evaluation and visual priors, at the cost of the open-domain generalization of a pure generative route. The deformable-object manipulation world model \cite{enhanced2025} pushes hard-physics formal guarantees from rigid to soft bodies: it builds a perception-dynamics model with a Neural ODE, forms closed-loop control through perception gradients, and, for the first time for deformable-object manipulation, provides convergence criteria and a Lyapunov stability proof---stability guarantees exceedingly rare in the generative-world-model literature.

\subsubsection{Soft Constraints}

\begin{figure*}[!t]
\centering
\includegraphics[width=.82\textwidth]{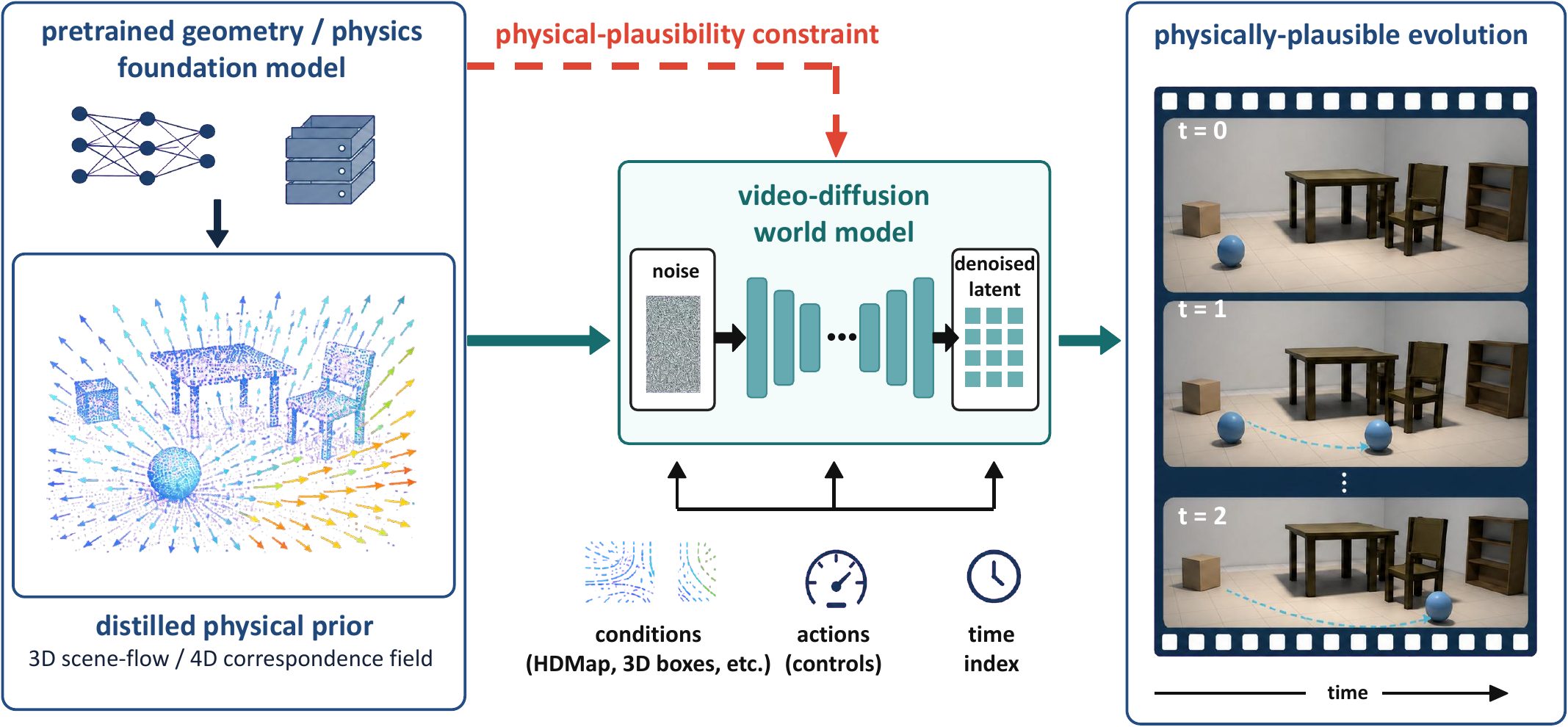}
\caption{\textbf{Soft physical constraints in a generative world model.} A pretrained geometry or physics model supplies a dense scene-flow or 4D-correspondence prior as a training constraint, while the conditioned video-diffusion model remains responsible for generating the evolving observations.}
\label{fig:physics-soft}
\end{figure*}

The alternative in Figure~\ref{fig:physics-soft} retains generative dynamics but constrains learning with distilled geometric or physical priors, improving plausibility without providing a formal physical guarantee.

Unlike building a physics engine from scratch, the vast majority of works inject physical constraints into an existing video-generation model---not changing the generative paradigm of the diffusion or autoregressive model, but adding physical guidance in the training objective or architecture. This sacrifices formal physical guarantees but retains the visual quality and open-domain capability of generative models.

GEM-4D \cite{2605.22882} represents an important methodology of physical soft constraints: rather than modelling physics from scratch, it distils geometric and physical knowledge from an existing pretrained model. Specifically, GEM-4D distils dense 4D-correspondence supervision---a per-pixel map of where each pixel should move in the next frame---from a pretrained geometric foundation model and explicitly injects this supervision into training the video-diffusion backbone. The elegance is that dense 4D correspondence encodes the scene's 3D geometric structure and motion, forcing the diffusion model to maintain geometric consistency while generating appearance. Its inverse-dynamics module further converts geometrically consistent video rollouts into executable robot trajectories, raising the success rate of real-world robotic manipulation from 61\% to 81\%. This 20-point gain is not merely a performance number; it directly quantifies the causal chain from geometric consistency to physical usability. FlowDreamer \cite{flowdreamer2026} shares the idea but goes further, embedding 3D scene flow itself as an explicit motion representation into an RGB-D world model: the model first predicts a 3D displacement field for each point in the scene, then generates the next-frame observation accordingly, making the physical continuity of motion an intrinsic constraint of the generation process rather than a post-hoc check. Mediating through explicit geometric motion is closer to the essence of physical evolution than directly regularizing in pixel space.

PEVA \cite{2506.21552} adopts another physical-soft-constraint strategy---structured kinematics as a prior. PEVA uses whole-body 3D joint pose (a sequence of human 3D skeletons) rather than an abstract latent vector as the action condition. The human skeleton is itself a physically constrained system: joint rotations are limited by anatomical ranges (an elbow cannot bend backward), and limb motion obeys kinematic-chain constraints (the hand's position is determined by the chained rotation of shoulder, elbow, and wrist); these constraints are already implicitly encoded in the skeleton structure, requiring no inference from pixels. An autoregressive conditional-diffusion model trained on the large-scale egocentric real dataset Nymeria can therefore learn how physical human motion shapes the first-person visual environment---e.g., a hand pushing a door occludes the view, and the view-sway frequency while walking matches the step frequency---rather than merely the statistical correlation of which pixels change given an action label. Using known physical structure as an architectural prior---not making the model cleverer at learning physics from data, but explicitly encoding physical structure into the model's input representation---is an important methodological direction for improving physical capability.

ChronoDreamer \cite{2512.18619} focuses on a specific but critical physical quantity: contact force. In precise robotic-manipulation tasks such as assembly, insertion, grinding, and wiping, precise perception and control of contact force is central to success, yet from RGB pixels alone it is nearly impossible to infer precisely whether the fingertip applies 3.2 N or 4.1 N. ChronoDreamer trains contact force as an explicit prediction target of the world model, so the model can directly output a structured force-feedback signal for downstream controllers without an extra force-estimation network inferring from pixels---because this inference is integrated end-to-end within the world model. Elevating a key physical quantity from implicit inference to explicit output has direct, irreplaceable value for tasks requiring precise force control.

Motus \cite{2512.13030} handles physical soft constraints from the optical-flow angle: by learning pixel-level incremental actions---an optical-flow field of how much each pixel should move from the current to the next frame---it makes the effect of actions precise to the pixel level. Compared with a global latent-vector action, pixel-level flow provides richer physical-motion information (which regions move, and their direction and speed), achieving 15\% and 11--48\% performance gains in simulation and real robotic scenes, respectively.

A particularly noteworthy soft-constraint strategy uses a physics simulator as a bridge. The core idea of RealWonder's PhysBridge component \cite{2603.05449} is not to let the generative model face unknown physics directly, but to let a traditional physics simulator act as an intermediate translation layer: given a robot action, the physics simulator first computes the physical consequence in physical space (an optical-flow field and an approximate RGB), and the diffusion model then needs only 4 denoising steps to refine the simulator's approximate output into realistic real-time interactive video. This division---physical correctness guaranteed by the physics simulator, visual quality by the generator---is demonstrated across rigid bodies, fluids, and granular media. The idea is consistent with the design in ReconDreamer-RL \cite{2508.08170}, where a kinematic model handles physical modelling and a diffusion prior handles appearance: physical correctness and visual quality are two distinct optimization goals, and assigning them to different subsystems is more effective than a single model juggling both.

On physically-soft-constrained generative backbones, several works explore from different angles. The physics-informed embedding strategy \cite{2506.23135} enhances the 3D geometric consistency of rendering by jointly training temporal depth prediction and keypoint dynamics, where depth provides the scene's geometric structure and keypoint dynamics provide a sparse physical constraint on motion. The frame-level action-conditioning module \cite{2406.14540} strengthens the physical coherence of robot--object interaction through fine-grained action-frame alignment (keyboard- and VR-controllable), and its policy evaluation on generated video correlates highly with policy evaluation in the real environment, meaning that physical plausibility in generated video can, to some extent, substitute for real-environment evaluation. The DPO post-training strategy \cite{2603.23376} post-trains the decoder of a 14B DiT, using a decoupled discriminator to directly penalize non-physical behaviour such as clipping and anti-gravity, representing a path that directly optimizes physical plausibility at the training-objective level, and its cross-embodiment action-control results indicate that physical constraints can transfer across robot platforms. WoW \cite{2509.22642} introduces VLM guidance on a comparably large (14B) generative world model, using an external vision-language model to give feedback-style corrections on the physical plausibility of generated results and improving physical realism from embodied-interaction data---a new form of soft constraint in which a strong model judges a weaker model's output.

\subsubsection{Physical Evaluation}

\begin{figure*}[!t]
\centering
\includegraphics[width=.82\textwidth]{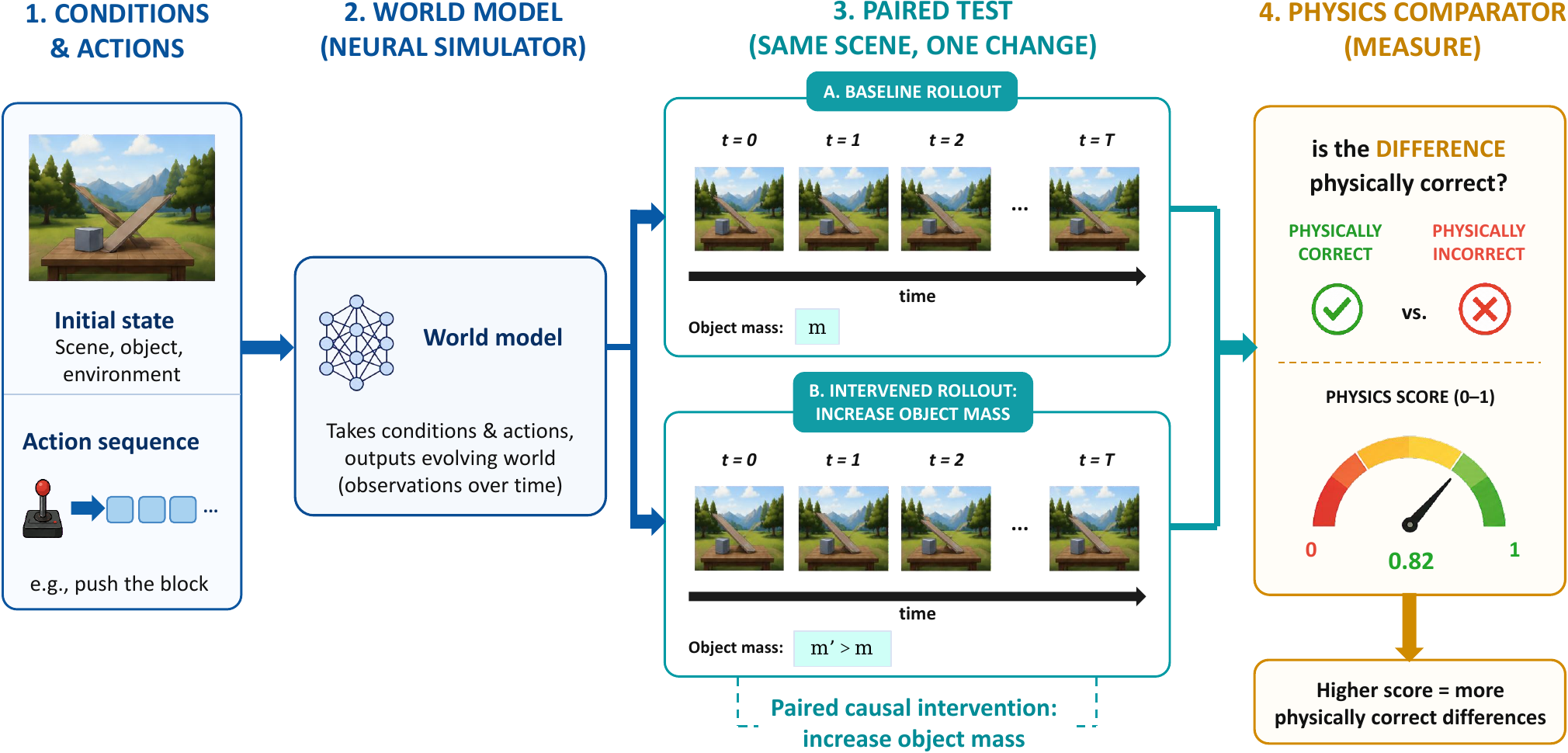}
\caption{\textbf{Physics evaluation through paired causal intervention.} A baseline and an intervened rollout share the scene and action sequence but differ in one physical variable, such as object mass; evaluation asks whether the resulting difference follows the expected physical consequence.}
\label{fig:physics-eval}
\end{figure*}

When physics is assessed rather than enforced, paired causal interventions provide a more diagnostic test than visual realism alone, as shown in Figure~\ref{fig:physics-eval}.

Whereas the hard-physics and soft-constraint routes above try to improve physical plausibility during generation, physical-evaluation benchmarks approach from another angle. What-If World \cite{2605.27589} uses 319 causal-intervention prompt pairs to test whether changing a physical variable produces the expected change while preserving the shared scene. Its APEO scoring covers adherence, physical consistency, environment preservation, and outcome difference. In its evaluated set, none of nine models exceeds a 52\% paired score and the reported open-source result reaches 28\%. Scores are also lower for visually subtle interventions, such as changing mass (14.2\%), than for visually salient changes to colour or size (40.4\%). These results indicate that visual salience may confound causal-physics evaluation and motivate more diagnostic testing rather than establishing a universal ranking of physical understanding.

ReactSim-Bench \cite{2606.14058} introduces a dedicated reactivity evaluation in the autonomous-driving domain; its core design decouples the ego vehicle's behaviour from the simulation of surrounding agents, using AV behaviour from an external planning model as an independent input to check whether a behaviour world model reacts physically reasonably to ego behaviour that deviates from the training distribution. This directly targets a core blind spot of data-driven behaviour simulators: they may be highly realistic within the training distribution but wholly unreasonable out of distribution. WorldBench \cite{2601.21282} further provides a diagnostic evaluation of single physical concepts, isolating physical constants such as friction and viscosity, and intuitive physics understanding, so researchers can locate on which specific physical concept a model is weakest.

The core gap in the physics-engine capability stems from a choice of technical orientation. A differentiable physics engine pursues precision and formal guarantees, but the physical phenomena it can simulate are limited to the forms of known equations---it can only simulate phenomena whose equations have been written down; a data-driven generative world model pursues flexibility and open-domain generalization, but what it learns may be a visual shortcut rather than a true physical law. There is tension between the two, and it cannot be eliminated by enlarging models or adding data. At least two fusion strategies are emerging: functional separation, where a physics simulator guarantees physical correctness and a generator guarantees visual quality (e.g., PhysBridge \cite{2603.05449}, ReconDreamer-RL \cite{2508.08170}); and physical parameterization, embedding key physical parameters as learnable variables in the generative model (e.g., PIN-WM \cite{2504.16693}, OrbiSim \cite{2605.16395}). Beyond these two empirically demonstrated strategies, there are also proposals from a formal-modelling angle, such as building a physics-consistent world model via Schr\"odinger-bridge optimal transport \cite{consistent2026}, which seeks to impose physical-consistency constraints on state evolution through the mathematical structure of optimal transport, though it remains at the method-proposal stage. Their commonality is turning physics from a learning target into an architectural component: the model no longer learns physics only implicitly from data, but reserves an explicit place for physics in the architecture.

\subsection{Interaction}\label{sec:c3}

Interaction is the essential feature distinguishing a simulator from a pure content-generation engine: a simulator must accept an agent's action input at any moment, update the world state in real time or near-real time, and feed the updated observation back to the agent, forming a closed loop. World models have made the most striking progress here among the eight capabilities. Of the 200 papers, 80 (40.0\%) list interaction as a principal contribution dimension, making it the second most studied capability after controllability. The analysis of interaction unfolds most naturally along two dimensions---the \emph{object} of interaction (what can be interacted with) and the \emph{depth} of interaction (the granularity and closed-loop nature)---because these two dimensions are precisely what separate an interactive content browser from a programmable simulation environment.

\subsubsection{The Object of Interaction}

The spectrum of interaction objects extends from the most basic camera control to object manipulation and then to multi-agent coordination, each extension marking the world model's evolution from a passive observer to an active participant.

\begin{figure*}[!t]
\centering
\includegraphics[width=.82\textwidth]{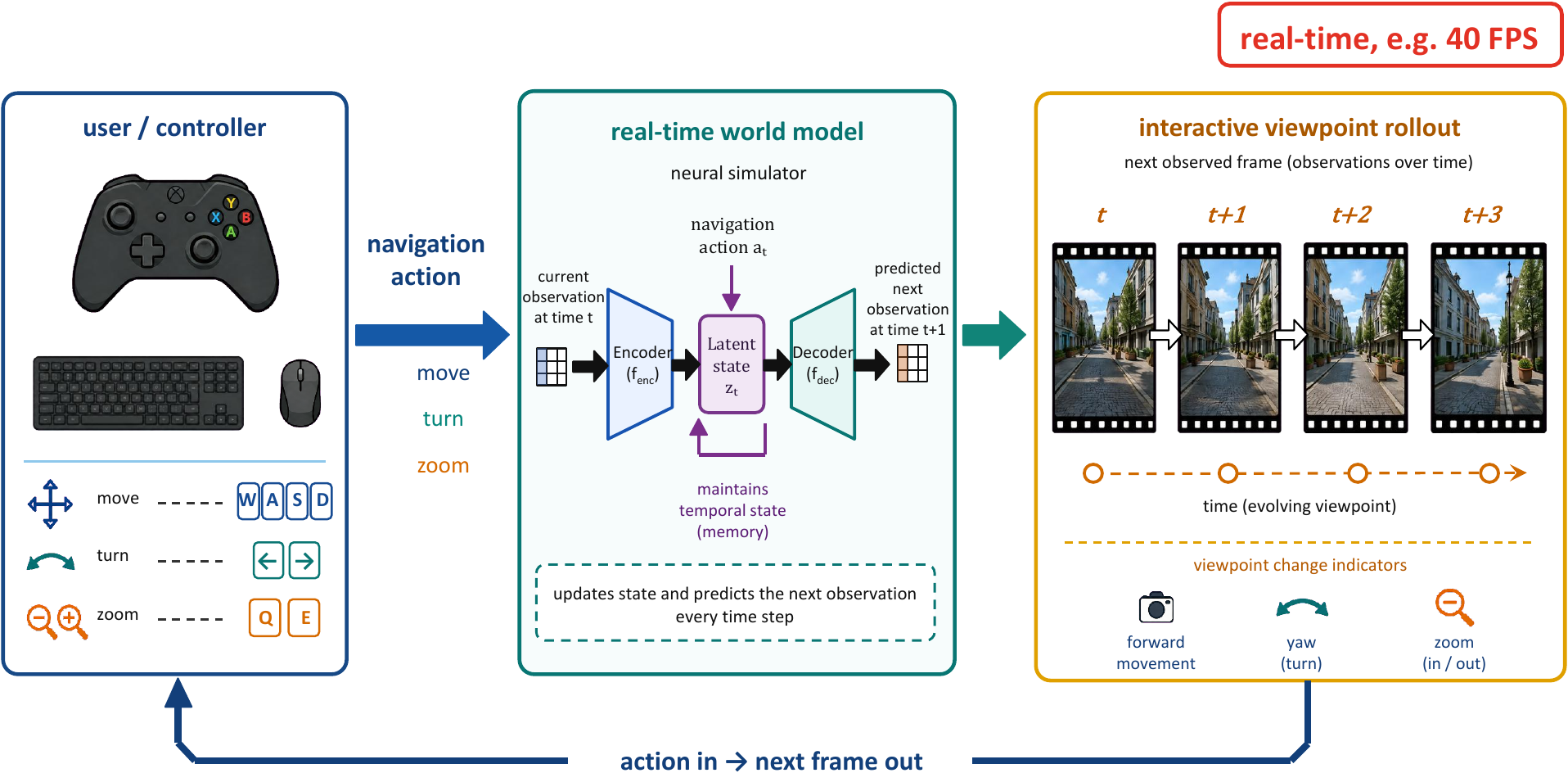}
\caption{\textbf{Camera-level interaction in a real-time world model.} Controller inputs such as movement, turning, and zoom update the model state and produce the next observation, forming an action-in--frame-out loop for interactive viewpoint navigation.}
\label{fig:interaction-camera}
\end{figure*}

Camera-level interaction is the most frequently demonstrated interaction form in the included implementation papers. Its basic closed loop is shown in Figure~\ref{fig:interaction-camera}: a navigation action updates the model's temporal state and returns the next observation from the requested viewpoint. GameNGen \cite{2408.14837}, as a proof of concept, showed that a diffusion model can provide real-time interaction in a constrained game domain: it uses two-stage training---an RL agent first self-plays in VizDoom \cite{kempka2016vizdoom} to collect 900 million frames of state and action, then an autoregressive diffusion model adapted from Stable Diffusion v1.4 \cite{rombach2022ldm} learns to predict the next frame---achieving about 20-FPS real-time interaction with 4-step DDIM sampling on a single TPU-v5. GameNGen is essentially an offline policy-distillation system relying on a traditional engine to generate training data and applies to only a single game, but its conceptual significance is proving that neural rendering can functionally substitute for a traditional pipeline within a specific constrained domain. DIAMOND \cite{2405.12399} demonstrates two uses of diffusion world models: on the Atari 100k benchmark, an RL agent trained inside the learned model reaches strong reported performance (mean HNS 1.46), while also scaling to 87 hours of CS:GO data as an interactive neural game engine, showing that interactive generation and functional simulation can coexist in one framework. Matrix-Game 3.0 \cite{2604.08995} pushes camera interaction to industrial grade: through an industrial-grade data engine (UE5 and AAA-game capture plus real-world augmentation), residual-prediction long-horizon consistency training, DMD multi-segment autoregressive distillation, and 75\% pruning of the VAE decoder, its 5B model generates at about 40 FPS at 720p resolution while maintaining minute-level memory consistency; its predecessor Matrix-Game 2.0 \cite{2512.14614} already reached 720p, 24 FPS via a continuous-plus-discrete dual-action representation and context forcing. miniWM \cite{2605.30263} provides a full-stack open-source alternative, converting existing bidirectional video-foundation models (e.g., Wan2.1, HY1.5) into few-step autoregressive generators through causal forcing and its improved training pipeline, covering the complete flow from data construction to streaming inference and providing a reproducible research substrate for academic teams without industrial-grade resources.

On the real-time front of camera interaction, several systems have made engineering breakthroughs. Hunyuan-GameCraft 2.0 \cite{2508.08601} achieves 1080P, 60-FPS real-time streaming infinite interaction via a low-latency 3D-VAE and KV cache, and decouples mechanisms such as physics and game logic from appearance rendering for editability---essentially a neural re-creation of the traditional game engine's game-logic-loop-plus-rendering-pipeline architecture. The few-step autoregressive diffusion model \cite{2508.13009} reaches minute-level long-horizon stable generation at 25 FPS through distillation. Block-diffusion semi-autoregressive inference \cite{2511.20714} achieves long-video stability via KV-cache chunked decoding and integrates the LV-Bench benchmark. The open-source high-fidelity real-time world simulator \cite{2601.20540} achieves minute-level long-term memory and sub-second-latency 16-fps interaction. Training-free attention acceleration \cite{2602.01801} maintains stable memory footprint over long rollouts through temporal-cache compression and sparse attention. This lineage of real-time infinite interaction also includes several representative systems: The Matrix \cite{2412.03568} achieves infinite-horizon, real-time controllable high-fidelity real-scene world simulation; MineWorld \cite{2504.08388} runs as a real-time, open-source autoregressive interactive world model on Minecraft, providing a reproducible baseline for the community; Matrix-Game \cite{2506.18701} offers precise action and camera control as an interactive game world foundation model; Yume \cite{2507.17744} realizes infinite generation via camera quantization plus masked video diffusion plus a memory module; converting a pre-trained video model into a controllable world simulator \cite{2502.07825} demonstrates the path of reusing large-scale video priors to obtain interaction capability at low cost; and Peekaboo \cite{peekaboo2024} gives a training-free masked-diffusion interactive video-generation scheme that injects interactive control without retraining. A common feature of these works is systematic attention to real-time performance as an engineering constraint---concerned not only with whether interaction is possible, but with whether it is fast and stable enough.

\begin{figure*}[!t]
\centering
\includegraphics[width=.82\textwidth]{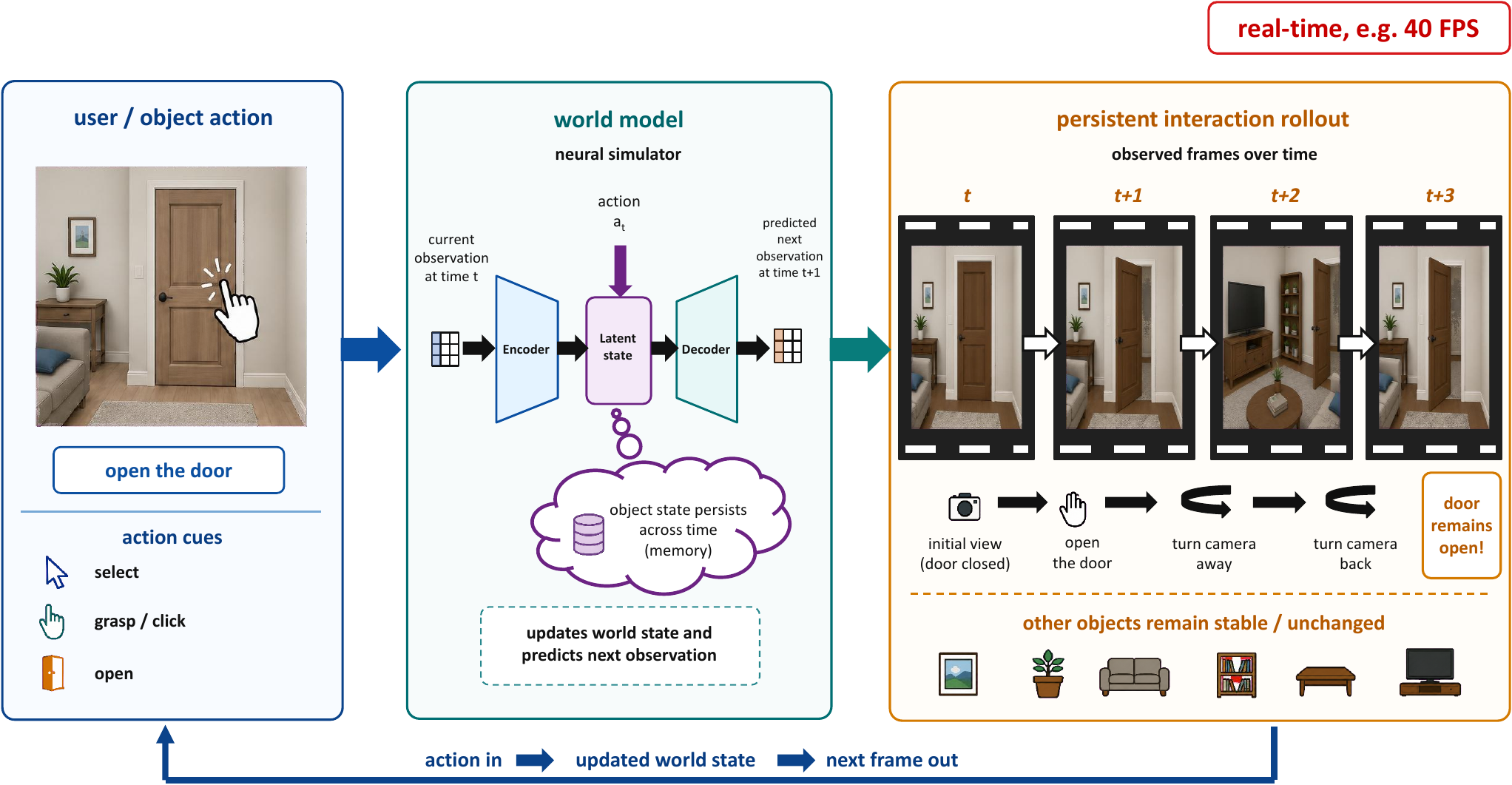}
\caption{\textbf{Object-level interaction with persistent state.} An object action updates the world state, and the modified state must remain valid across subsequent viewpoint changes; in the illustrated sequence, the door remains open after the camera turns away and returns.}
\label{fig:interaction-object}
\end{figure*}

Moving from camera navigation to object-level interaction introduces an additional state-management requirement. As Figure~\ref{fig:interaction-object} emphasizes, the action must update a persistent object state rather than only alter the current view. The action vocabulary of many camera-interaction systems in the included corpus is confined to navigation---move, rotate, zoom---while object-level operations such as grasping, opening, pushing, and assembling are less frequently demonstrated. ActWorld \cite{2606.17730} systematically identifies the root of this gap as a dual bottleneck. The first is the data bottleneck: human--object interaction videos lack precise, dense annotation, so one cannot know the event timing---that the hand contacts the doorknob at frame 137 and the door begins to rotate at frame 142. ActWorld therefore builds a dataset of 100k human--object interaction videos, each annotated with chain-of-thought-generated per-chunk captions. The second is the memory bottleneck: existing world models' bias toward recent-history compression discards the event-transition frames that causally determine an object's subsequent state, so when a user opens a door, walks away, and returns, the model has forgotten the door is open. ActWorld's hierarchical action-aware memory architecture routes history compression by interaction importance and maintains event updates and object-identity tokens in a persistent memory bank---designed precisely to address this bottleneck---so the model can both flexibly navigate the camera and perform meaningful interactive manipulation of objects in the scene. WorldCraft \cite{2605.25077} solves the same problem via a different path: Normalized World Trajectory (NWT) maps a user-drawn object-motion path into a camera-invariant world coordinate system, eliminating the confusion between object motion and camera motion in 2D projection; Spatial-Path LoRA (SP-LoRA) injects object-manipulation capability while preserving the pretrained camera controller; and Trajectory-Anchored State Persistence (TASP) treats world trajectories as persistent spatial state, so an object moved out of view can reappear at the correct new position---solving the basic causal-consistency problem that object state should be independent of the observation viewpoint.

In game scenes, object-level interaction has also advanced across several works. Genie \cite{2402.15391} infers a discrete latent-action space in an unsupervised manner from unlabelled internet videos---letting the model merely watch videos without being told that someone pressed the jump key---pioneering the paradigm of interactive game-world generation with zero human annotation. Its key is to reveal that interactivity can emerge from passive observational data: the video sequence itself contains information about actions causing state changes, only implicitly encoded in pixel changes. Genie-Envisioner \cite{2508.05635} extends this paradigm to robotic manipulation, forming a unified world-foundation platform for robotic manipulation from GE-Base (multi-view language-conditioned video diffusion), GE-Act (a lightweight action model), GE-Sim (an action-conditioned neural simulator), and EWMBench (an evaluation benchmark). GameFactory \cite{2501.08325}, through the GF-Minecraft keyboard/mouse-annotated dataset and a domain-adapter training strategy, decouples the open-domain diffusion prior from game-specific style learning and action control, achieving scene-generalizing game-video generation under precise keyboard/mouse control. Playable game generation \cite{2412.00887} verifies stable playability after 1000 frames in an autoregressive DiT diffusion and proposes the first playability-evaluation framework. The multi-person video world model \cite{2602.22208} extends interaction from single- to multi-person, achieving multi-agent synchronized multi-view interaction and long-term memory consistency via Checkpointed Self Forcing long-horizon training, with direct value for games and crowd-simulation scenes requiring multi-character simulation. The factorized latent-action model \cite{2602.16229} decomposes the scene into independent factors, inferring latent actions per entity or region, improving the precision of multi-agent interaction modelling.

In driving scenes, the objects of interaction are other vehicles, pedestrians, cyclists, and other traffic participants, whose behaviour must react physically reasonably to the ego vehicle's behaviour. OmniDreams \cite{2606.03159}, adapted by post-training from a Cosmos diffusion model and trained on 21k hours of driving data, is a real-time closed-loop autonomous-driving generative world model; its key breakthrough is using the world model not only as a sensor simulator generating realistic camera images but also as a policy backbone---its WAM (World-Action Model) variant surpasses the VLA-based Alpamayo~1.5 policy model on the NuRec dataset with one-fifth the parameters, showing the dual potential of a real-time world model as a policy architecture. The closed-loop large driving world model \cite{2412.09627} unifies perception, prediction, and planning via next-token prediction, achieving end-to-end closed-loop driving through a position-aware action tokenizer and modelling the whole flow from perception to prediction to planning. Multi-agent configurable traffic simulation \cite{2303.04116} proposes a framework generating configurable multi-agent traffic-interaction behaviour from destination plus implicit personality, giving agents aggressive or conservative driving personalities to generate more diverse, more realistic interaction patterns.

\subsubsection{The Depth of Interaction}

\begin{figure*}[!t]
\centering
\includegraphics[width=.82\textwidth]{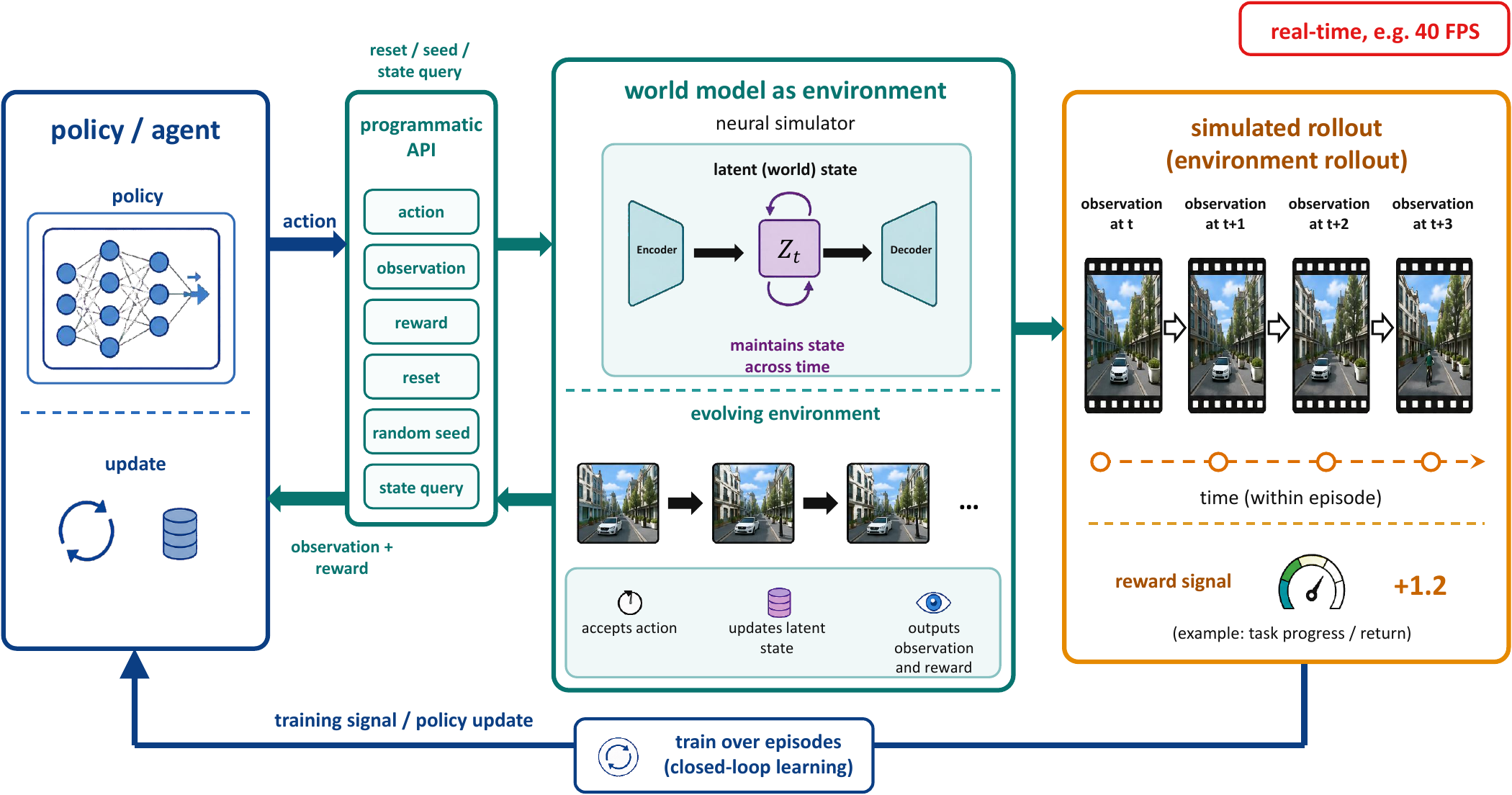}
\caption{\textbf{A world model as a programmable environment.} A policy interacts through an API carrying actions, observations, rewards, reset, random seed, and state queries, thereby closing the rollout and policy-update loop required for reinforcement-learning training.}
\label{fig:interaction-agent}
\end{figure*}

Figure~\ref{fig:interaction-agent} illustrates the interface required to move from human-in-the-loop interaction to algorithm-in-the-loop training: observations and rewards must return through a callable environment API, with reset, seed, and state-query operations available alongside actions.

The depth of interaction is the key dimension distinguishing an interactive content browser from a programmable simulation environment. The interaction mode of most existing systems is human-in-the-loop: a human operates in real time via keyboard/mouse, gamepad, or language instruction, and the model generates the next frame from the human input in real time. Algorithm-in-the-loop---e.g., an RL agent calling the world model through a programmatic API for policy training and evaluation---requires wholly different interface capabilities, including structured state queries (return the positions and velocities of all current objects), environment reset (restore the world to its initial state), random-seed control (reproduce the same interaction trajectory with a fixed seed), and reward-signal computation. These interfaces are standard in traditional simulators but are not consistently documented across the included world-model papers.

The few explorations in the algorithm-in-the-loop direction deserve special attention. DIAMOND \cite{2405.12399} predicts reward and termination signals simultaneously via an independent CNN-LSTM network and trains an RL agent inside the diffusion world model, making it one of the few world models with a complete algorithm-in-the-loop closed loop. The heterogeneous masked autoregressive model \cite{2502.04296} models cross-embodiment action-video dynamics through heterogeneous pretraining, its 15$\times$ inference speed-up making it computationally feasible to use the world model as a simulator to evaluate policies. The joint video-action latent representation \cite{2503.00200} supports, in a unified latent space, forward dynamics (state to next state), inverse dynamics (state to action), and policy (state to action) simultaneously, so a single model can serve the three core functions of forward rollout, action inference, and decision-making. The self-supervised discrete-action bottleneck \cite{2101.12195} controls video generation with a discrete action as a bottleneck, achieving frame-by-frame playable, action-controllable video generation, its discrete-bottleneck design making the action space compact and composable. iVideoGPT \cite{2405.15223} unifies vision, action, and reward into a token sequence to build an interactive, scalable autoregressive world model, and its explicit modelling of reward tokens makes it a natural fit for algorithm-in-the-loop calls from RL agents. The embodied-space world model \cite{2504.11419} shows that, in open-ended navigation tasks, a spatial world model supporting decision-making can emerge spontaneously from sparse-reward training alone, offering an example of how the world representation for algorithm-in-the-loop interaction can be acquired from the task itself.

A more fundamental interaction capability, hallmark of a traditional simulator, is action injection at any moment---the user can issue a new action command at any time during simulation, not only at the start. Pandora \cite{2406.09455} makes an important advance here, allowing language instructions to be injected at any moment of the rollout rather than only at the start of the sequence, enabling interactive mid-course intervention---e.g., opening a door in the scene at frame 100. DreamX-World 1.0 \cite{2606.16993} further enables composable run-time event triggering via event-instruction fine-tuning. Imaginative egocentric 3D exploration \cite{2411.11844} extends such anytime interaction to long-horizon exploration: the agent updates its belief about the world in continuously generated new observations and uses the updated belief to help an LLM make the next decision, forming a cognition--interaction closed loop. The video-diffusion adaptation \cite{2410.12822} converts a pretrained video-diffusion model into an action-conditioned world model via a lightweight adapter without accessing the original model parameters, a plug-and-play adaptation strategy with practical value for lowering the technical threshold of algorithm-in-the-loop interaction.

The gap in interaction is not whether interaction is possible---on the camera-navigation dimension the answer is already yes, and real-time performance (40 FPS \cite{2604.08995}) even exceeds that of some traditional simulators. The gap concentrates in three areas: the breadth of interaction objects, where object-level manipulation, physical-contact feedback, and multi-agent coordination remain nascent; the shift of interaction paradigm, where moving from human-in-the-loop to algorithm-in-the-loop requires completing interfaces for structured queries, environment reset, random seeds, and reward computation; and the persistence of interaction, where whether a door opened a thousand steps ago is still open remains a difficulty for world models relying on implicit memory.

\subsection{Controllability}\label{sec:c4}

Controllability is one of the fundamental attributes distinguishing a simulator from a free generative model. A traditional simulator provides fully deterministic control through a parametric API: set an object's position to $(1.0, 2.0, 3.0)$ and it sits precisely at that coordinate. The controllability of a world model faces a peculiar challenge: the action space is usually not a numerical physical parameter but a camera pose, language instruction, keyboard/mouse operation, or abstract latent variable, which introduces a fundamental trade-off between control precision and control generalization. Of the 200 papers, 125 (62.5\%) list controllability as a principal contribution dimension, making it the most widely studied capability---reflecting both that controllability is a common need across application scenarios and the fundamental difficulty of making a system that learns from data and inherently relies on stochastic sampling controllable.

\begin{figure*}[!t]
\centering
\includegraphics[width=.82\textwidth]{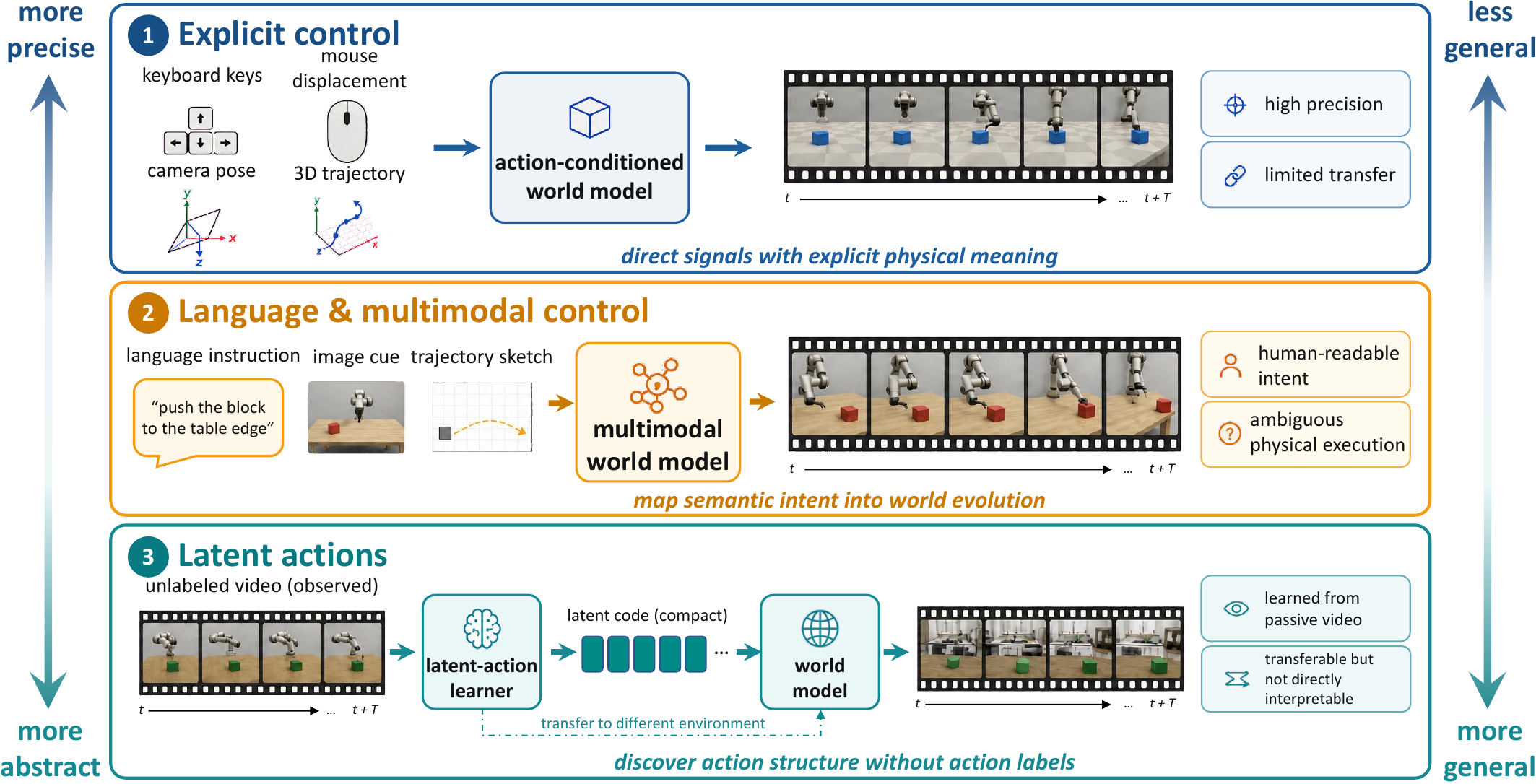}
\caption{\textbf{Controllability of world models across action-representation levels.} Explicit signals such as keyboard input, camera pose, and 3D trajectories provide precise control but transfer poorly; language and multimodal conditions express human-readable intent while leaving physical execution ambiguous; and latent actions can be learned from passive video and transferred across environments, but are not directly interpretable. Moving from explicit to latent control therefore trades physical precision for abstraction and generality, making cross-level action mapping an open problem.}
\label{fig:controllability}
\end{figure*}

The core tension of controllability lies between precision and generalization; it runs through the architectural design of all world models and exhibits a systematic trade-off pattern along the axis of the abstraction level of the action representation. We organize the analysis along this axis from most precise to most abstract, because the abstraction level of the action directly determines what granularity a user can control and how widely control generalizes. Figure~\ref{fig:controllability} summarizes this continuum and the corresponding shift from physically interpretable commands to learned but less transparent action codes.

\subsubsection{Explicit Control}

At the most precise end, actions are represented as signals with clear physical meaning, including keyboard keys, mouse displacement, camera pose, and 3D trajectory coordinates. This class of control is closest to the API-call mode of a traditional simulator---the user specifies precisely what is desired and the system executes it precisely.

DisCo \cite{2606.07967} poses a fundamental challenge to the mainstream scheme of using a continuous camera trajectory as the action condition. Its core finding is that continuous representations make the feature similarity between different motion patterns too high---the trajectories of \emph{advance 1.0 m} and \emph{advance 1.1 m} are nearly indistinguishable in continuous space---causing the model to confuse these close-but-different motion commands during denoising. DisCo's solution is to discretize the action space into a compact set of motion primitives, such as turn left 30$^\circ$, advance, stop, which are well separated in feature space, thereby greatly improving the reliability of action-following. The significance exceeds the specific scheme: it shows that in certain dimensions, coarse-grained discretization may outperform fine-grained continuity, because separability matters more than resolution. Prisma-World \cite{2606.09507} extends control from single- to multi-agent: via multi-agent RoPE distinguishing agent identity while keeping synchronized time coordinates, injecting an attention bias from relative camera geometry to steer overlapping viewpoints toward shared scene evidence, and an overlap-decay curriculum, it achieves cross-view consistent generation for any number of agents under complex camera trajectories. This has direct value for simulation scenes requiring multi-sensor fusion (e.g., surround-view multi-camera for autonomous driving) and multi-character game simulation requiring per-character behaviour specification.

A new dimension of control extension is multi-subject action binding. The subject-state token plus spatial bias \cite{2604.02330} decouples frame rendering from individual action updates, maintaining each player's action-following precision while controlling 7 players on the same screen---technically non-trivial, because without decoupling the action signals of 7 players would interfere in the shared pixel space. On the driving-control side, multi-view latent diffusion \cite{2503.20523} generates multi-camera spatio-temporally consistent driving video with fine-grained control of ego, agents, and semantics, covering rare scenes. Pose-memory retrieval plus frame-level action control \cite{2510.10125} achieves over-20-second long-term consistency without real-robot data, with direct value for robot-policy-ranking evaluation. The GPT-style driving world model \cite{2412.19505} achieves 40-second stable drift-resistant generation via spatio-temporal fusion and mask reweighting. Drive-WM \cite{2311.17918} generates multi-view controllable driving video through view-factorized joint spatio-temporal modelling and applies it to look-ahead planning, embodying the idea of controllable generation serving decision-making directly; EOT-WM \cite{vehicle2026} argues that controlling the ego trajectory alone is insufficient and further models ego and other-vehicle trajectories jointly, extending controllability to the interactions among traffic participants. ProphetDWM \cite{2505.18650} rollingly predicts future actions and video in driving scenes; its closed-loop rolling-prediction mode from action to video to action essentially re-creates the step loop of a traditional simulator inside the world model: given the current state and action, predict the next state, then select a new action based on the new state, and so on.

\subsubsection{Language and Multimodal Control}

Language instruction as an action condition has a natural advantage---humans can express intent intuitively---but also a natural disadvantage: a natural-language instruction such as \emph{push the block to the edge of the table} cannot precisely specify the end-effector trajectory. The core challenge of language control is the mapping from semantics to physics.

An important contribution of Pandora \cite{2406.09455} is to solve the problem of \emph{when} control can happen, allowing language instructions to be injected at any moment of the rollout rather than only at the start; this anytime-controllable interaction mode is closer to how a simulator works than one-off video generation. DreamX-World 1.0 \cite{2606.16993}, via event-instruction fine-tuning and RL alignment, restores camera control and visual quality after distillation, and its lightweight projected positional encoding E-PRoPE retains the ability to project camera geometry while reducing computation. The instruction-driven interactive game world model \cite{2511.23429} controls camera, character, and environment via text injection and is accompanied by the InterBench evaluation benchmark. Text-controlled world events plus keyboard interaction \cite{2512.22096} combines text-generated events (e.g., a car approaching from the left) with keyboard interaction (e.g., pressing the right-turn key), achieving a continuously explorable world through context compression. Controllable panoramic video generation \cite{2603.30045} achieves trajectory-controllable long-range scene roaming through a preview-and-refine two-stage strategy, first quickly generating a coarse panoramic point cloud at low resolution and then finely rendering at high resolution.

On the driving-control side, ADriver-I \cite{2311.13549} proposes the concept of interleaved vision-action pairs, unifying visual features and control signals into one format and, through an autoregressive loop of predicting the current-frame control signal, conditionally generating the next frame, and then predicting from the new frame, achieving continuous driving in its self-generated world. DrivingGPT \cite{2412.18607} unifies driving world modelling and planning via image-action-token interleaved autoregression, surpassing strong baselines on nuPlan and NAVSIM through standard next-token prediction on multimodal driving language, proving that world modelling and behaviour planning can be solved in the same sequence-modelling framework. The unified vision-language-world-action model \cite{2603.25741} further extends this idea, achieving personalized driving planning (e.g., choosing a scenic route or avoiding all toll booths) and trajectory generation in an instruction-driven manner, making the driving world model not only an environment simulator but also a behaviour decision-maker. UniDrive-WM \cite{2601.04453} follows a closely related, VLM-driven route, unifying scene understanding, behaviour planning, and video generation in one driving world model so that language-level intent propagates consistently to controllable generation and planning output.

On the robotic-control side, UniPi \cite{2302.00111} proposes a concise framework reformulating the policy problem as text-conditioned video generation: generate a video as a planning blueprint, then extract control actions from it for execution, achieving cross-task, cross-environment compositional generalization; this works because the compositionality of the video-generation space is far higher than that of the action space, and synthesizing an unseen task in video space is more feasible than in action space. AVDC \cite{2310.08576}, via the entirely different path of dense inter-frame correspondence rather than text guidance, infers actions from only RGB video and a text goal, requiring no action annotation. Its methodological significance is proving that action, as a kind of information, can be inversely inferred from state changes: given the before and after states, what action caused the change is learnable. The earlier Diffuser \cite{2205.09991} pioneered the paradigm of flexible behaviour synthesis by iteratively denoising an entire trajectory with a diffusion model, grounding controllability in direct modelling of the trajectory distribution. Building on this, several works push controllability to a finer robotic-manipulation granularity: ManipDreamer \cite{manipdreamer2026} organizes a manipulation world model with an action tree plus multimodal visual guidance, so a complex operation can be decomposed into controllable sub-steps; STARRY \cite{2604.26848} proposes action-centric spatio-temporal world modelling with a geometry-aware attention (GASAM), improving the geometric accuracy with which actions control scene evolution; and STORM \cite{2512.18477} combines a diffusion VLA, video prediction, and MCTS search into a search-guided generative world model, raising the controllability and success rate of manipulation actions by searching within the generated imagination space.

\subsubsection{Latent Actions}

At the most abstract end, actions are represented as learned latent variables, with the model itself deciding what constitutes a meaningful action unit, and humans unable to specify them directly. The core progress of the latent-action route can be understood as an evolution from finding actions, to understanding actions, to making actions meaningful.

Genie \cite{2402.15391} pioneered this paradigm, inferring a discrete latent-action space in an unsupervised manner from unlabelled internet videos. Genie's key contribution is proving that latent actions can emerge from passive video data: without anyone telling the model that the user pressed the left key here, the model automatically discovers action boundaries by judging which pixel changes are predictable, consistent, and compressible into a discrete code. But Genie's limitation is also here: the latent actions it discovers are independent per training environment, and a jump learned in a platformer cannot transfer to a racing game. AdaWorld \cite{2503.18938} advances a step, proposing transferable latent-action learning so that latent-action representations learned in one environment can transfer to another. Motus \cite{2512.13030} bridges latent action and the physical world, learning pixel-level incremental actions through optical flow so that latent actions can be decoded into an interpretable physical-motion field.

DiLA \cite{2605.15725} represents the deepest theoretical advance of the latent-action route so far, revealing and exploiting a previously under-recognized fact: the decoupling of content and structure co-evolves with latent-action learning. DiLA's core insight is that the prediction bottleneck in latent-action learning is itself the driver of decoupling. To compress future state changes into a limited latent-action code, the model is forced to separate \emph{what changes} (structure, i.e., spatial layout, object position) from \emph{what stays constant} (content, i.e., texture, colour, style) into different pathways, so that the latent-action codes in the structure pathway naturally align to physically meaningful changes. This symbiosis---decoupling drives action learning, and action learning in turn reinforces decoupling---lets DiLA simultaneously achieve a highly abstract semantic action space and high-fidelity visual generation, two goals previously thought mutually constraining. The principle DiLA reveals is of general significance: the quality of latent actions may ultimately depend on how well the model separates scene content from structure, rather than on the engineering design of the action space itself. This suggests the key to improving controllability may lie more in the decoupling mechanism.

After DiLA, the latent-action route continues to deepen in several directions. Learning continuous-constrained latent actions from in-the-wild video \cite{2601.05230} captures the complexity of in-the-wild actions, no longer confined to controlled game or robot scenes, and uses the learned latent actions as a planning interface. Sequence control-effect alignment \cite{2602.10104} proposes a strategy of aligning action semantics by anchoring temporal feature differences, achieving zero-shot action transfer---a push learned in one environment transfers zero-shot to a push in another. The hierarchical latent-action model \cite{2603.05815} further aggregates low-level latent actions into high-level skills, modelling multi-level temporal structure in long-horizon control tasks. Shared latent action \cite{2512.10016} reduces annotation-sample demand by an order of magnitude by aligning control signals with passively observed actions, of significant practical value in robotic applications with scarce annotated data.

On the downstream utility of latent actions, Being-H0.7 \cite{2605.00078} proposes a concise dual-branch training design: during training, the posterior branch uses embeddings of future observations to provide prescient information, while the prior branch uses only the current context to provide deployable information, and the two are aligned in latent space; at inference, the posterior branch is discarded and only the prior branch is used, thereby gaining the future-aware advantage of a world model while preserving VLA deployment efficiency. This train-time-fuse, inference-time-separate design may offer a general paradigm for unifying predictive world models and efficient VLA policies.

The unsolved challenges of controllability concentrate on mapping problems at two levels. The first is cross-abstraction-level mapping: how to build a bidirectional mapping among explicit control (precise but poorly generalizing), language instruction (semantically readable but low precision), and latent action (well generalizing but unreadable)? This requires a world model to simultaneously learn action representations of multiple granularities and their translation relations. The second is temporal compositionality---e.g., press the left key for 5 seconds, then the right key for 3 seconds; current systems remain unstable in supporting such temporally composed actions, because the segments of a composed action may be causally disconnected, with the conditional observation of the second segment already deviating from the true world state. This is essentially a problem between open-loop control and closed-loop feedback: a traditional simulator natively supports closed-loop feedback where the current state influences the next action, whereas current world models are more often open-loop sequence continuation. So far only V-JEPA~2-AC's \cite{2506.09985} receding-horizon MPC and WorldVLA's \cite{2506.21539} attention-mask strategy have explored true closed-loop control, and the fundamental challenge of temporal compositionality remains unsolved.

\subsection{Stability}\label{sec:c5}

Stability refers to the reliable consistency of generated results along the temporal and spatial dimensions---no unwarranted drift, distortion, or abrupt change---and it is the key dividing line between a demonstration tool and an engineering simulator. A traditional simulator's stability comes at almost no extra cost: being solver-based, it reproduces identical results infinitely for the same input sequence. A world model's stability is inherently constrained by its generative paradigm: whether autoregressive frame-by-frame generation or diffusion denoising followed by re-encoding, each step's tiny error accumulates over long horizons, forming a vicious cycle of error, deviation, generate-on-deviation, larger-deviation---i.e., \emph{autoregressive drift}. Of the 200 papers, 80 (40.0\%) list stability as a principal contribution dimension, a high proportion reflecting a broad consensus in the research community: autoregressive drift is the number-one engineering obstacle to the practical use of world models.

The analysis of stability is organized most naturally by the mechanism that maintains it, because different mechanisms represent not only different technical schemes but also different answers to the fundamental question of what really guarantees stability. The stability mechanisms in the current literature fall into four levels: self-adversarial training at the training-strategy layer, implicit and explicit memory at the memory-architecture layer, state persistence at the causal-consistency layer, and drift quantification at the theoretical-guarantee layer.

\subsubsection{Self-Adversarial Training}

\begin{figure*}[!t]
\centering
\includegraphics[width=.82\textwidth]{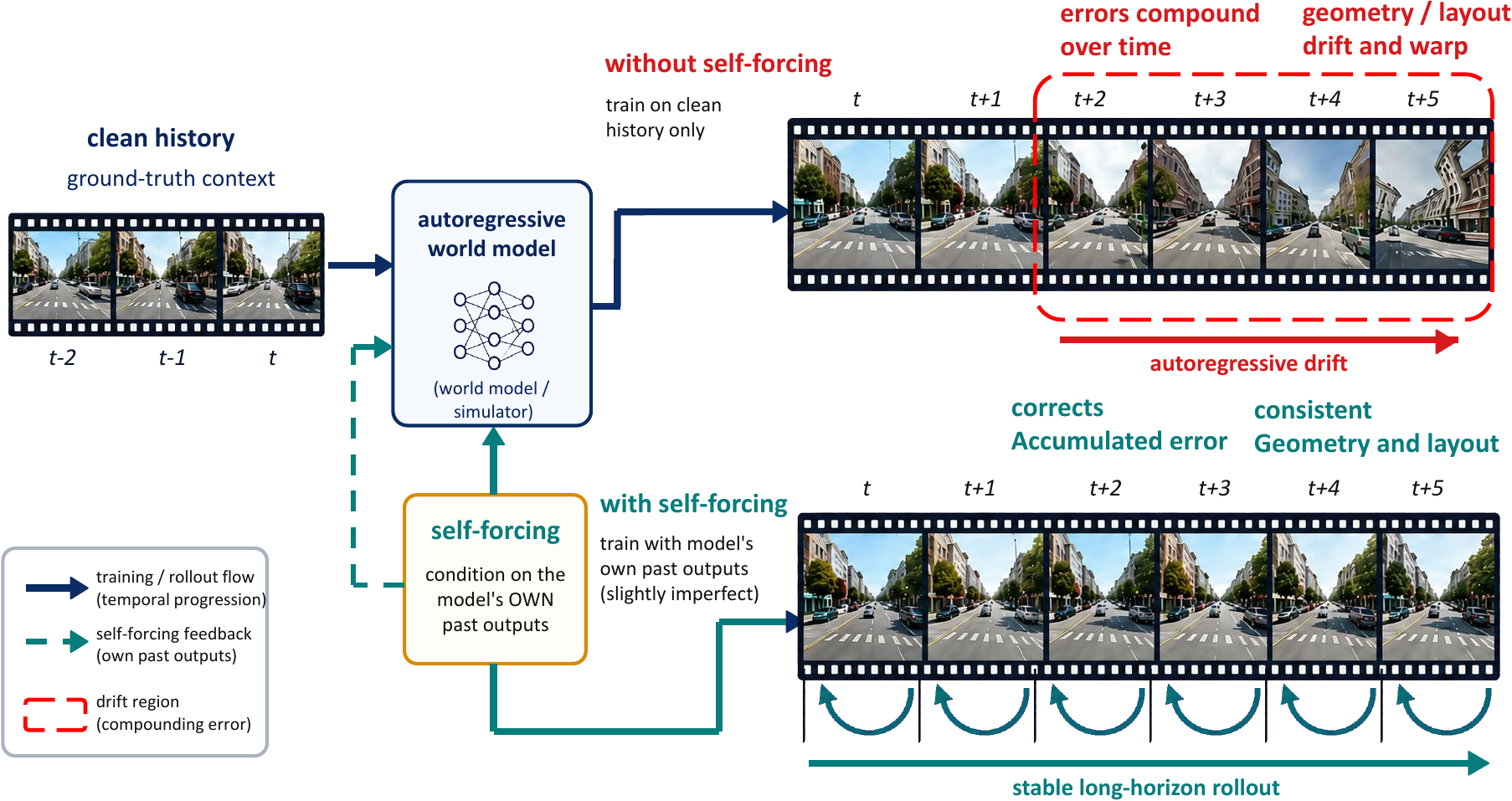}
\caption{\textbf{Self-forcing training against autoregressive drift.} Training only on clean history creates a train--test mismatch under which errors compound during rollout; conditioning on the model's own past outputs teaches recovery from imperfect context and improves long-horizon geometric consistency.}
\label{fig:stability-self-forcing}
\end{figure*}

The self-forcing mechanism in Figure~\ref{fig:stability-self-forcing} directly targets exposure bias by replacing clean-only conditioning with the model's own imperfect rollout history during training.

This is a stability-enhancement strategy independently verified as effective by multiple systems. The core idea is that, rather than trying to perfectly eliminate every step's autoregressive error (impossible in a stochastic-optimization framework), it is better to deliberately expose the model during training to its own imperfect outputs, forcing it to learn correction. Matrix-Game 2.0's \cite{2512.14614} context forcing is a representative implementation: during training it replaces the true history frame with the model's own previously generated history frame with some probability, as the condition, forcing the diffusion model to still generate a reasonable next frame when the context has slightly deviated from the true world state. Lyra~2.0's \cite{2604.13036} self-augmented history training follows the same logic, training the model to correct rather than propagate drift by exposing it to its own degraded outputs on long trajectories. DreamX-World 1.0's \cite{2606.16993} self-generated long-horizon context training and residual-recycling mechanism further enhance robustness to imperfect conditions. Checkpointed Self Forcing \cite{2602.22208} extends this strategy to multi-person scenes, proving that self-adversarial training is equally effective in more complex interaction scenes. The common insight of these methods distils into a design principle: in a system where errors are unavoidable, the ability to recover matters more than the ability to avoid---essentially a mapping of robust-control thinking into generative models.

\subsubsection{Implicit and Explicit Memory}

\begin{figure*}[!t]
\centering
\includegraphics[width=.82\textwidth]{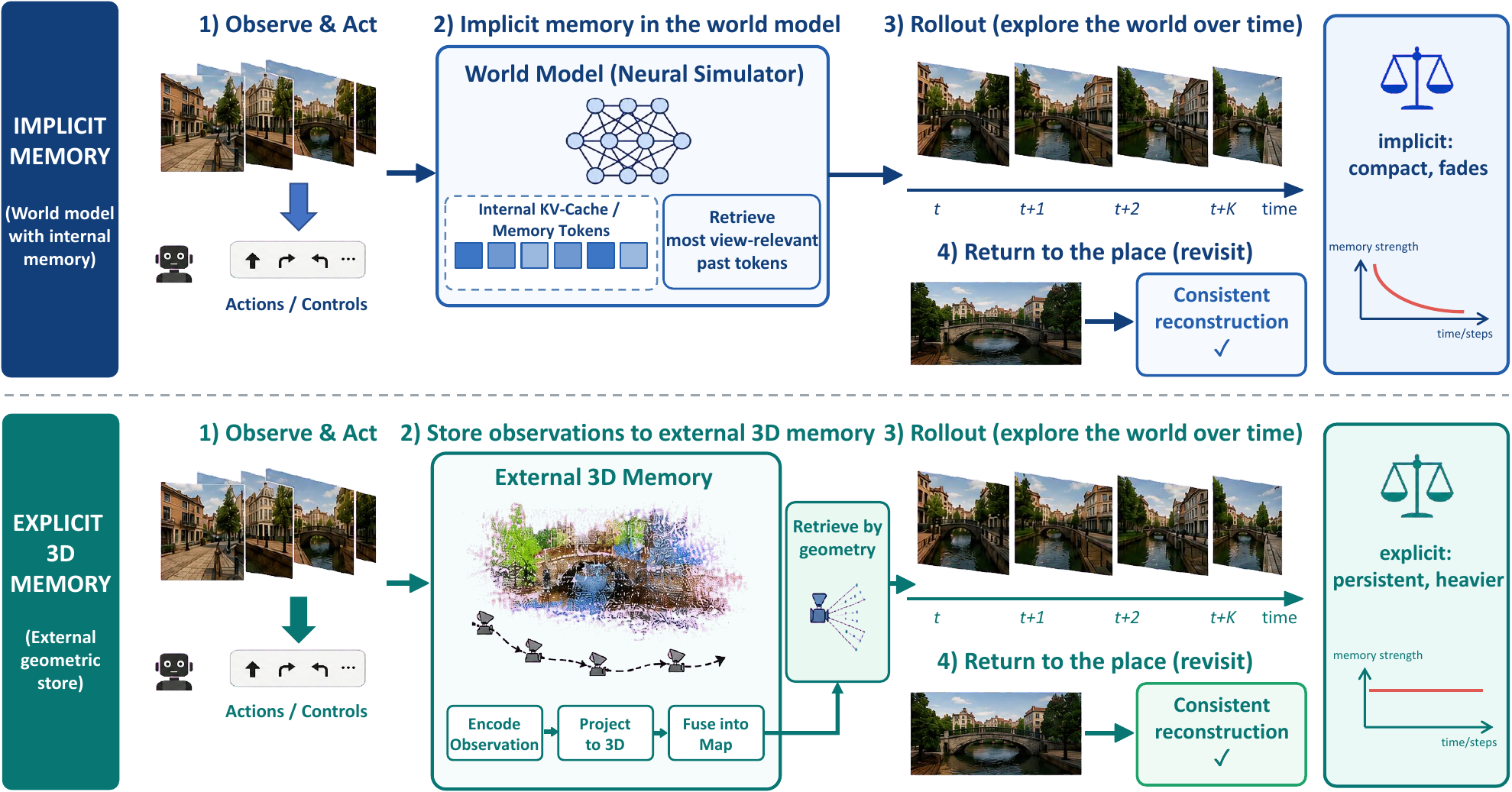}
\caption{\textbf{Implicit versus explicit memory for consistent world rollouts.} Implicit memory retrieves view-relevant internal states but may fade with horizon length; explicit 3D memory stores observations in a persistent geometric map and retrieves them when the camera revisits a location.}
\label{fig:stability-memory}
\end{figure*}

Figure~\ref{fig:stability-memory} contrasts the two principal memory designs: internal caches or tokens provide compact contextual recall, whereas an external geometric store preserves spatial information more persistently at greater computational and storage cost.

The essential cause of autoregressive drift is that past information is gradually forgotten in the sequence: if the model does not remember what the initial position looked like, it cannot maintain consistency. Therefore, how to remember the past is the core of stability-architecture design.

Implicit memory retains historical information through the model's internal state (e.g., RNN hidden state, Transformer KV cache, or learnable memory tokens) and is the choice of the vast majority of systems, its latest advances embodying delicate engineering innovation. The KV-cache camera-aware memory of RELIC and Oasis \cite{2512.04040} retrieves the most relevant historical KV pairs from the cache according to the current viewpoint, letting the model recall the past information most relevant to the current view rather than treating all history equally. DecMem \cite{2605.31336} takes a step toward theorization, systematically analysing two fundamental limitations of naive learnable memory architectures---computational inefficiency (full attention over long history grows as $O(n^2)$) and attention dispersion (attention weights diluted over many irrelevant tokens in long sequences)---and proposing a decoupled memory architecture: sparse global memory achieves $O(\log n)$-level fine-grained history access, and anchored local memory maintains dense attention within a local window to guarantee extrapolation quality; this decoupled design makes it one of the best current schemes for minute-level long-video consistency. WorldPack \cite{2512.02473} compresses history into compact memory tokens; memory-compressed KV cache \cite{2512.18741} maintains long-term consistency in a memory-efficient way and establishes the MAG-Bench benchmark. The GLP architecture \cite{2511.09057} fuses an LLM's latent dynamics with a diffusion decoder, achieving long-horizon consistent simulation and reasoning conditioned on natural-language actions, where the LLM provides abstract long-horizon memory and reasoning and the diffusion model provides concrete visual rendering.

Explicit 3D memory takes an entirely different approach: rather than relying on the model's internal learnable parameters to remember the past, it explicitly stores historical observations as a persistent 3D spatial representation (e.g., point cloud, voxels, 3D Gaussians). Long-term Spatial Memory \cite{2506.05284} represents this route, lifting each frame's observation into an explicit representation in 3D space and, when generating a new view, retrieving relevant information from this 3D store through geometry-aware feature retrieval. The fundamental advantage of explicit memory is that its memory does not decay with sequence length---a point placed in 3D space stays there, unconstrained by the number of autoregressive steps; but its cost is also clear: storage overhead grows linearly with scene scale, and updating and evicting stale memory in dynamic scenes is a non-trivial problem requiring additional mechanisms. Persistent global-state modelling \cite{2603.07145} solves the off-view problem of explicit memory in dynamic scenes by synchronously evolving off-view dynamics through a monitor mechanism, accompanied by the LiveBench benchmark for systematic evaluation. Explicit-3D-memory embodied simulation \cite{2505.05495} predicts RGB-D and aggregates it into a persistent 3D map as a condition for generating the environment, preliminarily fusing the advantages of implicit generation and explicit memory: the diffusion model predicts RGB-D to guarantee appearance quality and flexible motion modelling, while the explicit 3D map achieves persistence so memory does not decay. VMem \cite{vmem2025} offers a lightweight realization of this route: rather than maintaining a dense point cloud or voxels, it organizes historical-view memory indexed by surfels, retrieving the most relevant historical frames by geometric proximity as conditions when generating a new view, thereby maintaining long-term visual consistency under free camera roaming. This sparse explicit memory keyed by surfels strikes a more economical compromise between storage overhead and consistency than dense 3D storage.

\subsubsection{Causal Consistency}

\begin{figure*}[!t]
\centering
\includegraphics[width=.82\textwidth]{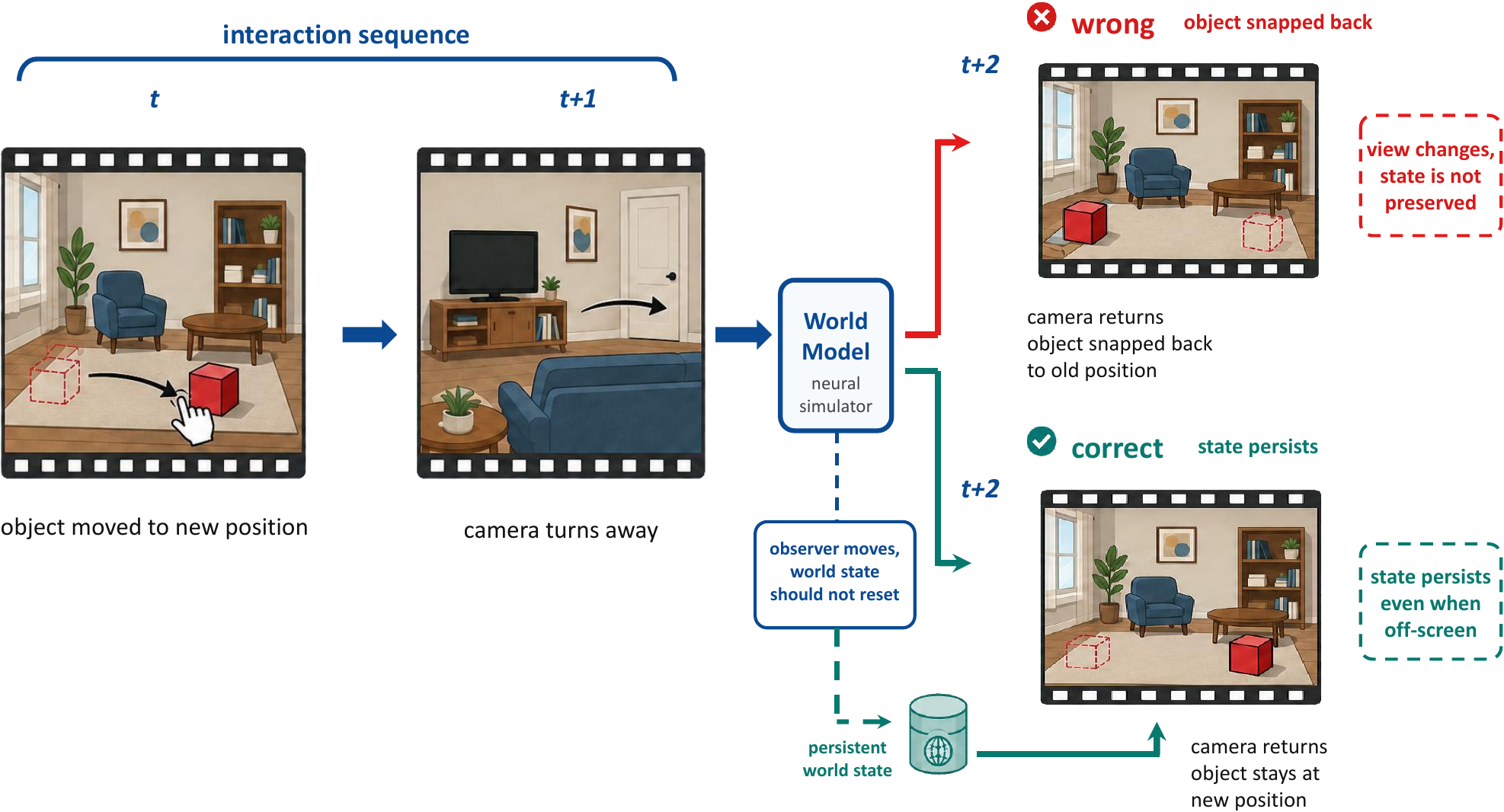}
\caption{\textbf{Causal consistency under viewpoint change.} After an object is moved and the camera turns away, a model without persistent state may place the object back at its former location; a consistent world model retains the intervened state when the object is observed again.}
\label{fig:stability-causal}
\end{figure*}

The off-screen persistence test in Figure~\ref{fig:stability-causal} isolates a basic causal requirement: changing the observer's view must not reset the underlying world state.

This is a special but instructive form of stability: an object's state (position, orientation, on/off) should be an objective fact independent of whether the observer is watching. WorldCraft's \cite{2605.25077} Trajectory-Anchored State Persistence (TASP) specifically solves this, so a moved object reappears at the correct new position after leaving the camera view rather than returning to its original place because the camera did not see it. This causal-consistency problem seems simple but is highly challenging in autoregressive video generation, because the model has no persistent world-state variable and its memory is entirely implicit and attached to recent frames. Prisma-World's \cite{2606.09507} multi-agent joint denoising processes all agents' video in one full-attention sequence, essentially ensuring that multiple observers see the same world: if two agents' viewpoints overlap, the overlapping region should show identical scene content. This cross-subject consistency is a natural generalization of temporal consistency to spatial consistency.

\subsubsection{Quantification of Drift and Theoretical Guarantees}

The quantitative characterization of autoregressive drift remains poorly standardized across model architectures, scene complexities, and action-sequence lengths. Existing statements of long-horizon stability---such as DecMem's minute-level \cite{2605.31336}, Matrix-Game 3.0's minute-level memory consistency \cite{2604.08995}, and the GPT-style driving model's 40 seconds \cite{2412.19505}---all lack a unified evaluation protocol: what counts as staying stable? PSNR not below a threshold, FVD \cite{unterthiner2018fvd} not diverging, or humans unable to tell real from fake? Different papers use different standards and cannot be compared directly. In driving, Fine-flow \cite{flow2026} proposes suppressing drift via coarse-to-fine distillation of optical flow: constraining the global motion trend at coarse granularity first, then progressively refining to local detail, keeping long-horizon generated motion coherent without diverging---a concrete drift-suppression mechanism grounded in motion consistency. MiLA \cite{2503.15875} attacks the long-horizon-consistency problem from the data and generation end, with multi-view high-fidelity long-term driving-video generation; SWM \cite{2603.15583} instead anchors generation to a real-world metropolis via retrieval augmentation and stabilizes long-horizon rollouts with a Virtual Lookahead Sink mechanism---the two representing complementary drift-suppression paths in the demanding driving setting. Within the included corpus, we did not identify an analysis of worst-case theoretical stability bounds: for a given model architecture and action distribution, is there a theoretical upper bound on the growth rate of autoregressive drift? If so, how do model capacity, training-data scale, and distillation strategy systematically raise this bound? NWM \cite{2412.03572} demonstrates the feasibility of using a diffusion world model for trajectory simulation and evaluation in navigation scenes---planning a path through simulation in familiar environments and imagining a navigation trajectory from a single image in unfamiliar ones---but its long-horizon stability (how far one can simulate before it becomes unreliable) is not systematically characterized. PlaNet's \cite{1811.04551} latent overshooting and DreamerV1's \cite{1912.01603} latent-space imagination, as early milestones of the latent-dynamics route, demonstrated the feasibility of long-horizon imagination in latent space (multi-step prediction not collapsing); MuZero \cite{1911.08265}, with a rule-free latent model plus tree search, repeatedly invokes its learned dynamics in long-horizon planning without diverging, corroborating from a planning perspective that a compact latent space supports long-horizon stability. But these works are five to seven years old, and whether their stability strategies remain effective or need adjustment in today's larger-scale models lacks systematic comparative study.

The second stability subattribute is run reproducibility: whether a fixed input sequence and a controlled source of randomness produce a repeatable rollout. Stochastic generation does not rule out reproducibility in principle, but the included corpus provides little directly comparable evidence on exposed seeds, deterministic kernels, tolerance bands, or repeat-run variance. Without such a protocol, A/B testing, regression testing, and safety analysis cannot separate model changes from sampling variation. Future evaluations should therefore report both long-horizon consistency and repeat-run reproducibility, rather than treating success on either one as evidence for the other.

\subsection{State Feedback}\label{sec:c6}

State feedback is the nervous system of a traditional simulator: a simulator must not only correctly evolve the world state but also return key information about the current state to external callers (human or algorithm) in a structured, consumable format. Traditional simulators have a natural advantage here---ego six-DoF pose, depth maps, LiDAR point clouds, semantic and instance segmentation, collision events, object bounding boxes, reward signals, and ground-truth labels are all explicitly modelled in the simulator design and accessible at any time via API. Of the 200 papers, only 45 (22.5\%) list state feedback as a principal contribution dimension---a proportion so low that it points precisely to a clear bias in the current research paradigm.

To characterize this absence precisely, we perform a five-variable state-interface audit of all 163 implementation papers. B1 is ego information (camera pose, vehicle speed, robot joint angles); B2 is sensor-level output (depth, semantics, LiDAR, normals, occupancy); B3 is task-evaluation signals (reward, termination flag); B4 is a runtime interface for queryable, named non-ego entity states or physical parameters, such as an agent's position, velocity, type, and size or an object's pose, mass, and friction; and B5 is closed-loop interaction (accepting actions and continuously generating). B4 excludes sensor-space predictions assigned to B2, unreadable latent vectors, and structure used only as an input, training target, or evaluation reference. Table~\ref{tab:feedback-audit} reports present, absent, and unresolved judgements separately. Confirmed presence is recorded for B1 in 65 papers, B2 in 45, B3 in 20, B4 in six, and B5 in 87.

\begin{table*}[!t]
\centering
\caption{\textbf{State-interface audit for all 163 implementation papers.} ``Present'', ``absent'', and ``unresolved'' are reported separately; each row sums to 163. B4 requires a runtime, queryable entity/physics annotation interface and excludes sensor-space estimates, latent vectors, and training-only labels. B5 is a cross-cutting closed-loop attribute.}
\label{tab:feedback-audit}
\small
\setlength{\tabcolsep}{9pt}
\begin{tabular}{@{}p{.08\textwidth}p{.58\textwidth}rrr@{}}
\toprule
Variable & Operational meaning & Present & Absent & Unresolved \\
\midrule
B1 & Ego information: camera/ego pose, vehicle state, robot joint or end-effector state & 65 & 74 & 24 \\
B2 & Sensor output: depth, semantics, LiDAR, normals, occupancy, or comparable sensor modality & 45 & 113 & 5 \\
B3 & Task signal: reward, termination, or an equivalent task-evaluation signal & 20 & 137 & 6 \\
B4 & Runtime entity/physics annotations: named non-ego entity states or physical parameters & 6 & 157 & 0 \\
B5 & Closed-loop interaction: repeated action-conditioned generation or model-mediated control loop & 87 & 56 & 20 \\
\bottomrule
\end{tabular}
\end{table*}

The paper-level evidence nevertheless exposes an important interface distinction. Many video-diffusion and autoregressive systems generate RGB and rely on a downstream perception model to recover structure. This pipeline may be useful, but its estimated output is not equivalent to state read directly from an explicit simulator. The distinction matters for information loss, calibration, and auditability, and it should be reported independently of output modality.

Among works that confront the state-feedback problem, three technical strategies can be identified, ordered by how closely their outputs approach the interface of a traditional simulator. These strategies differ not only in output modality but also in whether the returned state is human-readable, physically grounded, or intended primarily for machine consumption.

\subsubsection{Multi-Task Auxiliary Output}

\begin{figure*}[!t]
\centering
\includegraphics[width=.82\textwidth]{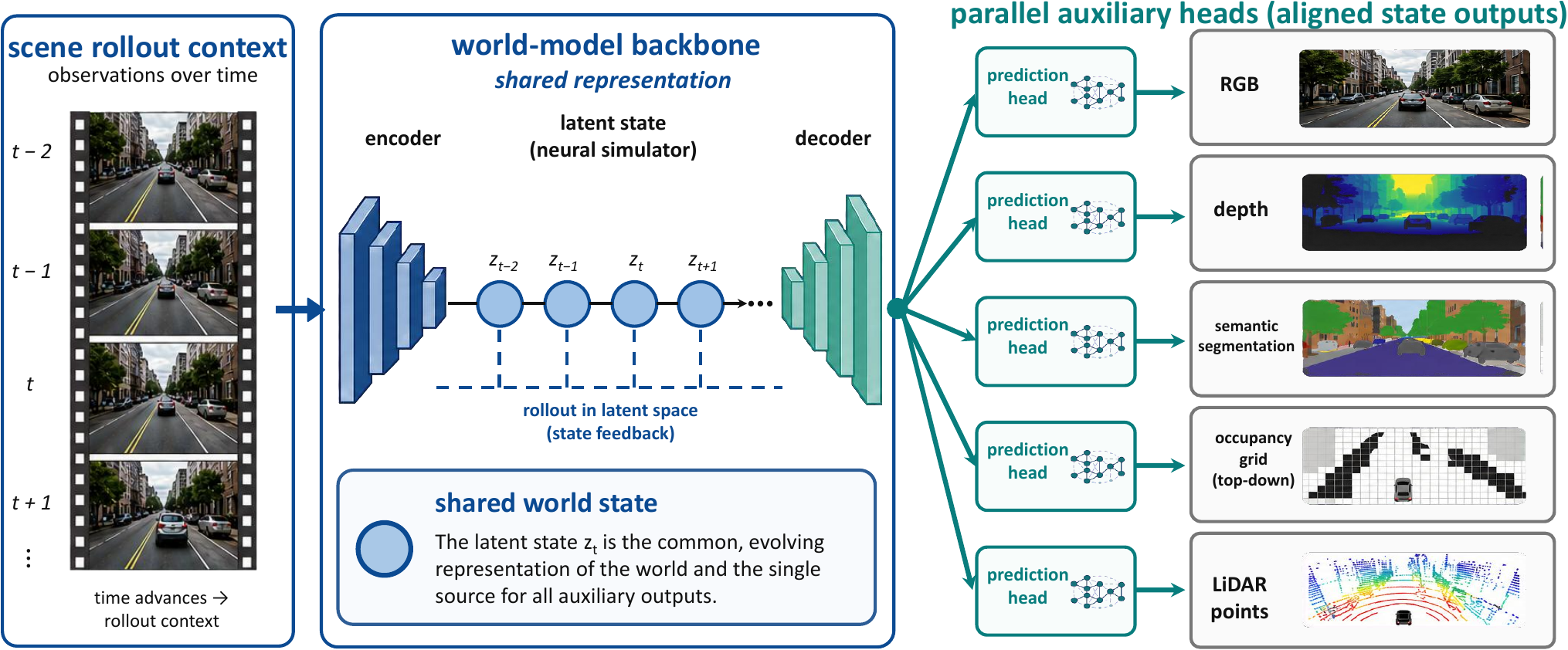}
\caption{\textbf{Structured feedback through multi-task auxiliary outputs.} Parallel prediction heads decode a shared temporal representation into aligned RGB, depth, semantic segmentation, occupancy, and LiDAR outputs, providing downstream systems with complementary views of the same simulated state.}
\label{fig:feedback-multitask}
\end{figure*}

As shown in Figure~\ref{fig:feedback-multitask}, multi-task prediction heads expose several aligned sensor and scene representations from a shared evolving world state instead of returning RGB alone.

The most direct strategy is to add multiple prediction heads on an RGB-generation backbone, simultaneously outputting complementary modalities such as depth, semantics, occupancy, normals, and optical flow. Tesseract \cite{2504.20995} demonstrates the feasibility of joint 4D generation of RGB, depth, and normals, learning a unified spatio-temporal representation from multi-view video and simultaneously outputting three essentially different but geometrically related modalities. The multimodal egocentric world model \cite{2412.11198} outputs RGB and depth while finely controlling objects, ego, and human pose, and proposes the COM (Control Over Multi-objects) metric to measure control precision, taking control precision as a quantifiable feedback-quality metric for the first time. The multimodal autoregressive model \cite{2510.09036} simultaneously generates RGB, depth, and arm masks, where depth provides 3D geometric information and the arm mask provides direct localization of the manipulation region, offering directly usable structured feedback for robotic-manipulation tasks. In driving, this strategy has recently been especially active: MUVO \cite{muvo2025} synchronously predicts LiDAR point clouds and 3D occupancy grids in addition to RGB, aligning the world model's output directly with the native sensor formats of the autonomous-driving perception stack; DiST-4D \cite{dist2025} outputs metric depth while generating multi-view RGB through decoupled spatio-temporal diffusion, serving as a bridge from pixels to geometry; and GaussianDWM \cite{2512.23180} uses 3D Gaussians as an intermediate representation to jointly render both RGB and depth. Together these works show that sensor-level feedback (B2) is spreading from a few robotic-manipulation scenes to driving, the application with the largest data scale.

UniScene \cite{2412.05435} adopts a progressive-generation strategy that goes beyond simply adding prediction heads: it first generates semantic occupancy as a meta-scene representation rich in semantics (class labels) and geometry (3D spatial occupancy), then generates video and LiDAR data respectively conditioned on the occupancy. This architecture, from a centralized scene representation to downstream multimodal sensor output, essentially rebuilds in the generation pipeline the information flow of a traditional simulator from world state (occupancy) to multiple sensors (camera, LiDAR), rather than end-to-end generating RGB and outputting other information incidentally. OccLLaMA \cite{2409.03272} and OccSora \cite{2405.20337} integrate occupancy prediction into a language model and a diffusion framework respectively, demonstrating the portability of occupancy as a general scene representation across different generative backbones.

Diffusion-flow on vector maps \cite{2603.17652} represents a severely underrated feedback paradigm---outputting an ego-lane and agent vector map rather than pixels. Designing structured vectors rather than pixels as the native output format has a unique advantage for driving simulation requiring deterministic spatial information (e.g., a surrounding vehicle at 12.3 m travelling at 15.2 m/s), because vector information eliminates the information loss of re-extracting structure from pixels. In LiDAR and semantic prediction, UniScene \cite{2412.05435} and street-view generation \cite{2310.02601} demonstrate the possibility of scene understanding and semantic information as a natural by-product of the world model rather than an independent task requiring an extra model.

\subsubsection{Deriving Structured State from Physical Simulation}

\begin{figure*}[!t]
\centering
\includegraphics[width=.82\textwidth]{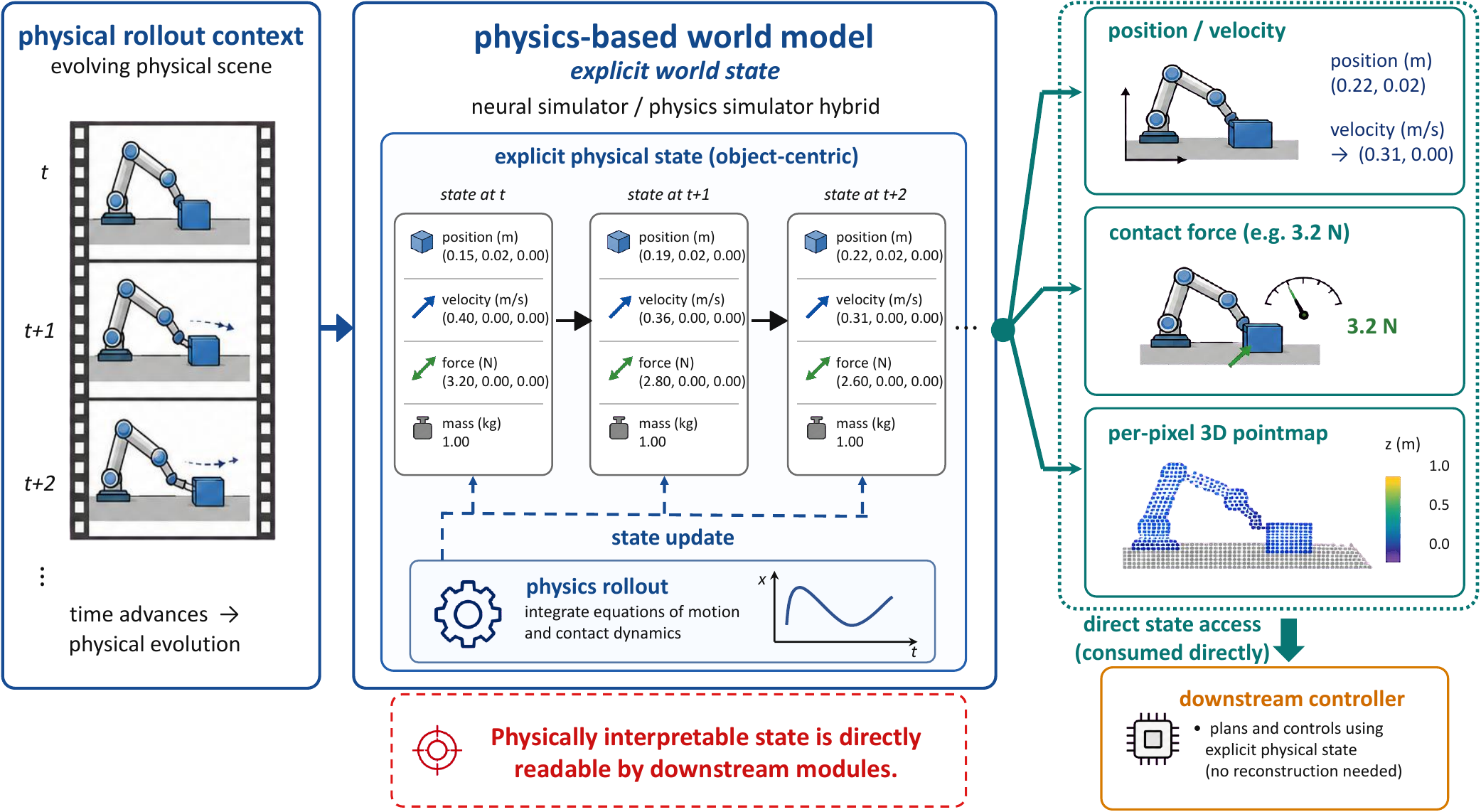}
\caption{\textbf{Structured state derived from physical simulation.} An explicit object-centric trajectory exposes position, velocity, force, and mass over time, from which a controller can directly consume quantities such as contact force and per-pixel 3D pointmaps.}
\label{fig:feedback-physics}
\end{figure*}

Figure~\ref{fig:feedback-physics} illustrates why a physics-based state is particularly useful as feedback: physical quantities can be read directly by downstream controllers without being reconstructed from rendered pixels.

The core advantage of this strategy is the physical interpretability of the state: state derived from a physics engine naturally carries clear physical meaning and can be directly consumed by downstream controllers. OrbiSim's \cite{2605.16395} dual-module architecture naturally separates physical-state computation from visual rendering explicitly; OrbiSim-Dynamics runs in an explicit physical-state space where all intermediate physical quantities (position, velocity, acceleration, contact force, potential energy, inferred physical parameters) exist in structured format and can be accessed downstream with zero information loss. The value of this separation is that downstream consumers can choose which layer to draw information from as needed: a control policy needs only the low-dimensional state vector of position and velocity, a perception model needs the high-dimensional sensor simulation of RGB and depth, and a physical-debugging tool may need contact forces and collision events---all obtainable from the same world model without training a separate model per downstream task.

ChronoDreamer \cite{2512.18619} takes contact force as an explicit prediction target---e.g., the contact force between a robot gripper and the table is 3.2 N---a form of structured feedback with direct, irreplaceable practical value in precision manipulation. Kinema4D \cite{2603.16669} outputs RGB and pointmap as dual feedback, the pointmap providing per-pixel 3D coordinates so that a downstream control system can directly learn the position of a pixel in the world coordinate system without indirect inference through stereo or RGB-D sensors. GEM-4D's \cite{2605.22882} inverse-dynamics module converts a video rollout into an executable robot trajectory---answering what robot action sequence corresponds to this video sequence---a translation from visual state feedback to action state feedback. The 3D-point-flow world model \cite{2601.03782} redefines action as a per-pixel 3D-displacement prediction, unifying the representation of state and action and plugging directly into an MPC framework for real-time planning, with key practical value for robotic control systems requiring millisecond-level response. MVISTA-4D \cite{2602.09878} builds on a view-consistent 4D world model and, via test-time action inference (imagine-then-act), back-solves the action to execute from the generated 4D state, so the structured geometric state output by the world model closes directly onto manipulation decisions---another instance of geometric state feedback driving control.

\subsubsection{Latent Representation as Alternative Feedback}

\begin{figure*}[!t]
\centering
\includegraphics[width=.82\textwidth]{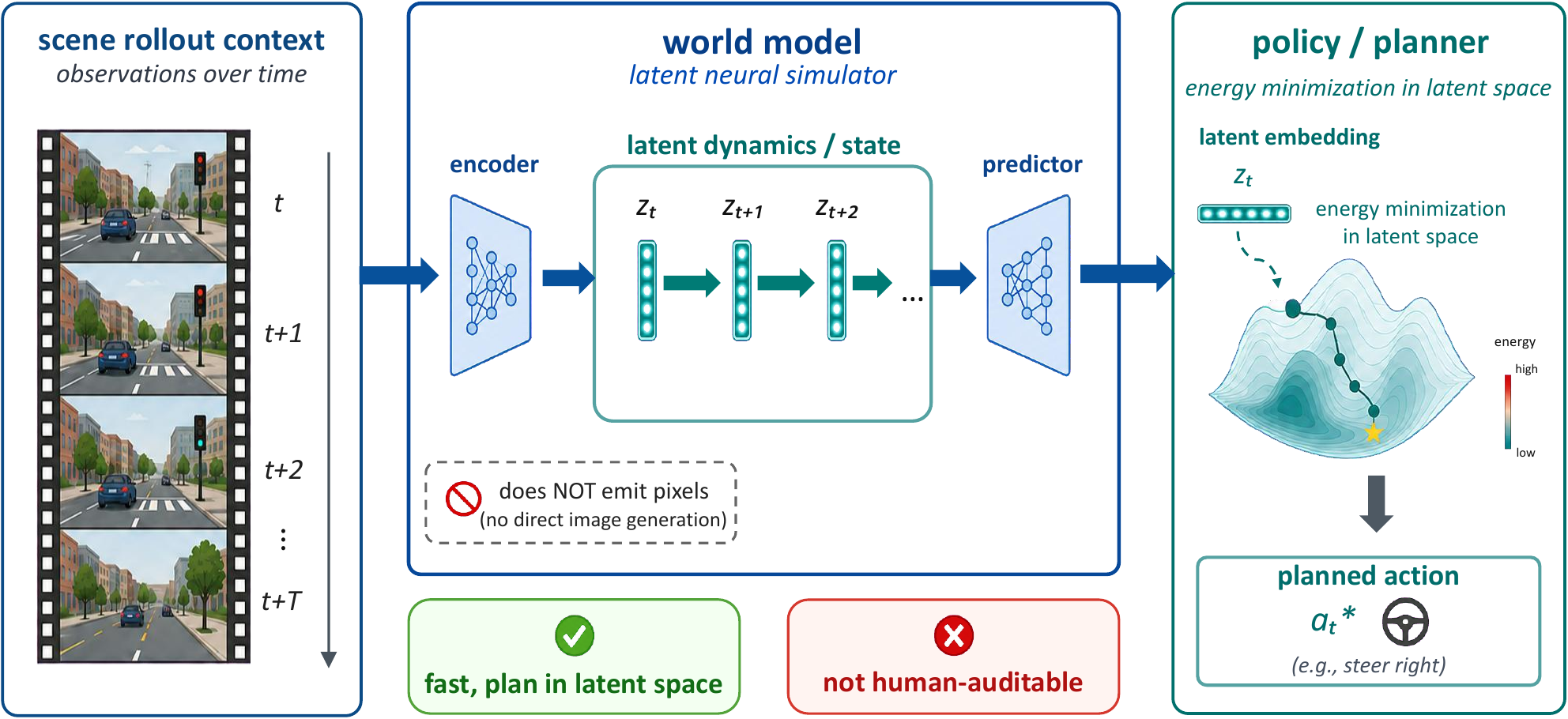}
\caption{\textbf{Latent representation as machine-facing state feedback.} The world model predicts a compact latent trajectory rather than reconstructing pixels, and a policy or planner selects actions by optimizing directly in that representation space; efficiency is gained at the cost of interpretability.}
\label{fig:feedback-latent}
\end{figure*}

At the opposite end of the interface spectrum, Figure~\ref{fig:feedback-latent} shows a compact predictive representation consumed directly by a planner: it is efficient for machine decision-making but not directly auditable by humans.

The JEPA route gives the most radical answer to the state-feedback problem: rather than providing human-readable state feedback, it provides, in latent space, a representation directly consumable by downstream neural networks. V-JEPA~2's \cite{2506.09985} closed-loop planning capability---planning an action sequence via energy minimization in latent space and zero-shot-controlling a real robot---proves that feedback that outputs no readable state can equally be effective feedback; its 15$\times$ speed advantage (16 seconds per action versus Cosmos's 4 minutes) shows that latent-space feedback has engineering superiority under specific conditions. Cosmos Policy \cite{2601.16163} defines the world model's feedback as a latent representation directly fed into a policy model. UWM-JEPA \cite{2605.25313} demonstrates the ability to maintain belief-space structural integrity under the more demanding condition of blind rollouts with occluded observations, reaching 0.77 accuracy versus 0.53 for a parameter-matched LSTM-JEPA, indicating that the internal structure of latent space may be more effective than pixel space at resisting information loss.

But latent-space feedback is constrained by a fundamental problem: auditability. In safety-critical systems, a human operator needs to understand and verify the system's state, which a latent-space vector cannot afford. A trainable representation-translator module that decodes latent representations into human-readable structured information may be key to bridging this gap, but no work in the current literature has systematically explored it.

\subsubsection{Runtime Annotations versus Sensor-Space Estimates}

The audit separates two interface designs that are often conflated. Sensor-space estimates---semantics, depth, occupancy, point flow, or contact maps---remain B2 even when they are produced alongside RGB. B4 instead requires named entity or physics fields that are directly available during rollout. Six systems meet this criterion: Playable Environments \cite{2203.01914} maintains per-object position and pose state; TrafficBots \cite{2303.04116} rolls forward typed agent states; PIN-WM \cite{2504.16693} identifies object dynamics and physical parameters; GigaWorld-0 \cite{2511.19861} exposes joint dynamics and system-identification parameters; VectorWorld \cite{2603.17652} produces a lane--agent vector graph with typed agent fields; and OrbiSim \cite{2605.16395} evolves explicit object-centric physical state. This distinction measures whether a simulator-like annotation interface exists, without treating a sensor prediction or a training label as such an interface.

\subsection{Diversity}\label{sec:c7}

Diversity refers to the richness and controllable stochasticity of a model's generated results in environment layout, object appearance, character behaviour, future evolution, and so on, and is the dimension where a generative world model holds the greatest potential advantage over a traditional simulator. A traditional simulator, limited by a hand-modelled asset library and hand-written behaviour logic, can only generate content from existing models, whereas a world model learns from large-scale internet data and can in principle cover a broader diversity of scenes and behaviours. Of the 200 papers, 46 (23.0\%) list diversity as a principal contribution dimension.

The analysis of diversity unfolds most naturally from the \emph{source} of diversity, because different sources determine the type and boundary of diversity and its trade-off pattern with controllability: coverage inherited from large-scale data, novel combinations induced by compositional instructions, and controlled variation along explicit parameters.

\subsubsection{Scale-Driven Diversity}

\begin{figure*}[!t]
\centering
\includegraphics[width=.82\textwidth]{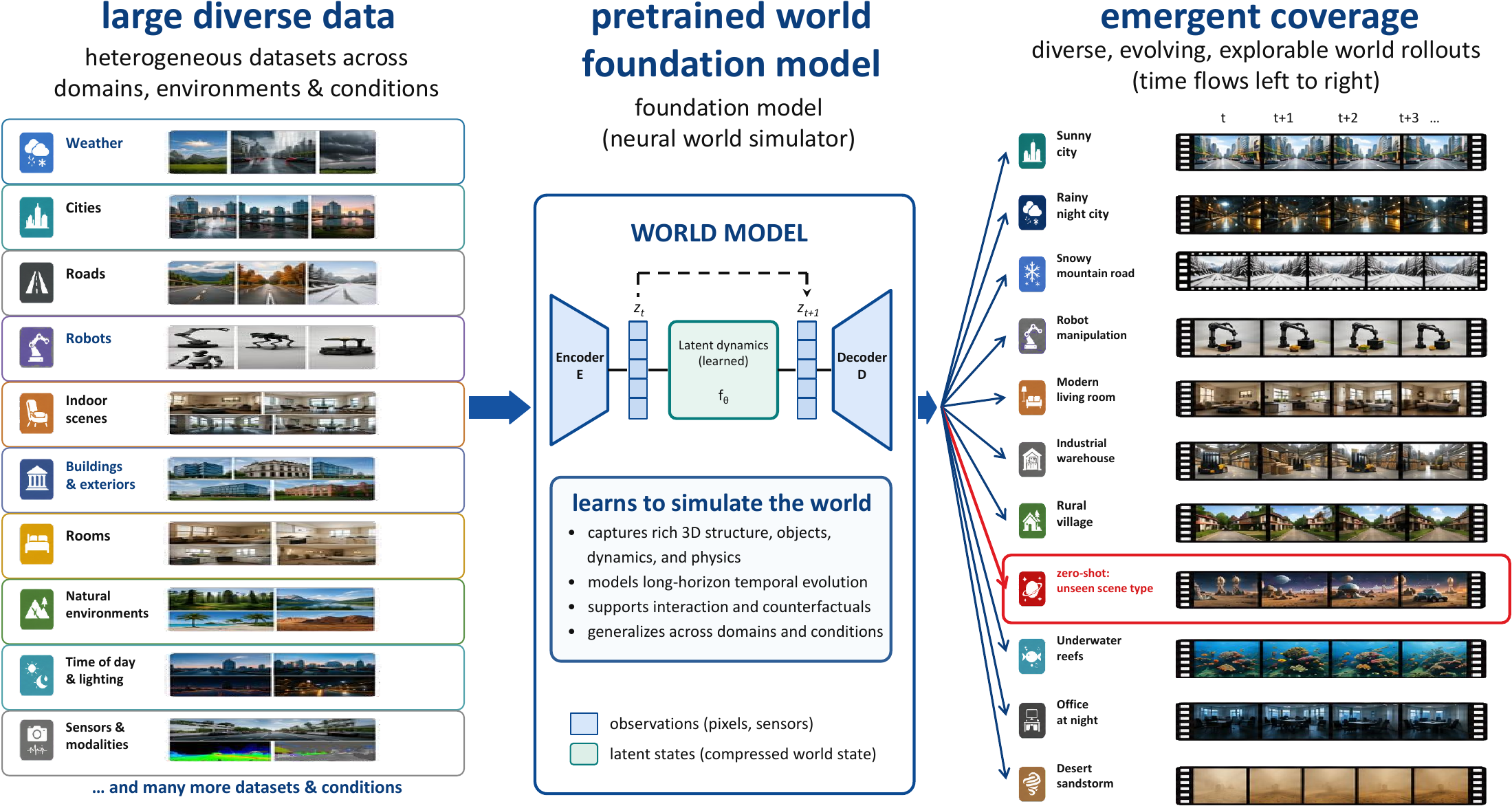}
\caption{\textbf{Scale-driven diversity in a world foundation model.} Heterogeneous data across environments, conditions, and embodiments are compressed into a pretrained model that produces diverse evolving rollouts, including zero-shot coverage beyond the scene types emphasized during training.}
\label{fig:diversity-scale}
\end{figure*}

Figure~\ref{fig:diversity-scale} summarizes the scale-driven route, in which heterogeneous training data broaden the support of a pretrained world model and can yield coverage of unseen scene types.

The most direct source of diversity is large-scale, diverse training data. Cosmos's \cite{2501.03575,2601.16163} WFM design idea---pretrain a general world foundation model for downstream fine-tuning into scene-specific world models---embodies exactly the driving role of data scale on diversity: the large-scale diverse data encountered during pretraining is compressed into model parameters, and fine-tuning releases diversity along specific dimensions per downstream need. GenAD \cite{2403.09630}, trained on over 2000 hours of global multi-weather, multi-traffic driving data, demonstrates zero-shot generalization on unseen driving datasets, proving that diverse training data not only improves performance on known scenes but also lets the model generalize to scene types never directly present in the training data. UniSim \cite{2310.06114} learns a universal real-world interaction simulator by orchestrating multi-source data from driving, robotic manipulation, and indoor navigation, mapping cross-domain data diversity directly into cross-domain scene-generation diversity. The 44k-hour general robot world model \cite{2602.06949} transfers interaction knowledge by learning continuous latent actions on large-scale human video and, via knowledge distillation, raises inference speed to a real-time 10.81 FPS, letting diverse interaction behaviours execute efficiently. VQ-VAE-plus-GPT autoregressive video generation \cite{2104.10157}, as a representative of early autoregressive video generation, already demonstrated the feasibility of a Transformer autoregressively generating diverse, high-fidelity natural-video content in the discrete latent space of a VQ-VAE.

\subsubsection{Composition-Driven Diversity}

\begin{figure*}[!t]
\centering
\includegraphics[width=.82\textwidth]{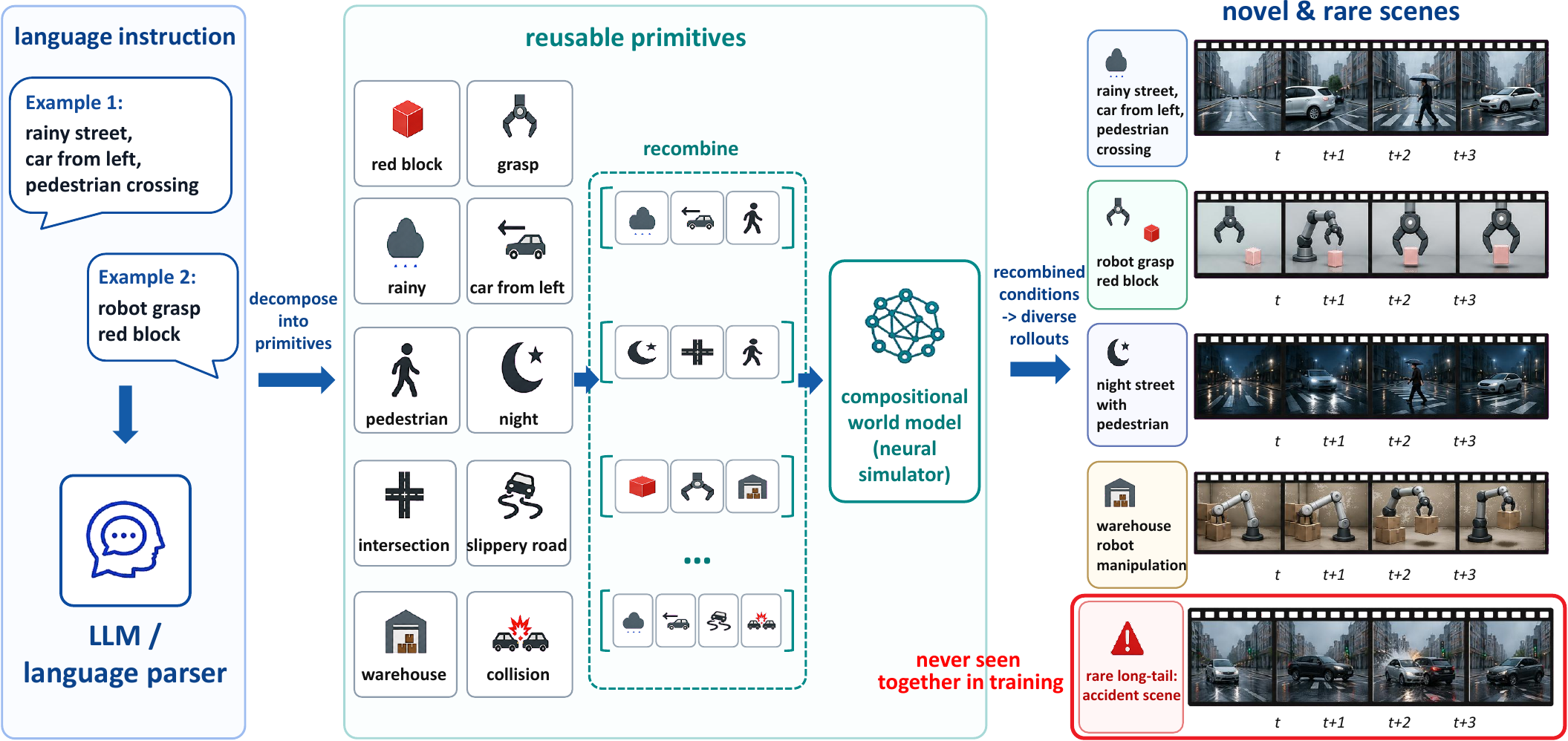}
\caption{\textbf{Compositional diversity through language.} A language parser decomposes instructions into reusable primitives whose new combinations condition a world model to generate novel or rare rollouts, including long-tail interaction and accident scenarios.}
\label{fig:diversity-compositional}
\end{figure*}

The compositional route in Figure~\ref{fig:diversity-compositional} obtains novelty by decomposing instructions into reusable entities, actions, and conditions and recombining them into scenarios not observed as complete examples during training.

The compositionality of natural language is a clever strategy for controllable diversity, specifying which parts should change and into what via language instruction. RoboDreamer \cite{2404.12377} exploits the compositionality of natural language to parse a complex instruction into a set of primitives (e.g., red block plus grasp) and conditions multiple generative models to synthesize planning videos for primitive combinations never seen in training. This decompose-then-recompose strategy achieves \emph{compositional generalization}, letting the model flexibly combine learned primitives in unseen ways. DriveDreamer-2 \cite{2403.06845} introduces an LLM as a diversity amplifier, using the LLM's world knowledge and commonsense reasoning to generate scene descriptions not directly present in the training data, thereby expanding scene diversity in semantic space. Multi-view latent diffusion \cite{2503.20523} generates multi-view video covering rare driving scenes under fine-grained control conditions. The synthetic-data generation pipeline \cite{2506.09042} drives a WFM to generate controllable multi-view long-tail edge scenes, directly used to alleviate the long-tail problem of autonomous-driving perception models. Accident-scene generation \cite{end2025} pushes this idea to extremely rare events: it drives a UniAD agent in a closed loop within the world model to actively generate accident-type dashcam scenes scarce in real data for data augmentation, its value lying precisely in that such high-risk, high-value samples are nearly impossible to obtain through real collection, so the diversity advantage of a generative world model here translates into direct completion of the safety-critical long-tail distribution.

\subsubsection{Controllable Parametric Diversity}

\begin{figure*}[!t]
\centering
\includegraphics[width=.82\textwidth]{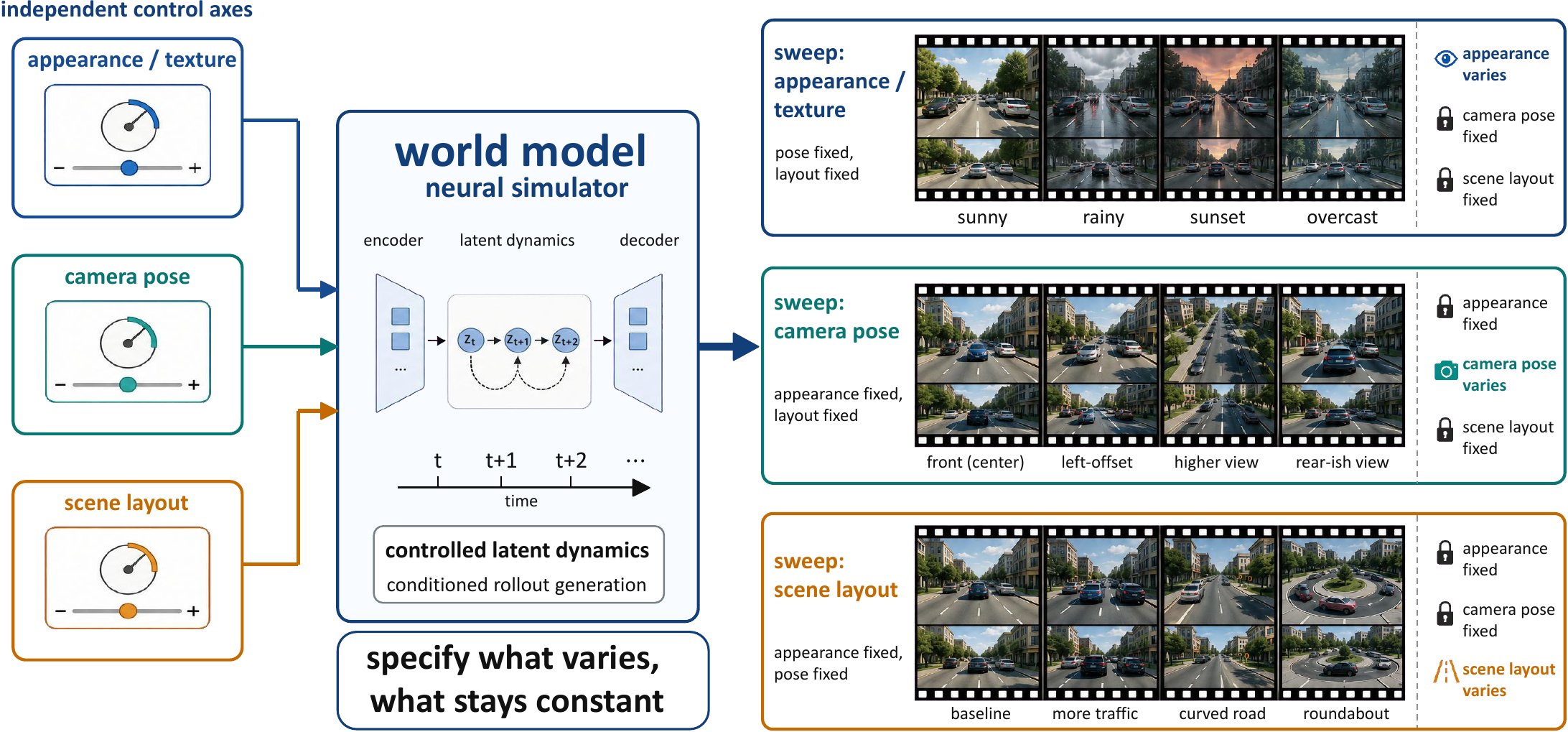}
\caption{\textbf{Parametrically controllable diversity.} Independent controls over appearance, camera pose, and scene layout allow a world model to sweep one factor while keeping the others fixed, separating purposeful variation from unconstrained stochastic generation.}
\label{fig:diversity-parametric}
\end{figure*}

Figure~\ref{fig:diversity-parametric} presents a more controlled form of diversity in which one parameter axis is varied while the remaining scene factors are held fixed.

There is a trade-off between diversity and controllability---higher diversity means poorer control, more precise control means a narrower space---and this trade-off can be managed through parametrically controllable diversity. GigaWorld-0 \cite{2511.19861} demonstrates such fine-grained diversity control, generating diverse embodied sequences and 3D scenes under independent control of appearance (texture, material, lighting), camera view (pose, trajectory), and scene layout (object placement, spatial configuration), so users can precisely specify along which dimensions the world should be diverse and along which consistent. Adaptive multimodal spatial control \cite{2503.14492} supports Sim2Real data augmentation through multimodal spatial-weighted control of segmentation, depth, and edges, where different types of control maps provide precise spatial constraints on which regions should change and which should stay. Pose-free feed-forward 4D reconstruction \cite{2601.00393} achieves scalable 4D reconstruction and novel-view-trajectory generation from in-the-wild monocular video, demonstrating the feasibility of diverse scene reconstruction without the constraint of camera-pose annotation.

\subsubsection{Theoretical Insight and Missing Evaluation of Diversity}

The challenge diversity faces is highly symmetric to controllability: high-diversity models (e.g., Cosmos- and Sora-like) often produce uncontrollable, irrelevant visual content on precise tasks, while high-controllability models degrade sharply out of the training distribution. The root cause of this trade-off is that, in current model design, diversity and controllability share the same capacity budget, and the model must simultaneously compress how many ways the world can look and which one the user wants, information that is redundant in some dimensions and adversarial in others. How to design a controllable-diversity mechanism that lets users specify in a structured way which aspects may vary freely and which must precisely match the specification is a core design challenge for the next generation of world models. In addition, the quantitative evaluation of diversity is severely lacking; at present almost all diversity claims are qualitative, and systematic metrics need to be built from an information-theoretic (conditional entropy, mutual information) or coverage (long-tail coverage, rare-event frequency) angle.

\subsection{Evaluation Metrics}\label{sec:c8}

Evaluation metrics largely determine how fast a field can move: what cannot be measured usually cannot be improved. A traditional simulator has a mature, direct evaluation system, including environment-reward-based RL performance, ground-truth-based perception accuracy, deterministic-reproduction-based A/B testing, and the sim-to-real transfer gap. Evaluating a world model as a simulator faces a dual challenge: as a generative model, its output quality must be evaluated (visual fidelity, temporal consistency); as a simulator, its downstream usability must be evaluated (whether it can replace the real environment to support policy or perception-model training). The two are not always consistent: a visually perfect-looking video may be physically impossible \cite{2605.27589}, and a model with a high open-loop generation score may drift catastrophically in closed-loop interaction \cite{2510.18135}. Of the 200 papers, 51 (25.5\%) list evaluation as a contribution dimension, of which 18 are dedicated benchmarks.

\begin{figure*}[!t]
\centering
\includegraphics[width=.82\textwidth]{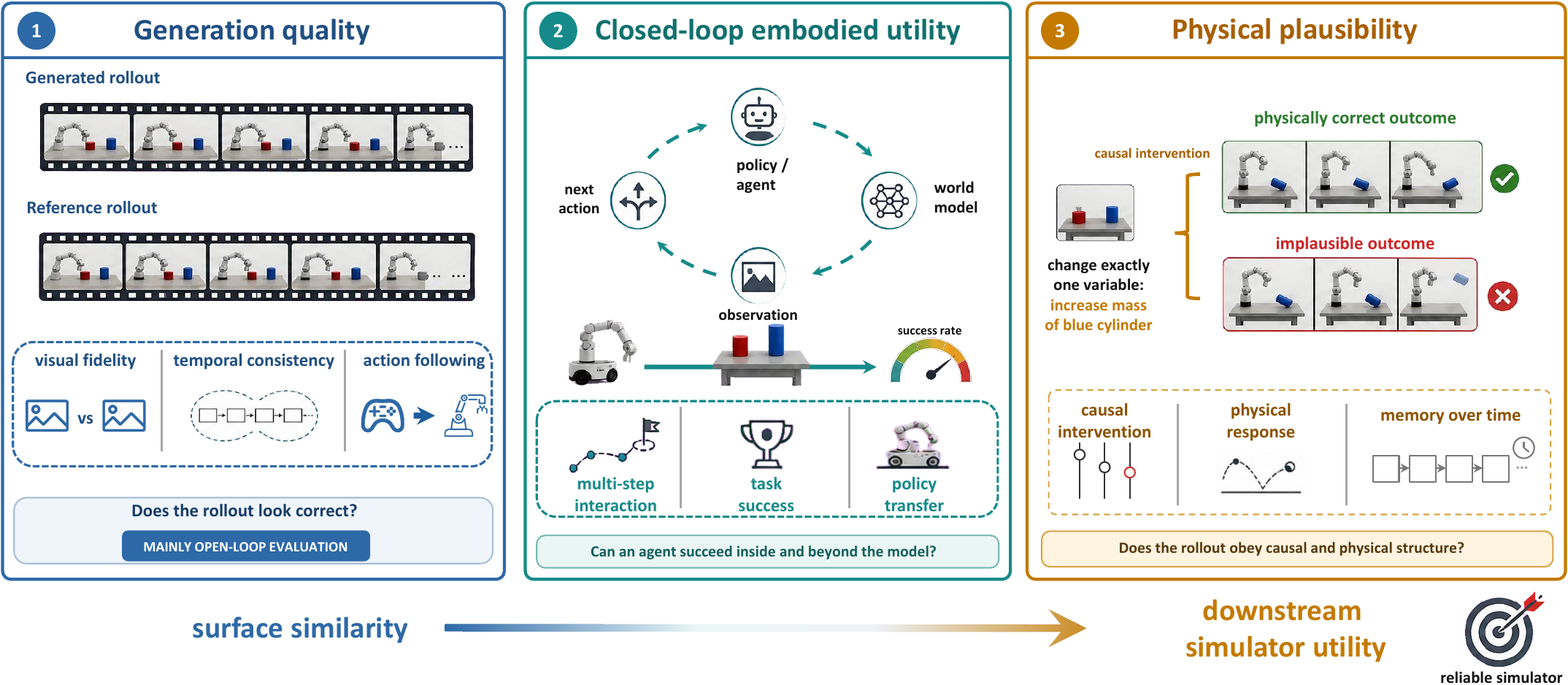}
\caption{\textbf{Three complementary lenses for evaluating world models as simulators.} Generation-quality evaluation measures visual fidelity, temporal consistency, and action following, but remains predominantly open loop. Closed-loop embodied evaluation tests multi-step interaction, task success, and policy transfer. Physical-plausibility evaluation uses causal interventions to assess whether rollouts preserve physical response and state over time. Together, these lenses shift the target from surface similarity toward downstream simulator utility.}
\label{fig:evaluation-lenses}
\end{figure*}

Evaluation works fall naturally into three evaluation lenses ordered by their proximity to downstream utility---generation quality, closed-loop embodied evaluation, and physical plausibility---together with a fourth category that identifies cross-cutting structural gaps in the evaluation ecosystem. Figure~\ref{fig:evaluation-lenses} summarizes the three lenses and emphasizes that simulator evaluation must progress beyond surface similarity toward task-level and causal validity.

\subsubsection{Unified Generation-Quality Evaluation}

WorldScore \cite{2504.00983} is a unified benchmark aiming to provide standardized comparison across world models, covering dimensions such as video quality, temporal consistency, action-following, and physical plausibility. By analogy with GLUE \cite{wang2019glue} and SuperGLUE \cite{wang2019superglue} in NLP, a unified benchmark is crucial for the comparability of community progress, but the task span of video world models is too large---from open-loop FVD \cite{unterthiner2018fvd} to closed-loop success rate to physical-causality verification---so the comprehensiveness of any single metric's coverage is questioned. Nano World Models' \cite{2605.23993} unified evaluation protocol covers dimensions such as generation target, model scale, action conditioning, latent space, evaluation, and long rollout, providing a consistency substrate for reproducible academic evaluation. In the vertical domain, DrivingGen \cite{2601.01528} establishes a comprehensive benchmark for generative video world models in autonomous driving, systematically covering the generation-quality and controllability dimensions of driving scenes and filling the gap left by general benchmarks in characterizing driving-specific needs.

\subsubsection{Closed-Loop Embodied Evaluation}

Closed-loop embodied evaluation is particularly relevant to simulator claims because it measures downstream utility rather than only output appearance. World-in-World \cite{2510.18135} reports that visual quality does not guarantee task success rate, controllability matters more than visual quality, and it reports a scaling relationship for embodied-world-model data. This finding questions the implicit assumption that better FVD equals a better simulator. RoboWM-Bench \cite{2604.19092} realizes an executability-evaluation closed loop from generated video to extracted action to physical-execution verification, directly testing whether the video generated by a world model can translate into a physically feasible robotic manipulation. WorldArena 2.0 \cite{2605.17912} systematically extends the evaluation system along three dimensions---modality (visuo-tactile), function (RL training environment), and platform (simulation plus real robot). iWorld-Bench \cite{2605.03941}, at the evaluation scale of 330k video clips, 6 task categories, and 14 models, provides a large standardized evaluation of interaction capability within the included benchmark records. The 1X World Model Challenge \cite{2510.07092}, in the form of a technical report and a public challenge, establishes a unified task and evaluation protocol for generative world models in humanoid embodied settings, advancing the standardization of closed-loop embodied evaluation in real-robot-related scenarios. WBench \cite{2605.25874} highlights the multi-round nature of interaction and cross-paradigm comparability: with 289 cases and 1058 interaction rounds covering interactive world models of different technical paradigms, it specifically checks whether a model can maintain consistent world evolution under consecutive multi-round actions rather than evaluating only single-step response---which strikes exactly at the essence distinguishing a simulator from single-shot generation, namely that state must stay coherent across repeated interaction. SpatialWorld \cite{2606.09669} evaluates spatial reasoning at the scale of 8 heterogeneous simulation backends and 760 human-annotated tasks and reports a 17.4\% success rate for GPT-5 in its evaluated setting. The Table30 benchmark \cite{2510.17950} provides a reproducible, scalable real-robot VLA-policy evaluation method; the 4D interaction-response benchmark \cite{2603.22212} evaluates the interaction quality of a world model by measuring the causal impact of interaction actions on state transitions; and the verifiable game-agent benchmark \cite{2604.07429} realizes standardized closed-loop interaction evaluation on 34 games and 170 tasks.

\subsubsection{Physical-Plausibility Evaluation}

What-If World's \cite{2605.27589} causal-intervention methodology and APEO scoring use 319 paired interventions to test whether generated consequences change in the expected direction. ReactSim-Bench \cite{2606.14058}, with decoupled ego and agent control, checks reaction plausibility under out-of-distribution behaviour. WorldBench \cite{2601.21282} isolates physical concepts such as friction, viscosity, and intuitive physics. MIND \cite{2511.20937} evaluates memory retention through forward- and inverse-modelling reranking with 8972 question--answer pairs; in its evaluated setting, the performance gap between VLMs and humans widens as the rollout horizon increases.
\subsubsection{Four Evaluation Limitations}

Across the included benchmark literature, four recurring evaluation limitations are visible:
\begin{itemize}
\item \textbf{Benchmark fragmentation.} There is no recognized central benchmark, and the protocols, dimensions, and datasets of different benchmarks are mutually incompatible.
\item \textbf{Open-loop bias.} The closed-loop cumulative-error effect is an important attribute of a simulator, yet most evaluations cannot capture it---what truly needs attention is whether the model is still in a reasonable world after 1000 steps, not the next-frame PSNR under teacher forcing.
\item \textbf{Systematic absence of steppable-environment attributes.} Almost no benchmark measures the incremental success rate---the success rate of deploying to the real environment a policy trained inside the world model---which is precisely the most direct measure of a simulator's value.
\item \textbf{Missing comparison with traditional simulators.} Existing benchmarks only make relative comparisons among world models, lacking an absolute baseline of how traditional simulators perform on these metrics.
\end{itemize}
The root cause of these gaps is that evaluation thinking remains centred on generation quality rather than downstream utility. World-in-World \cite{2510.18135} and RoboWM-Bench \cite{2604.19092} are exemplary in the paradigm shift, but their current community influence remains limited.

\section{Future Directions and Challenges}\label{sec:future}

Based on the comparative analysis in Section~\ref{sec:comparison}, the gap between world models and a rigorous simulator is real and structural, not a marginal quantitative issue. Yet the clear identification of these gaps is itself progress, because only by knowing where the gaps lie can we design targeted research to close them.

\subsection{Formalized Physics}

The physical simulation of current world models relies entirely on soft constraints implicitly learned from data; a breakthrough requires deep fusion of a differentiable physics engine (e.g., OrbiSim \cite{2605.16395}) with the generative pipeline, using the physics solver as a conditional regularizer and a source of structured state feedback. PIN-WM's \cite{2504.16693} parameter identification, ChronoDreamer's \cite{2512.18619} contact-force prediction, and PhysBridge's \cite{2603.05449} physics-simulator bridge already show explicit physical modelling to be feasible; CP4D's \cite{2606.09187} physics-prior-plus-diffusion-commonsense hybrid and ReconDreamer-RL's \cite{2508.08170} kinematics-diffusion separation show modular physical modelling to be practical. The real difficulty is the tension identified in Section~\ref{sec:c2}: a differentiable engine is exact but can only simulate phenomena whose equations have been written down, whereas a data-driven model is flexible but prone to visual shortcuts, and this conflict cannot be removed by enlarging the model or the data. The open question is therefore not whether physics can be injected, but whether generated output can carry step-wise verifiable conservation constraints while retaining open-domain generalization---What-If World's \cite{2605.27589} causal-intervention protocol offers a way to measure such progress, and whether certification can extend from rigid bodies to fluids and deformation will set the reach of formalized physics.

\subsection{Unified Action Interface}

Explicit control is precise but generalizes poorly, language instruction is semantically readable but imprecise, and latent action generalizes well but is unreadable; the paradigm chasm among the three needs bridging in two directions. The first is the joint learning of multi-scale action representations: DiLA's \cite{2605.15725} content--structure separation and DisCo's \cite{2606.07967} discrete primitives both indicate that, along certain dimensions, coarse-grained discretization outperforms fine-grained continuity because separability matters more than resolution. The second is incorporating LLM world knowledge into the action-translation pipeline, where DreamX-World 1.0 \cite{2606.16993} and DriveDreamer-2 \cite{2403.06845} show preliminary feasibility. Harder than representation learning is the temporal compositionality dissected in Section~\ref{sec:c4}: when action segments are concatenated, the conditional observation of a later segment has already drifted from the true world state, so the segments lose causal coupling---at root a gap between open-loop sequence continuation and closed-loop feedback. Only V-JEPA~2's \cite{2506.09985} receding-horizon MPC and WorldVLA's \cite{2506.21539} attention mask so far touch genuine closed-loop control, and how to transplant the mature feedback-plus-feedforward framework of control theory into a high-dimensional generative model, so that the action interface stays closed-loop-consistent over millisecond-level rollouts, remains unresolved.

\subsection{First-Class State Feedback}

The state-interface audit motivates three distinct development targets. Multi-task prediction heads, as used by Tesseract \cite{2504.20995}, UniScene \cite{2412.05435}, and multimodal egocentric world models \cite{2412.11198}, can expose estimated structure alongside RGB. Explicit-state systems such as OrbiSim \cite{2605.16395} can provide quantities tied to an interpretable dynamics state. Representation translators may make JEPA-style latent state inspectable. The key research question is not simply whether a model can emit semantics, depth, or boxes, but whether it can expose named entity and physical fields that remain consistent with the generated observation throughout a rollout. Benchmarks should therefore score sensor-space estimates separately from runtime entity/physics annotations.

\subsection{Long-Horizon Stability}

Suppressing autoregressive drift admits two complementary strategies: a periodic reset-and-anchor mechanism, for which Lyra~2.0's \cite{2604.13036} geometry routing and DecMem's \cite{2605.31336} sparse global memory provide a technical basis; and quantitative characterization of how stability changes with scale and horizon, for which World-in-World \cite{2510.18135} reports a scaling relationship and MIND's \cite{2511.20937} memory evaluation supplies a methodology. As Section~\ref{sec:c5} notes, drift claims across architectures, scene complexities, and action-sequence lengths still lack a common protocol. Two open questions follow. First, for a given architecture and action distribution, can the growth of rollout error be bounded, and how do model capacity and distillation affect it? Second, how should repeat-run reproducibility be reported under controlled randomness? The included corpus provides little comparable evidence on exposed seeds, deterministic kernels, tolerance bands, or repeat-run variance. Standardizing these factors is necessary for A/B testing, regression testing, and safety analysis built on world-model rollouts.

\subsection{Downstream-Utility Evaluation}

The core metric should shift from generation quality to the incremental success rate---the success rate, in the real environment, of an RL policy trained inside the world model. RoboWM-Bench's \cite{2604.19092} generation-and-execution verification and World-in-World's \cite{2510.18135} practice of training agents inside the world model are early exemplars. The resistance to this shift lies in the four structural gaps summarized in Section~\ref{sec:c8}: benchmark fragmentation, open-loop bias, the absence of steppable-environment attributes, and the missing absolute comparison with traditional simulators. The most fundamental is that evaluation thinking remains centred on generation quality---a model with a strong FVD may, after 1000 closed-loop steps, already sit in an implausible world, which current metrics do not catch. A downstream-utility-centred paradigm therefore has to answer two questions: how to measure the incremental success rate cheaply without a full real-robot deployment each time, and how to use a traditional simulator as an absolute baseline that quantifies how far a world model still falls short, rather than only ranking world models against one another.

\subsection{Cross-Route Fusion}

The first five directions each target a single capability, whereas the capability-by-capability comparison of Section~\ref{sec:comparison} reveals a more holistic opportunity: the routes' strengths are complementary---diffusion in vision and interaction, latent dynamics in control decisions, JEPA in efficiency, explicit 3D/4D in geometric consistency. Several fusion directions therefore stand out: embedding a differentiable physics engine into the diffusion pipeline (OrbiSim \cite{2605.16395} and PhysBridge \cite{2603.05449}); combining explicit 3D memory with implicit video generation (Lyra~2.0 \cite{2604.13036} and GWM \cite{2508.17600}); unifying VLA and world models ($\tau_0$-WM \cite{2606.01027} and WorldVLA \cite{2506.21539}); and the pretrain-plus-fine-tune paradigm of the world model as a holistic data engine (GigaWorld-0 \cite{2511.19861} and Cosmos \cite{2501.03575}). The fundamental obstacle is not engineering assembly but that the routes' optimization objectives are not naturally compatible---pixel-reconstruction loss, latent-space energy, and policy return live in different metric spaces, and naively summing them lets the terms work against one another, which calls for re-examining the routes from the unified framework $T\colon\mathcal{S}\times\mathcal{A}\to\Pi(\mathcal{S}\times\mathcal{F})$. Being-H0.7's \cite{2605.00078} train-time-fuse, inference-time-separate design hints at one way around the objective conflict, but how to compose progress on the six directions into a single system that meets all requirements of a rigorous simulator at once remains the open question with no precedent and the greatest value.

\section{Conclusion}\label{sec:conclusion}

This study asks whether the generative process of a world model can become a simulator in the strict sense. Using the eight capabilities of a traditional simulator as a unified yardstick, it performs a systematic comparative analysis and paper-level evidence mapping of a curated 200-paper corpus.

On the core question of whether it can become a rigorous simulator, our judgement is: under restricted conditions it can, but in the strict sense it remains a critical step away. Generative world models, with their open-domain generalization learned from data and their visual quality approaching photorealism, are advancing rapidly toward a new generation of intelligent simulation environments. In interaction and controllability, through autoregressive-diffusion distillation \cite{2604.08995,2605.30263}, latent-action learning \cite{2402.15391,2605.15725}, 4D geometric-control representation \cite{2601.05138}, and world-foundation-model platforms \cite{2501.03575}, world models have demonstrated the feasibility of functional substitution in specific scenarios. However, on the formal guarantees of the physics engine, the richness and structuredness of state feedback---the complete implementation audit confirms B2 in 45 of 163 papers and B4 in only six---the theoretical guarantee of long-horizon stability, and the downstream-utility orientation of the evaluation system, world models remain a critical step away from traditional simulators.

On the question of how far each route has progressed, the capability-by-capability comparison shows that autoregressive models have the broadest interaction coverage, while diffusion models remain prominent in high-fidelity visual generation \cite{2408.14837,2604.08995}; both families nevertheless have weak coverage of physics and state feedback. Latent dynamics is distinguished by its support for control decisions \cite{2301.04104} but has limited visual-generation ability; the JEPA route has outstanding efficiency, with a roughly 15$\times$ speed advantage \cite{2506.09985}, but unreadable output; and the explicit 3D/4D route has a structural advantage in geometric consistency \cite{2604.13036,2503.18945,2508.17600} but appearance detail inferior to pure diffusion methods. Table~\ref{tab:matrix} provides the corpus-level coverage evidence for these route-dependent trade-offs and, in particular, shows that state feedback C6 remains a structural gap across most model families.

On where to go next, we distil six research directions---formalized physics, a unified action interface, first-class state feedback, long-horizon stability, downstream-utility evaluation, and cross-route fusion---each supported by existing nascent work as evidence of feasibility, together forming an actionable roadmap toward the next generation of true simulators.

We hope the eight-capability comparison framework and the full-corpus evidence map established by this study can provide a systematic coordinate system for tracking progress. We also call on the research community to turn the proposition of ``world model as simulator'' from an aspirational slogan into a research agenda with rigorous definitions of capability, gap, and evaluation. Only when the goal expands from generating good-looking video to outputting consumable, verifiable, reproducible world state can a generative world model truly be used as a world, rather than merely looking like one.

\appendices

\section{Complete List of Search Queries}\label{app:queries}

This appendix reports the complete keyword protocol used in Section~\ref{sec:collection}. Each query is assigned a stable identifier and a short scope label so that the coverage of the search can be inspected without parsing an undifferentiated code block. The Boolean expressions are reproduced verbatim apart from line wrapping. Results were ranked by relevance, truncated to the top entries for each query, de-duplicated against the accumulated corpus, and then screened manually. The search cut-off was 30 June 2026.

\subsection{Mainstream Routes and Application Settings}

These queries cover the principal generative architectures, interaction settings, and application domains used to form the core candidate pool.

\begingroup
\sloppy
\queryitem{M1}{Interactive video simulation}{interactive AND world AND model AND simulation AND video AND generation}
\queryitem{M2}{Playable neural game engines}{world AND model AND game AND engine AND playable AND real-time}
\queryitem{M3}{Robot manipulation}{action AND conditioned AND world AND model AND robot AND manipulation AND simulation}
\queryitem{M4}{Latent-action models}{latent AND action AND world AND model AND video}
\queryitem{M5}{Autoregressive video models}{autoregressive AND world AND model AND video AND tokenizer AND interactive}
\queryitem{M6}{Autonomous-driving simulation}{driving AND world AND model AND autonomous AND simulation AND generative}
\queryitem{M7}{Embodied physical simulation}{embodied AND world AND model AND physics AND simulator}
\queryitem{M8}{Diffusion game simulators}{neural AND game AND simulator AND diffusion AND world AND model}
\queryitem{M9}{Evaluation benchmarks}{world AND model AND benchmark AND evaluation AND interactive AND simulation}
\queryitem{M10}{General surveys}{general AND world AND model AND survey AND simulator}
\queryitem{M11}{Real-time streaming}{real-time AND generative AND world AND model AND streaming}
\queryitem{M12}{Explorable 4D worlds}{4D AND world AND model AND scene AND generation AND explorable}
\endgroup

\subsection{Supplementary Representations and Capabilities}

These queries target representations and capabilities that broad world-model searches tend to miss, including predictive embeddings, explicit geometry, memory, multimodality, and physics-oriented evaluation.

\begingroup
\sloppy
\queryitem{S1}{Joint-embedding prediction}{joint AND embedding AND predictive AND architecture AND world AND model AND JEPA}
\queryitem{S2}{Self-supervised representation learning}{self-supervised AND world AND model AND representation AND learning AND prediction}
\queryitem{S3}{Language models for planning}{large AND language AND model AND as AND world AND model AND planning AND reasoning}
\queryitem{S4}{Text-based simulation}{language AND model AND world AND model AND text AND adventure AND simulation}
\queryitem{S5}{3D Gaussian representations}{3D AND gaussian AND splatting AND world AND model AND scene AND generation}
\queryitem{S6}{Neural radiance fields}{neural AND radiance AND field AND world AND model AND dynamic AND scene}
\queryitem{S7}{Differentiable physics}{differentiable AND physics AND simulation AND world AND model AND learned}
\queryitem{S8}{Neural rigid-body dynamics}{physics AND engine AND neural AND world AND model AND rigid AND body AND dynamics}
\queryitem{S9}{Long-horizon consistency}{long-term AND memory AND consistent AND world AND model AND video AND generation}
\queryitem{S10}{Persistent spatial memory}{persistent AND world AND model AND spatial AND memory AND exploration}
\queryitem{S11}{Embodied navigation}{embodied AND navigation AND world AND model AND spatial AND reasoning}
\queryitem{S12}{Occupancy prediction}{occupancy AND world AND model AND 3D AND scene AND autonomous AND prediction}
\queryitem{S13}{Multimodal prediction}{multimodal AND world AND model AND audio AND video AND tactile AND prediction}
\queryitem{S14}{Physical-plausibility evaluation}{physical AND plausibility AND world AND model AND evaluation AND benchmark AND video}
\queryitem{S15}{General-purpose foundation models}{foundation AND world AND model AND general AND purpose AND interactive AND simulation}
\queryitem{S16}{Camera- and action-controllable video}{controllable AND video AND generation AND world AND model AND camera AND action}
\endgroup

\medskip
\noindent\textit{Execution note.} On arXiv, each expression was applied to the \texttt{all} field; on Google Scholar, DBLP, and Crossref, the same concept-term conjunction was used without a field prefix. Citation-chain expansion and publication-status verification, described in Section~\ref{sec:collection}, were conducted as separate cross-checks and are therefore not represented as keyword queries above.

{\small
\bibliographystyle{IEEEtran}
\bibliography{references}}

\end{document}